\documentclass[acmsmall,natbib=true]{acmart}\usepackage[]{graphicx}\usepackage[]{xcolor}
\makeatletter
\def\maxwidth{ %
  \ifdim\Gin@nat@width>\linewidth
    \linewidth
  \else
    \Gin@nat@width
  \fi
}
\makeatother

\definecolor{fgcolor}{rgb}{0.345, 0.345, 0.345}

\usepackage{framed}
\makeatletter
\newenvironment{kframe}{%
 \def\at@end@of@kframe{}%
 \ifinner\ifhmode%
  \def\at@end@of@kframe{\end{minipage}}%
  \begin{minipage}{\columnwidth}%
 \fi\fi%
 \def\FrameCommand##1{\hskip\@totalleftmargin \hskip-\fboxsep
 \colorbox{shadecolor}{##1}\hskip-\fboxsep
     \hskip-\linewidth \hskip-\@totalleftmargin \hskip\columnwidth}%
 \MakeFramed {\advance\hsize-\width
   \@totalleftmargin\z@ \linewidth\hsize
   \@setminipage}}%
 {\par\unskip\endMakeFramed%
 \at@end@of@kframe}
\makeatother

\definecolor{shadecolor}{rgb}{.97, .97, .97}
\definecolor{messagecolor}{rgb}{0, 0, 0}
\definecolor{warningcolor}{rgb}{1, 0, 1}
\definecolor{errorcolor}{rgb}{1, 0, 0}
\newenvironment{knitrout}{}{} 

\usepackage{alltt}

\usepackage{multirow}
\usepackage{epstopdf}
\usepackage{subcaption}
\usepackage{tikz}
\usetikzlibrary{positioning, shapes, arrows, shadows, arrows.meta}

\def \parrotpdf {\includegraphics[]{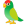}}
\DeclareUnicodeCharacter{1F99C}{\parrotpdf}
\usepackage{tabularx}
\usepackage[utf8]{inputenc}
\usepackage{wrapfig}
\usepackage[T1]{fontenc}
\usepackage{textcomp}
\usepackage{listings}
\usepackage{xcolor}

\definecolor{codegreen}{rgb}{0,0.6,0}
\definecolor{codegray}{rgb}{0.5,0.5,0.5}
\definecolor{codepurple}{rgb}{0.58,0,0.82}
\definecolor{backcolour}{rgb}{0.95,0.95,0.92}

\lstdefinestyle{mystyle}{
  backgroundcolor=\color{backcolour}, commentstyle=\color{codegreen},
  keywordstyle=\color{magenta},
  numberstyle=\tiny\color{codegray},
  stringstyle=\color{codepurple},
  basicstyle=\ttfamily\footnotesize,
  breakatwhitespace=false,         
  breaklines=true,                 
  captionpos=b,                    
  keepspaces=true,                 
  numbers=left,                    
  numbersep=5pt,                  
  showspaces=false,                
  showstringspaces=false,
  showtabs=false,                  
  tabsize=2
}

\usepackage{graphicx}
\usepackage{enumerate}

\usepackage{amsmath}

\usepackage{subcaption}

\hypersetup{colorlinks=true, linkcolor=black, citecolor=black, filecolor=blue,
     urlcolor=blue, unicode=true}

\usepackage[american]{babel}
\usepackage{csquotes}

\IfFileExists{upquote.sty}{\usepackage{upquote}}{}
\begin{document}
\title{Misclassification in Automated Content Analysis Causes Bias in Regression. Can We Fix It? Yes We Can!}

\author{Nathan TeBlunthuis}
\authornote{This article has been accepted for publication in Communication Methods \& Measures, published by Taylor \& Francis.
Dr. TeBlunthuis lead the project conception, wrote the initial draft, and developed the simulations, real-data example, and error correction methods. Dr. Hase designed and lead the literature review. Dr. Chan contributed to the project conception, to the literature review, by leading the R package development, and by advising. All authors contributed to writing. The bulk of Dr. TeBlunthuis' contributions were done as an affiliate of Northwestern University.}
\email{nathante@uw.edu}
\affiliation{\institution{School of Information, University of Michigan}\country{USA}}
\affiliation{\institution{Department of Communication Studies, Northwestern University}
\country{USA}}

\author{Valerie Hase}
\affiliation{\institution{Department of Media and Communication, LMU Munich}\country{Germany}}

\author{Chung-hong Chan}
\affiliation{\institution{GESIS - Leibniz-Institut für Sozialwissenschaften}\country{Germany}}

\renewcommand{\shortauthors}{TeBlunthuis et al.}

\begin{abstract}
Automated classifiers (ACs), often built via supervised machine learning (SML), can categorize large, statistically powerful samples of data ranging from text to images and video. They have become widely popular measurement devices in communication science and related fields.
Despite this popularity, even highly accurate classifiers make errors that cause misclassification bias and misleading results when input to downstream statistical analyses—unless such analyses account for these errors.
As we show in a systematic literature review of SML applications, 
communication scholars largely ignore misclassification bias.
In principle, existing statistical methods can use ``gold standard'' validation data, such as that created by human annotators, to correct misclassification bias.
We introduce and test such methods, including a new method we design and implement in the R package \texttt{misclassificationmodels}, via Monte Carlo simulations designed to reveal each method's limitations, which we also release. Based on our results, we recommend our new error correction method as it is versatile and efficient. In sum, automated classifiers, even those below common accuracy standards or those making systematic misclassifications, can be useful for measurement with careful study design and appropriate error correction methods.
\end{abstract}

\maketitle

\emph{Automated classifiers} (ACs) based on supervised machine learning (SML) have rapidly gained popularity
as part of the \emph{automated content analysis} toolkit in communication science \citep{baden_three_2022}. With ACs, researchers can categorize large samples of text, images, video or other types of data into predefined categories \citep{scharkow_thematic_2013}. Studies for instance use SML-based classifiers to study frames \citep{burscher_teaching_2014}, tonality \citep{van_atteveldt_validity_2021}, 
or civility \citep{hede_toxicity_2021} in news media texts or social media posts.

However, there is increasing concern that automated classifiers' imperfections may compromise measurement validity for studying theories and concepts from communication science \citep{baden_three_2022, hase_computational_2022}. 
Research areas where ACs have the greatest potential—e.g., content moderation, social media bots, affective polarization, or radicalization—are haunted by the specter of methodological questions related to misclassification bias \citep{rauchfleisch_false_2020}: How accurate must an AC be to measure a variable? Can an AC built for one context be used in another \citep{burscher_using_2015,hede_toxicity_2021}? Is comparing automated classifications to some external ground truth sufficient to claim validity?
How do biases in AC-based measurements affect downstream statistical analyses \citep{millimet_accounting_2022}? 

Social scientists using ACs are broadly aware that they have imperfections, and consensus best practices are to acknowledge this fact and report quantifiers of classifier performance such as precision and recall.  Such acknowledgment is an important step, but clearly falls short of a satisfying answer to the methodological questions above.
We introduce and analyze \emph{misclassification bias}: How misclassifications by ACs distort statistical findings in downstream analysis---unless correctly modeled \citep{fong_machine_2021}.

Our study begins with a demonstration of misclassification bias in a real-world example based on the Perspective toxicity classifier.
Next, we provide a systematic literature review of $N = 48$ studies employing SML-based text classification.
Although communication scholars have long scrutinized related questions about manual content analysis for which they have recently proposed statistical corrections  \citep{bachl_correcting_2017, geis_statistical_2021}, misclassification bias in automated content analysis is largely ignored.
Our review demonstrates a troubling lack of attention to the threats ACs introduce and virtually no mitigation of such threats. As a result, in the current state of affairs, researchers are likely to either draw misleading conclusions from inaccurate ACs or avoid ACs in favor of costly methods such as manually coding large samples \citep{van_atteveldt_validity_2021}. 
 
Our primary contribution, an effort to rescue ACs from this dismal state, is to \emph{introduce and test methods for correcting misclassification bias} \citep{carroll_measurement_2006, buonaccorsi_measurement_2010, yi_handbook_2021}. We consider three recently proposed methods: \citet{fong_machine_2021}'s generalized method of moments calibration method, \citet{zhang_how_2021}'s pseudo-likelihood models,  and \citet{blackwell_unified_2017-1}'s application of imputation methods. To overcome these methods' limitations, we draw a general likelihood modeling framework  from the statistical literature on measurement error \citep{carroll_measurement_2006} and tailor it to the problem of misclassification bias. Our novel implementation is the experimental R package \texttt{misclassificationmodels}.

 We test these four error correction methods and compare them against ignoring misclassification (the naïve approach) and refraining from automated content analysis by only using manual coding (the feasible approach). We use Monte Carlo simulations to model four prototypical situations identified by our review: Using ACs to measure either (1) an independent or (2) a dependent variable where the classifier makes misclassifications that are either (a) easy to correct (when an AC is unbiased and misclassifications are uncorrelated with covariates i.e.,  \emph{nonsystematic misclassification}) or (b) more difficult (when an AC is biased and misclassifications are correlated with covariates i.e., \emph{systematic misclassification}).

According to our simulations, even biased classifiers without high predictive performance can be useful in conjunction with appropriate validation data and error correction methods.
As a result, we are optimistic about the potential of ACs and automated content analysis for communication science and related fields—if researchers can correct for misclassification. 
Current practices of ``validating'' ACs by making misclassification rates transparent via metrics such as the F1 score, however, provide little safeguard against misclassification bias.

In sum, we make a methodological contribution by introducing the often-ignored problem of misclassification bias in automated content analysis, testing error correction methods to address this problem via Monte Carlo simulations, and introducing a new method for error correction.
Profoundly, we conclude that automated content analysis will progress not only---or even primarily---by building more accurate classifiers but through rigorous human annotation and statistical error modeling.

\section{Why Misclassification is a Problem: an Example Based on the Perspective API}

There is no perfect AC. All ACs make errors.
This inevitable misclassification causes bias in statistical inference \citep{carroll_measurement_2006, scharkow_how_2017}, leading researchers to make both type-I (false discovery) and type-II errors (failure to reject the null) in hypotheses tests. To illustrate the problematic consequences of this misclassification bias, we focus on real-world data and a specific research area in communication research: detecting and understanding harmful social media content. Communication researchers often employ automated tools such as the Perspective toxicity classifier \citep{cjadams_jigsaw_2019} to detect toxicity in online content \citep[e.g.,][]{hopp_social_2019, kim_distorting_2021, votta_going_2023}.
As shown next, however, relying on toxicity scores created by ACs such as the Perspective API as (in-)dependent variables produces different results than using measurements created via manual annotation.

To illustrate this, we use the Civil Comments dataset released in 2019 by Jigsaw, the Alphabet corporation subsidiary behind the Perspective API. Methodological details on the data and our example are available in Appendix \ref{appendix:perspective}. The dataset has 448,000 English-language comments made on independent news sites. It also includes manual annotations of each comment concerning its toxicity (\emph{toxicity}), whether it discloses aspects of personal identity like race or ethnicity \emph{(identity disclosure)}, and the number of likes it received \emph{(number of likes)}.

In addition to manual annotations of each comment, we obtained AC-based toxicity classifications from the Perspective API in November 2022. Perspective's toxicity classifier performs very well, with an accuracy of 92\% and an F1 score of 0.79. Nevertheless, if we treat human annotations as the ground-truth, the classifier makes systematic misclassifications: It disproportionately misclassifies comments as toxic if they disclose racial or ethnic identity (Pearson's $\rho=0.121;~ 95\%~CI=[0.12,0.124]$). This seemingly small degree of misclassification can have nontrivial consequences for the overall accuracy of results as shown below. We will illustrate this through two separate research questions (RQs) researchers may test based on automated classifications from the Perspective API.

Consider the following exemplary research question that social scientists may ask: How does positive feedback (likes) to more-or-less toxic behavior in online comments depend on whether comments also disclose identity? They may use a logistic regression model to predict whether a comment contains \emph{identity disclosure} using \emph{number of likes}, \emph{toxicity}, and their interaction as independent variables (IVs).
\begin{figure}[htbp!]
\centering
\begin{subfigure}{\linewidth}
\begin{knitrout}
\definecolor{shadecolor}{rgb}{0.969, 0.969, 0.969}\color{fgcolor}
\includegraphics[width=\maxwidth]{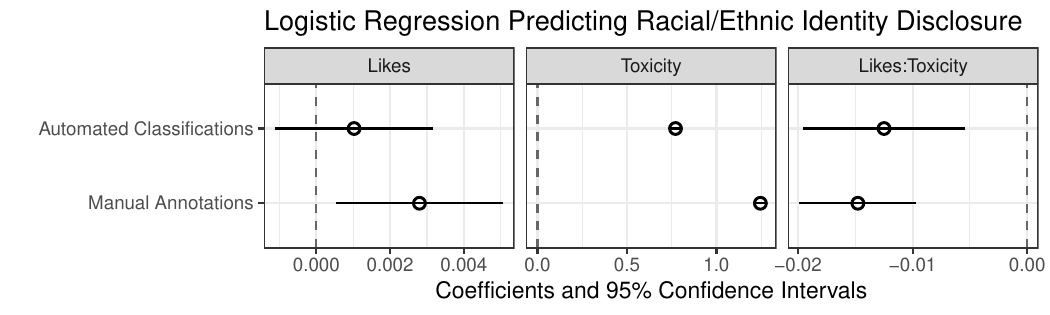} 
\end{knitrout}
\subcaption{\emph{Example 1}: Misclassification in an independent variable.\label{fig:real.data.example.iv}}
\end{subfigure}

\begin{subfigure}{\linewidth}
\begin{knitrout}
\definecolor{shadecolor}{rgb}{0.969, 0.969, 0.969}\color{fgcolor}
\includegraphics[width=\maxwidth]{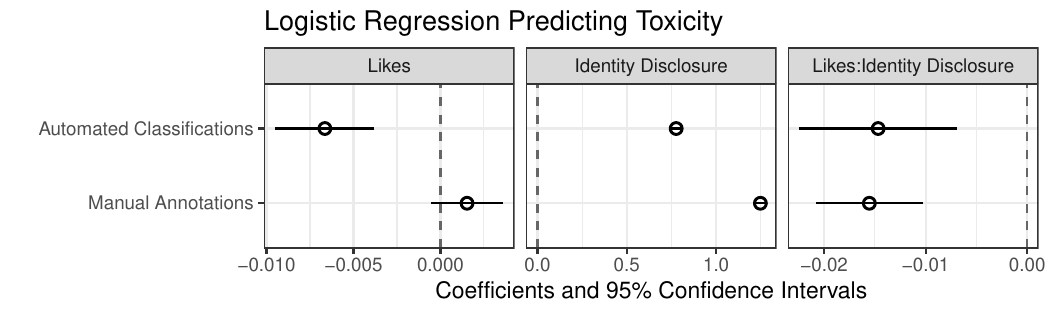} 
\end{knitrout}
\subcaption{\emph{Example 2}: Misclassification in a dependent variable. \label{fig:real.data.example.dv}}

\end{subfigure}

\caption{Bias through Misclassification: a Real-World Example Using the Perspective API and the Civil Comments Dataset. 
\label{fig:real.data.example}
}
\end{figure}
Figure \ref{fig:real.data.example.iv}  illustrates what happens in this case, i.e. when \emph{ignoring misclassification in an IV}: Researchers may reject a hypothesized direct relationship between likes and identity disclosure because the model indicates that that their correlation is mediated by toxicity, but the direct relationship is not well estimated.

Thus, even a very accurate AC can introduce type-II errors, i.e. researchers failing to rejecting a null hypothesis due to misclassification.

Now consider a second exemplary RQ: How does toxic behavior in online comments depend on whether comments disclose identity and on the positive feedback they receive (likes)? Researchers of this question may predict the \emph{toxicity} of a comment with \emph{number of likes}, \emph{identity disclosure} in a comment, and their interaction as IVs.
Figure \ref{fig:real.data.example.dv} illustrates what happens in this case, i.e., when \emph{ignoring misclassification in a dependent variable (DV)}: Using Perspective's classification of toxicity results in a small negative direct effect of likes. However, there is no detectable relationship when using manual annotations. In this case misclassification leads to type-I error, i.e., false discovery of a nonzero relationship.

\section{Why Transparency about Misclassification Is Not Enough}

Although the Perspective API is certainly accurate enough to be useful to content moderators, the example above demonstrates that this does not imply usefulness for social science  \citep{grimmer_machine_2021-1}.
Machine learning takes the opposite position on the bias-variance trade-off than conventional statistics does and achieves high predictiveness at the cost of more biased inference \citep{breiman_statistical_2001}. As a growing body of scholarship critical of the hasty adoption of machine learning in criminal justice, healthcare, or content moderation demonstrates,
ACs boasting high performance often have biases related to social categories \citep{barocas_fairness_2019}. Such biases often result from non-representative training data and spurious correlations that neither reflect causal mechanisms nor generalize to different populations \citep{bender_dangers_2021}.
Much of this critique targets unjust consequences of these biases to individuals. Our example shows that these biases can also contaminate scientific studies using ACs as measurement devices. Even very accurate ACs can cause both type-I and type-II errors, which become more likely when  classifiers are less accurate or more biased or when effect sizes are small.

We argue that current common practices to address such limitations are insufficient. These practices aim to characterize the quality of measurement tools by reporting classifier performance on manually annotated data quantified via metrics like accuracy, precision, recall, or the F1 score \citep{hase_computational_2022, baden_three_2022, song_validations_2020}.
While such steps may promote confidence in the reliability of results and are an important first step, they cannot provide conclusive evidence for or against misclassification bias. Despite high predictiveness, bias can still flow downstream into statistical inferences and, ultimately, threaten the validity of results, as our example demonstrates. 
Instead of relying only on transparency rituals to ward off misclassification bias, researchers can and should use validation data to correct it.

These claims may surprise because of the wide-spread misconception that misclassification causes only conservative bias (i.e., bias towards null effects). This is believed because it is true for bivariate least squares regression when misclassifications are nonsystematic
\citep{carroll_measurement_2006, loken_measurement_2017, van_smeden_reflection_2020}.\footnote{The measurement error literature defines categories of measurement error, of which we will discuss four:
(1) Measurement error in an IV is \emph{classical} when the measurement equals the true value plus noise.  In other words, it is nonsystematic and the variance of an AC's predictions is greater than the variance of the true value \citep{carroll_measurement_2006}. (2) \emph{Berkson} measurement error in an IV is when the truth equals the measurement plus noise. In agreement with prior work, we assume that nonsystematic measurement error from ACs is classical because an AC is an imperfect model of reality \citep{fong_machine_2021, zhang_how_2021}. Similarly, it is hard to imagine how an AC would have Berkson errors as predictions would then have lower variance than the training data. We thus do not consider Berkson error. (3) Measurement error in an IV is called \emph{differential} if it is not conditionally independent of the DV given the other IVs. 
(4) Measurement error in a DV is called \emph{systematic} when it is correlated with an IV. We use this more general term to simplify our discussions that pertain equally to misclassified IVs and DVs.}  As a result, researchers interested in a hypothesis of a statistically significant relationship may not consider misclassification an important threat to validity \citep{loken_measurement_2017}. 

As our example shows, however, misclassification bias can be anti-conservative \citep{carroll_measurement_2006, loken_measurement_2017, van_smeden_reflection_2020}. In regression models with more than one IV or in nonlinear models, such as the logistic regression we used in our example, even nonsystematic misclassification can cause bias away from 0. 
Moreover, systematic misclassification can bias inference in any direction.

ACs designed in one context and applied in another are likely to commit systematic misclassification. The Perspective API, for example, was developed for social media comments but performs much worse when applied to news data \citep{hede_toxicity_2021}. Systematic misclassification may also arise when an AC used for measurement shapes behavior in a sociotechnical system under study. As examples, the Perspective API is used for online forum moderation \citep{hede_toxicity_2021}, as is the ORES API for Wikipedia moderators \citep{teblunthuis_effects_2021}. 
Misclassification from such classifiers can be systematic because they have causal effects on outcomes related to moderation.

If ACs become standard measurement devices, for instance
Google's Perspective API for measuring toxicity \citep[see critically][]{hosseini_deceiving_2017} or Botometer for classifying social media bots \citep[see critically][]{rauchfleisch_false_2020}, entire research areas may be subject to systematic biases.
Even if misclassification bias is usually conservative, it can slow progress in a research area.   Consider how \citet{scharkow_how_2017} argue that media's ``minimal effects'' on political opinions and behavior in linkage studies may be an artifact of measurement errors in both manual content analyses and self-reported media use in surveys.  Conversely, if researchers selectively report statistically significant hypothesis tests, misclassification can introduce an upward bias in the magnitude of reported effect sizes and contribute to a replication crisis \citep{loken_measurement_2017}.

\section{Quantifying the Problem: Error Correction Methods in SML-based Text Classification}

To understand how social scientists, including communication scholars, engage with the problem of misclassification in automated content analysis,
we conducted a systematic literature review of studies using supervised machine learning (SML) for text classification (see Appendix \ref{appendix:lit.review} in our Supplement for details).\footnote{Automated content analysis includes a range of methods both for assigning content to predefined categories (e.g., dictionaries) and for assigning content to unknown categories (e.g., topic modeling) \citep{grimmer_text_2013}. While we focus on SML, our arguments extend to other approaches such as dictionary-based classification and even beyond the specific context of text classification.}
Our sample consists of studies identified by similar reviews on automated content analysis \citep{baden_three_2022, hase_computational_2022, junger_unboxing_2022, song_validations_2020}. Our goal is not to comprehensively review all SML studies 
but to provide a picture of common practices, with an eye toward awareness of misclassification and its statistical implications. 

We identified a total of 48 empirical studies published between 2013 and 2021, more than half of which were published in communication journals. Studies used SML-based text classification for purposes such as to measure frames \citep{opperhuizen_framing_2019} or topics \citep{vermeer_online_2020}. They often employed SML-based ACs to create dichotomous (50\%) or other categorical variables (23\%).\footnote{Metric variables were created in 35\% of studies, mostly via the non-parametric method by \citet{hopkins_method_2010}.} Of these empirical studies, many used SML-based ACs as IVs (44\%) or DVs (40\%) in multivariate analyses. 90\% reported univariate statistics such as proportions.

Overall, our review reveals that \emph{studies almost never reflected upon nor corrected misclassification bias}. According to our review, 85\% of studies reported metrics such as recall or precision, but only 19\% of studies explicitly stated that an AC misclassified texts which may introduce measurement error. Only a single article reported using error correction methods. This indicates a clear need to better understand and address misclassification bias, we now turn towards error correction methods for doing so.

\section{Addressing the Problem: Approaches for Correcting Misclassification}
Statisticians have extensively studied measurement error (including misclassification), the problems it causes for statistical inference, and methods for correcting these problems \citep[see][]{carroll_measurement_2006, fuller_measurement_1987}. 
In this paper, we propose \emph{Maximum Likelihood Adjustment} (MLA), method capable of fixing misclassification bias when an automatic classifier measures either an IV or a DV and makes either random or systematic misclassifications. 
This section explains our proposed approach and then compares it with three alternatives for error correction: \citet{fong_machine_2021}'s GMM calibration method (GMM), multiple imputation (MI) \citep{blackwell_multiple_2012}, and \citet{zhang_how_2021}'s pseudo-likelihood models (PL). Table \ref{tab:methods.tested} summarizes each method and the conditions where they are effective according to our simulations. \footnote{Statisticians have studied other methods including simulation extrapolation and score function methods. As we argue in Appendix \ref{appendix:other.methods}, these error correction methods are not advantageous when manually annotated data is available, as is often the case with ACs.}

\begin{table}
\centering
\footnotesize
    \caption{Overview of error correction methods}
    \label{tab:methods.tested} 
\begin{tabular}{lllcccc}
\toprule
    Method & Abbrev. & Key Citation & \multicolumn{2}{l}{Random Misclass.} & \multicolumn{2}{l}{Systematic Misclass.} \\ \cmidrule(l){4-5}\cmidrule(l){6-7} 
    & & & IV & DV & IV & DV \\ \hline
    GMM calibration & GMM & \citet{fong_machine_2021} & Yes & No & No & No \\
    Multiple Imputation & MI & \citet{blackwell_multiple_2012} & Yes & Yes & Yes & No \\
    \multirow{2}{10em}{Pseudo-likelihood modeling} & PL & \citet{zhang_how_2021} & No & No & No & No \\ & & & & & &\\
    \multirow{1}{10em}{Maximum likelihood adjustment (ours)} & MLA &  \citet{carroll_measurement_2006} & Yes & Yes & Yes & Yes \\ & & & & & &  \\ \hline
\end{tabular}

\end{table}

For clarity, we first restate the problem in mathematical notation. In \emph{the IV case}, we want to estimate a regression model $Y=B_0+B_1X+B_2Z+\varepsilon$ where $X$ is only observed in a small sample of manually annotated data $\tilde{X}$, but automated classifications $W$ of $X$ are observed.
To illustrate, in our first real-world example, $X$ is toxicity, $\tilde{X}$ are the civil comment annotations, $W$ are the Perspective API's toxicity classification, $Z$ are likes, and $Y$ is identity disclosure.
Similarly, in \emph{the DV case}, we only observe the true value of $Y$ in a small sample $\tilde{Y}$.  In our second real-word example, $Y$ is toxicity.

Across both cases, a sample of annotated data ($\tilde{X}$ or $\tilde{Y}$) may be too small to provide convincing evidence, but collecting additional annotations is too expensive. In contrast, an AC can make classifications $W$ for the entire dataset but introduces misclassification bias. How can we correct this bias in an automated content analysis?

Say the sample of annotated data $X^*$ is too small to convincingly test a hypothesis, but collecting additional annotations is too expensive.
In contrast, an AC can make classifications $W$ for the entire dataset but introduces misclassification bias. How can we correct this bias in an automated content analysis?

In this paper, we propose \emph{Maximum Likelihood Adjustement} (MLA) for correcting misclassification bias. 
MLA tailors \citet{carroll_measurement_2006}'s presentation of the general statistical theory of likelihood modeling for measurement error correction to the context of automated content analysis.\footnote{In particular see Chapter 8 (especially example 8.4) and Chapter 15 (especially 15.4.2).} We give a brief intuitive explanation here. A more rigorous explanation is found in Appendix \ref{appendix:mla}.

MLA works in slightly different ways depending on if an IV or a DV is automatically classified, but the intuition is similar.
In both cases, MLA depends on an \emph{error model} of the automated classifications $W$ given $X$, $Y$, and $Z$ \citep{carroll_measurement_2006}.
It may seem surprising that the error model predicts the automated classifications using validation data instead of the other way around, as is the case in the first stages of the MI and GMM approaches (see below). In MLA, the role of the error model is not to make new automated classifications, but to account for the relationship between automated classifications and the annotations and how this relationship varies depending on other observable variables.

In the IV case, MLA constitutes the use of a single \emph{joint model} comprised of the error model, the \emph{main model} of scientific interest, and an \emph{exposure model} giving the probability of $X$.  Unlike the MI and GMM techniques discussed below, MLA estimates the joint model all at once using likelihood maximization. When annotations $\tilde{X}$ are available, the data are entirely observed and are included directly in the model. Otherwise, when $X$ is unobserved, the model uses a synthetic copy of the unobserved data for each possible value of $X$ (e.g., if $X$ is binary, there are two synthetic datasets: one with $X=1$ and another with $X=0$.) The synthetic data are weighted by the probability of $W$ given $X$ according to the error model and by the probability of $X$ according to the exposure model. This way, MLA accounts for all possible values of $X$ and ``integrates out'' the unobserved data. Intuitively, when $X$'s true value is unknown, MLA uses information about how the probability of the observed $W$ depends on the possible values of $X$ and the probability that $X$ takes each of these values instead. The difference in the DV case is that there is no exposure model because the main model gives the probability of $Y$.



MLA has four advantages in the context of ACs that reflect the benefits of integrating out partially observed discrete variables and maximum likelihood estimation (MLE)\footnote{It is worth noting that an analogous method can be implemented in a Bayesian framework instead of MLE. A Bayesian approach may provide additional strengths in flexibility and uncertainty quantification, but bears additional complexity.}. First, it is generally applicable to any model with a convex likelihood (i.e., an optimizer can find a unique set of parameters that best fit the data). Such models include generalized linear models (GLMs) and generalized additive models (GAMs).
Second, assuming the model is correctly specified, MLA estimators are fully consistent (i.e., as sample size increases, estimates converge to the true value) \citep{carroll_measurement_2006}.  This is related to MLA's excellent statistical efficiency; it requires less manually annotated data to make precise estimates. Third, it is applicable when either an IV or the DV is automatically classified. Finally, MLA uses established methods for estimating confidence intervals, such as the Fischer information and profile likelihood methods, that effectively account for uncertainty from all the joint model's components.

\subsection{Other Approaches to Correcting Misclassification}

We now compare our proposed MLA approach to others recently discussed by social scientists and discuss how the MLA approach is advantageous.

\subsubsection{Regression Calibration}
\emph{Regression calibration} uses observable IVs, including automated classifications $W$ and other variables measured without error $Z$, to approximate the true value of $X$ \citep{carroll_measurement_2006}. \citet{fong_machine_2021} propose a regression calibration procedure designed for SML that we refer to as \emph{GMM calibration} or GMM.\footnote{\citet{fong_machine_2021} describe their method within an instrumental variable framework, but it is equivalent to regression calibration, the standard term in measurement error literature.} For their calibration model, \citet{fong_machine_2021} use 2-stage least squares (2SLS). They regress observed variables $Z$ and AC predictions $W$ onto the manually annotated data and then use the resulting model to approximate $X$ as $\hat{X}$. They then use the generalized method of moments (gmm) to combine estimates based on the approximated IV $\hat{X}$ and estimates based on the manually annotated data $\tilde{X}$. This makes efficient use of manually annotated data and provides an asymptotic theory for deriving confidence intervals. An additional strength compared to the other methods is that GMM does not require the analyst to make assumptions about the distribution of the outcome $Y$. 

The GMM approach bears limitations that MLA overcomes. Systematic misclassification can invalidate GMM when AC predictions $W$ are not conditionally independent of $Y$ given $X$.  In the MLA framework, including $Y$ in the error model for $X$ can account for such systematic misclassification. 
This technique, however, does not translate to the 2SLS framework. 2SLS assumes that the instrumental variable (i.e., $W$) is conditionally independent of $Y$ given $X$. In other words, regression calibration assumes that measurement error is non-differential \citep{carroll_measurement_2006}.

\subsubsection{Multiple Imputation}
\emph{Multiple imputation} (MI) treats misclassification as a missing data problem. Communication scholars may be familiar with MI as a method for analyzing datasets with missing values, for instance in survey studies \citep{myers_goodbye_2011,cho_geography_2011}. MI can also be applied to misclassification bias when validation data are available. The true value of $X$ can be treated as observed in manually annotated data $\tilde{X}$ and missing otherwise \citep{blackwell_unified_2017-1}. 

Like MLA, MI uses a model to infer likely values of possibly misclassified variables. The difference is that multiple imputation samples several (hence \emph{multiple} imputation) entire datasets by filling in the missing data from the predictive probability distribution of $X$ conditional on other variables $\{W,Y,Z\}$. MI then runs a statistical analysis on each of these sampled datasets and pools the results of each of these analyses \citep{blackwell_multiple_2012}. Note that $Y$ can be included among the imputing variables, giving MI the potential to address systematic misclassification.
 \citet{blackwell_multiple_2012} claim that MI is relatively robust when it comes to small violations of the assumption of nondifferential error. Moreover, in theory, MI can be used for correcting misclassifications both in IVs and DVs. 

MLA's advantages compared to MI come not from its flexibility (MI also affords flexible specifications), but from its constraints tailored to the problem of misclassification bias.  MI is designed for applicability to a wide range of problems involving missing continuous as well as categorical variables. This generality is unnecessary for addressing misclassified variables, which are categorical by definition. Somewhat contradictorily, MI assumes that the covariance matrix of the data ($Y$,$X$,$Z$) is multivariate normal, an assumption that is never true for categorical data and helps explain why (as shown below) MI struggles particularly in the case where an AC predicts $Y$. In sum, when it comes to fixing misclassification bias, we argue that MI is both more complex and requires stronger assumptions than necessary.

\subsubsection{Pseudo-Likelihood}
\emph{``Pseudo-likelihood''} methods (PL)—even if not always explicitly labeled this way—are another approach for correcting misclassification bias. \citet{zhang_how_2021} proposes a method that approximates the error model using quantities from the AC's confusion matrix—the positive and negative predictive values in the case of a mismeasured IV and the AC's false positive and false negative rates in the case of a mismeasured DV.  Because quantities from the confusion matrix are neither data nor model parameters, \citet{zhang_how_2021}'s method is technically a ``pseudo-likelihood'' method. A clear benefit is that this method does not require validation data, only summary statistics which may be easier to obtain. 

PL is conceptually similar to MLA. Indeed, for non-systematic misclassification, PL and MLA maximize almost identical likelihoods. The difference is that instead of estimating the error model from data, PL uses parameters derived from quantities that summarize misclassification (e.g., precision and recall).  These statistics do not account for systematic misclassification and so PL is unsuited for such.  Moreover, PL overstates the precision of its estimates because it does not account for uncertainty in these statistics.

 


\section{Evaluating Misclassification Models: Monte Carlo Simulations}

We now present four Monte Carlo simulations (\emph{Simulations 1a}, \emph{1b}, \emph{2a}, and \emph{2b}) with which we evaluate existing methods (GMM, MI, PL) and our approach (MLA) for correcting misclassification bias.

Monte Carlo simulations are a tool for evaluating statistical methods, including (automated) content analysis \citep[e.g.,][]{song_validations_2020,bachl_correcting_2017,geis_statistical_2021, fong_machine_2021,zhang_how_2021}.
They are defined by a data generating process from which datasets are repeatedly sampled. Repeating an analysis for each of these datasets provides an empirical distribution of results the analysis would obtain over study replications. Monte Carlo simulation affords exploration of finite-sample performance, robustness to assumption violations, comparison across several methods, and ease of interpretability \citep{mooney_monte_1997}. 
Such simulations allow exploration of how results depend on assumptions about the data-generating process and analytical choices and are thus an important tool for designing studies that account for misclassification.

\subsection{Parameters of the Monte Carlo Simulations}

The simulations reported in our main text are designed to (1) verify that error correction methods are effective when they should be effective in theory, and (2) to demonstrate the promising potential of error correction methods in automated content analysis. The classifiers in these scenarios have predictive performance substantially lower than is commonly held to be necessary for reliable measurement via ACs. We chose to not study all possible combination of simulation parameters to limit the energy consumption of fitting thousands or even millions of models to generated datasets, but focus on the most illustrative and common cases. Our simulations have relatively small samples and large effect sizes, but simulations with larger samples and smaller effects would inevitably obtain qualitatively similar results. Our substantive findings about the methods' effectiveness are not sensitive to these choices as shown by additional simulations in appendix \ref{appendix:robustness} with varying classifier accuracy and systematic error, imbalance in predicted variables, and correlation between $X$ and $Y$.

In our simulations, we tested four error correction methods: \emph{GMM calibration} (GMM) \citep{fong_machine_2021}, \emph{multiple imputation} (MI) \citep{blackwell_multiple_2012}, \emph{Zhang's pseudo-likelihood model} (PL) \citep{zhang_how_2021}, and our \emph{Maximum Likelihood Adjustment} approach (MLA). We use the \texttt{predictionError} R package \citep{fong_machine_2021} for the GMM method, the \texttt{Amelia} R package with 200 imputed datasets for the MI approach, and our own implementation of \citet{zhang_how_2021}'s PL approach in R.
We develop our MLA approach in the R package \texttt{misclassificationmodels}. 
For PL and MLA, we quantify uncertainty using the Fisher information quadratic approximation.\footnote{The code for reproducing our simulations and our experimental R package is available here: \url{https://osf.io/pyqf8/?view_only=c80e7b76d94645bd9543f04c2a95a87e}.} 

In addition, we compare these error correction methods to two common approaches in communication science: the \emph{feasible} estimator (i.e., conventional content analysis that uses only manually annotated data and not ACs)
and the \emph{naïve} estimator (i.e., using AC-based classifications $W$ as stand-ins for $X$, thereby ignoring misclassification).
According to our systematic review, the \emph{naïve} approach reflects standard practice in studies employing SML for text classification.

We evaluate each of the six analytical approaches in terms of \emph{consistency} (whether the estimates of parameters $\hat{B_X}$ and $\hat{B_Z}$ have expected values nearly equal to the true values $B_X$ and $B_Z$), \emph{efficiency} (how precisely the parameters are estimated and how precision improves with additional data), and \emph{uncertainty quantification} (in terms of the empirical coverage probability of the 95\% confidence intervals estimated in our simulations. Ideally, 95\% confidence intervals contain the true parameter value in 95\% of simulations).
To evaluate efficiency, we repeat each simulation with different amounts of total observations, i.e., unlabeled data to be classified by an AC (ranging from 1000 to 10000 observations), and manually annotated observations (ranging from 100 to 400
observations). Since our review indicated that ACs are most often used to create binary variables, we restrict our simulations to misclassifications related to a binary (in-)dependent variable.

\subsection{Four Prototypical Scenarios for Our Monte Carlo Simulations}

We simulate regression models with two IVs ($X$ and $Z$). This sufficiently constrains our study's scope but the scenario is general enough to be applied in a wide range of research studies. 

Whether the methods we evaluate below are effective or not depends on the conditional dependence structure among IVs $X$ and $Z$, the DV $Y$, and automated classifications $W$.
This structure determines if adjustment for systematic misclassification is required \citep{carroll_measurement_2006}.
Bayesian networks provide a useful graphical representation of conditional independence structure \citep{pearl_fusion_1986} and we use such networks to illustrate our scenarios in Figure \ref{bayesnets}.\footnote{Bayesian networks are similar to causal directed acyclic graphs (DAGs), an increasingly popular representation in causal inference methodology.  DAGs are Bayesian networks with directed edges that indicate the direction of cause and effect. We use Bayesian networks for generality and because causal direction is not important in the measurement error theory we use.}
An edge between two variables in these networks indicates that they have a direct relationship.  Two nodes are \emph{conditionally independent} given a set of other variables that \emph{separate} them (i.e., the two nodes would be disconnected if the set of nodes is removed from the network).
For example, in Figure \ref{fig:simulation.1a}, the automatic classifications $W$ are conditionally independent of $Y$ given $X$ because all paths between $W$ and $Y$ contain $X$. This indicates that the model $Y=B_0 +B_1 W+ B_2 Z$ (the \emph{naïve estimator}) has non-differential error because the automated classifications $W$ are conditionally independent of $Y$ given $X$. However, in Figure \ref{fig:simulation.1b}, there is an edge between $W$ and $Y$ to indicate that $W$ is not conditionally independent of $Y$ given other variables. Therefore, the naïve estimator has differential error. To illustrate, this figure corresponds to our first real-data example in which Perspective's toxicity classifications $W$ are correlated with identity disclosure $Y$ even after accounting for $X$ (as follows from the correlation of missclassification ($W-X$) with $Y$).

We first simulate two cases where an AC measures an IV without (\emph{Simulation 1a}) and with differential error (\emph{Simulation 1b}). Then, we simulate using an AC to measure a DV, either one with misclassifications that are uncorrelated (\emph{Simulation 2a}) or correlated with an IV (\emph{Simulation 2b}). GMM is not designed to correct misclassifications in DVs, so we omit this method in \emph{Simulations 2a} and \emph{2b}. 

Our main simulations below are designed to demonstrate how error correction methods can use validation data together with low-quality automated classifications to obtain precise and consistent estimates and to clarify the conditions where our proposed method is effective and others are not. Additional simulations presented in appendix \ref{appendix:robustness} evaluate the methods with varying degrees of classifier accuracy, systematic classification, skewness in the classified variable, error model misspecification, violation of parametric assumptions, and correlation between the IVs.  These additional simulations show that our main qualitative results are not sensitive to our choices of simulation parameters and help support our recommendations. 

\tikzset{
  observed/.style={circle, draw},
  partly observed/.style 2 args={draw, fill=#2, path picture={
      \fill[#1, sharp corners] (path picture bounding box.south west) -|
      (path picture bounding box.north east) -- cycle;},
     circle},
    unobserved/.style={draw, circle, fill=gray!40},
    residual/.style={draw, rectangle}
}
\begin{figure}[htbp!]
\centering
\begin{subfigure}[t]{0.48\textwidth}
\centering
\begin{tikzpicture}

  \node[observed] (y) {$Y$};
  \node[unobserved, above=of y] (x) {$X$};
  \node[observed, left=of x] (w) {$W$};

  \node[observed,right=of x] (z) {$Z$};

  \draw[-] (z) to (y);
  \draw[-] (z) -- (x);
  \draw[-] (x) -- (y);
  \draw[-] (x) -- (w);

\end{tikzpicture}
\caption{In \emph{Simulation 1a}, classifications $W$ are conditionally independent of $Y$ so a model using $W$ as a proxy for $X$ has non-differential error. \label{fig:simulation.1a}}
\end{subfigure}
\hfill
\begin{subfigure}[t]{0.48\textwidth}
\centering
\begin{tikzpicture}

  \node[observed] (y) {$Y$};
  \node[unobserved, above=of y] (x) {$X$};
  \node[observed, left=of x] (w) {$W$};

  \node[observed,right=of x] (z) {$Z$};

  \draw[-] (z) to (y);
  \draw[-] (z) -- (x);
  \draw[-] (x) -- (y);
  \draw[-] (x) -- (w);

  \draw[-] (x) to (y);

  \draw[-] (w) -- (y);

\end{tikzpicture}
\caption{In \emph{Simulation 1b}, the edge from $W$ to $Y$ signifies that the automatic classifications $W$ are not conditionally independent of $Y$ given $X$, indicating differential error.
\label{fig:simulation.1b}
}
\end{subfigure}
\\
\hfill
\begin{subfigure}[t]{0.48\textwidth}
\centering
\begin{tikzpicture}
  \node[unobserved] (y) {$Y$};

  \node[observed, above=of y] (x) {$X$};
  \node[observed, right=of y] (w) {$W$};

  \node[observed,right=of x] (z) {$Z$};
  \draw[-] (z) to (y);
  \draw[-] (x) -- (y);
  \draw[-] (y) -- (w);
  \draw[-] (x) -- (z);

\end{tikzpicture}
\caption{In \emph{Simulation 2a}, an unbiased classifier measures the outcome. \label{fig:simulation.2a}}
\end{subfigure} \hfill
\begin{subfigure}[t]{0.48\textwidth}
\centering
\begin{tikzpicture}
  \node[unobserved] (y) {$Y$};

  \node[observed={white}{gray!40}, above=of y] (x) {$X$};
  \node[observed, right=of y] (w) {$W$};

  \node[observed,right=of x] (z) {$Z$};
  \draw[-] (x) -- (y);
  \draw[-] (z) -- (w);
  \draw[-] (y) -- (w);
  \draw[-] (x) -- (z);
  \draw[-] (z) -- (y);
\end{tikzpicture}
\caption{In \emph{Simulation 2b}, the edge connecting $W$ and $Z$ signifies that the predictions $W$ are not conditionally independent of $Z$ given $Y$, indicating systematic misclassification. \label{fig:simulation.2b}}
\end{subfigure}
\vspace{1em}
\begin{subfigure}[t]{0.2\textwidth}
\centering
\begin{tikzpicture}
  \matrix [draw, below, font=\small, align=center, column sep=2\pgflinewidth, inner sep=0.4em, outer sep=0em, nodes={align=center, anchor=center}] at (current bounding box.south){
    \node[observed,label=right:observed] {}; \\
    \node[unobserved,label=right:automatically classified]{}; \\
  };
\end{tikzpicture}
\end{subfigure}
\caption{
Bayesnet networks representing the conditional independence structure of our simulations. \label{bayesnets}
}
\end{figure}

\subsubsection{Misclassification in an Independent Variable (\emph{Simulations 1a} and \emph{1b})}

We first consider studies with the goal of testing hypotheses about the coefficients for $X$ and $Z$ in a least squares regression.
In this simulated example, $Y$ is continuous DV, $X$ is a balanced ($P(X)=0.5$) binary IV measured with an AC, $Z$ is a normally distributed IV with mean 0 and standard deviation 0.5 measured without error, and $X$ and $Z$ are correlated (Pearson's $\rho=0.24$).
To represent a study design where an AC is needed to obtain sufficient statistical power, $Z$ and $X$ can explain only 10\% of the variance in $Y$. 

In \emph{Simulation 1a} (Figure \ref{fig:simulation.1a}), we simulate an AC with 72\% accuracy.\footnote{Classifier accuracy varies between our simulations because it is difficult to jointly specify classifier accuracy and the required correlations among variables and due to random variation between simulation runs. We report the median accuracy over simulation runs.}  This reflects a situation where $X$ may be difficult to predict, but the AC, represented as a logistic regression model having linear predictor $W^*$, provides a useful signal. 
We simulate nondifferential misclassification because $W=X+\xi$, $\xi$ is normally distributed with mean $0$, and $\xi$ and $W$ are conditionally independent of $Y$ given $X$ and $Z$.

In our real-data example, we included an example where the Perspective API disproportionately misclassified comments as toxic if they disclosed aspects of identities which resulted in differential misclassification. 
In \emph{Simulation 1b} (Figure \ref{fig:simulation.1b}), we test how error correction methods can handle such differential error by making AC predictions similarly depend on the DV $Y$.
This simulated AC has $74\%$ accuracy and makes predictions $W$ that are negatively correlated with the residuals of the linear regression of $X$ and $Z$ on $Y$ (Pearson's $\rho=-0.17$). As a result, this AC makes fewer false-positives and more false-negatives at greater levels of $Y$. 


\subsubsection{Measurement Error in a Dependent Variable (\emph{Simulation 2a} and \emph{2b})}

We then simulate using an AC to measure the DV $Y$ which we aim to explain given a binary IV $X$ and a continuous IV $Z$. The goal is to estimate coefficients for $X$ and $Z$ in a logistic regression model. 
In \emph{Simulation 2a} (see Figure \ref{fig:simulation.2a}) and \emph{Simulation 2b} (see Figure \ref{fig:simulation.2b}) $X$ and $Z$ are, again, balanced ($P(X)=P(Z)=0.5$) and correlated
 (Pearson's $\rho=0.24$).  

In \emph{Simulation 1}, we chose the variance of the normally distributed outcome given our chosen coefficients $B_X$ and $B_Z$, but this is not appropriate for \emph{Simulation 2}'s logistic regression. We therefore choose, somewhat arbitrarily, $B_X=0.7$ and $B_Z=-0.7$. We again simulate ACs with moderate predictive performance.
The AC in \emph{Simulation 2a} is 72\% accurate and the AC in \emph{Simulation 2b} is 73\% accurate. In \emph{Simulation 2a}, the misclassifications are nonsystematic as $\xi$ has mean $0$ and is independent of $X$ and $Z$.  However, in \emph{Simulation 2b}  the misclassifications $\xi$ are systematic and correlated with $X$ (Pearson's $\rho = 0.1$).

\section{Simulation Results}

For each method, we visualize the consistency and efficiency of estimates across prototypical scenarios by using quantile dot plots with 95\% coverage intervals \citep{fernandes_uncertainty_2018}. We report our evaluation of uncertainty quantification in the form of empirical coverage probability in appendix \ref{appendix:main.sim.plots} to save space. 
For example, Figure \ref{fig:sim1a.x} visualizes results for \emph{Simulation 1a}. Each subplot shows a simulation with a given total sample size (No. observations) and a given sample of manually annotated observations (No. annotations). 
To assess a method's consistency, we locate the expected value of the point estimate across simulations with the center of the black circle and observe if the estimates become more precise as sample sizes increase. As an example, see the leftmost column in the bottom-left subplot of Figure \ref{fig:sim1a.x}. For the naïve estimator, the circle is far below the dashed line indicating the true value of $B_X$. Here, ignoring misclassification causes bias toward 0 and the estimator is inconsistent. To assess a method's efficiency, we mark the region in which point estimate falls in 95\% of the simulations with black lines.
The black lines in the bottom-left subplot of Figure \ref{fig:sim1a.x}, for example show that the feasible estimator, which uses only manually annotated data, is consistent but less precise than estimates from error correction methods.

\subsection{\emph{Simulation 1a:} Nonsystematic Misclassification of an Independent Variable}

\begin{figure}[htbp!]
\begin{knitrout}
\definecolor{shadecolor}{rgb}{0.969, 0.969, 0.969}\color{fgcolor}
\includegraphics[width=\maxwidth]{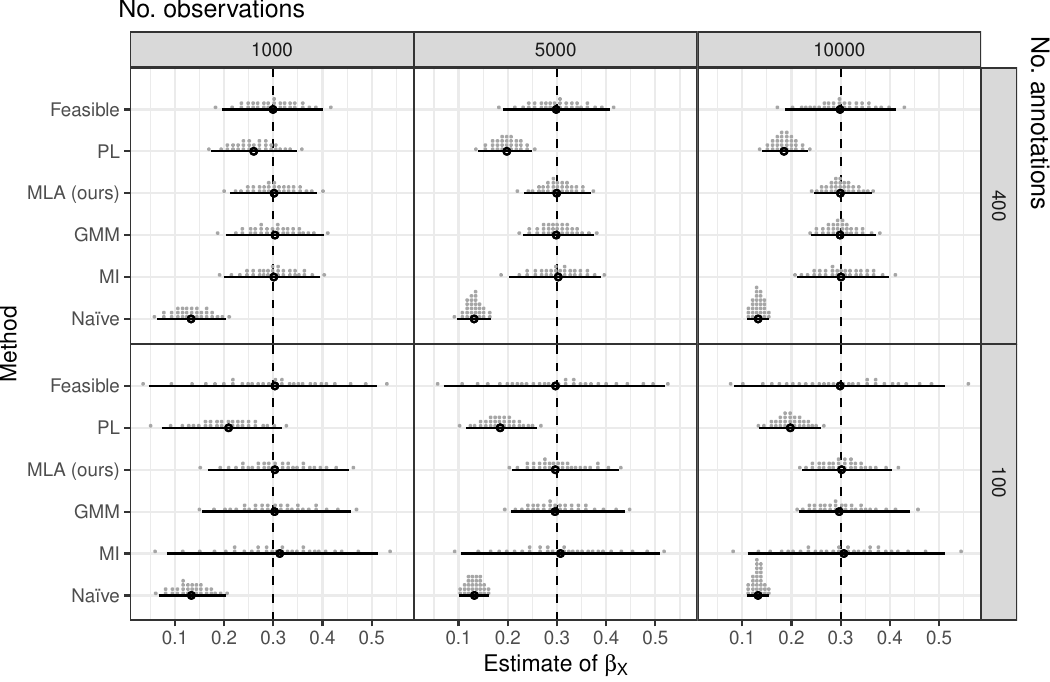} 
\end{knitrout}
\caption{Simulation 1a: Nonsystematic misclassification of an independent variable. Error correction methods, except for PL, obtain precise and accurate estimates given sufficient manually annotated data. \label{fig:sim1a.x}}
\end{figure}

Figure \ref{fig:sim1a.x} illustrates \emph{Simulation 1a}. Here, the naïve estimator is severely biased in its estimation of $B_X$.
Fortunately, error correction methods (GMM, MI, MLA) produce consistent estimates and acceptably accurate confidence intervals.
Notably, the PL method is inconsistent, and considerable bias remains when the sample of annotations is much smaller than the entire dataset.  This is likely due to $P(X=x)$ missing from the PL estimation.\footnote{Compare Equation \ref{eq:mle.covariate.chainrule.4} in Appendix \ref{appendix:derivation} to Equations 24-28 from \citet{zhang_how_2021}.} Figure
\ref{fig:sim1a.x} also shows that MLA and GMM estimates become more precise in larger datasets.
As \citet{fong_machine_2021} also observed, this precision improvement is less pronounced for MI estimates, indicating that
GMM and MLA use automated classifications more efficiently than MI.

In brief, when misclassifications cause nondifferential error, MLA and GMM are effective, efficient, and provide accurate uncertainty quantification.  They complement each other due to different assumptions: MLA depends on correctly specifying the likelihood but its robustness to incorrect specifications is difficult to analyze \citep{carroll_measurement_2006}. The GMM approach depends on the exclusion restriction instead of distributional assumptions \citep{fong_machine_2021}. In cases similar to \emph{Simulation 1a}, we therefore recommend both GMM and MLA to correct for misclassification.  Appendix \ref{appendix:robustness.vi} further explores the trade-off between the two approaches.

\subsection{\emph{Simulation 1b:} Systematic Misclassification of an Independent Variable}

\begin{figure}[htbp!]
\begin{knitrout}
\definecolor{shadecolor}{rgb}{0.969, 0.969, 0.969}\color{fgcolor}
\includegraphics[width=\maxwidth]{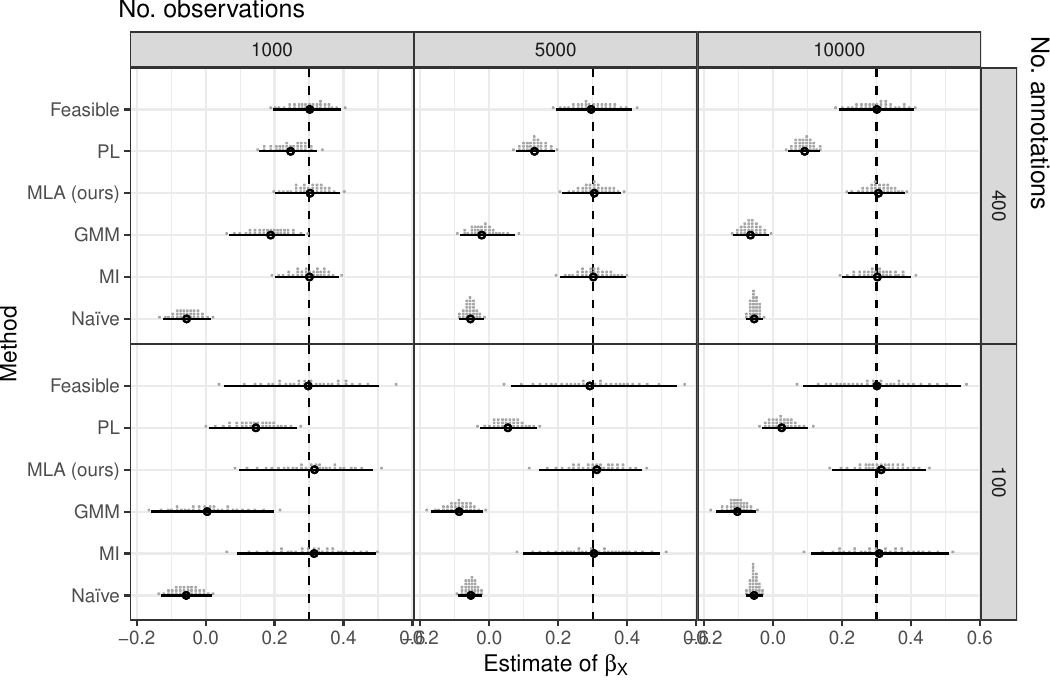} 
\end{knitrout}
\caption{\emph{Simulation 1b:} Systematic misclassification of an independent variable. Only the MLA and MI approaches obtain consistent estimates of $B_X$. \label{fig:sim1b.x}}
\end{figure}

Figure \ref{fig:sim1b.x} illustrates \emph{Simulation 1b}. Here, systematic misclassification gives rise to differential error and creates extreme misclassification bias that is more difficult to correct.
As Figure \ref{fig:sim1b.x} shows, the naïve estimator is opposite in sign to the true parameter. 
Of the four methods we test, only the MLA and the MI approach provide consistent estimates. This is expected because they use $Y$ to adjust for misclassifications. The bottom row of Figure \ref{fig:sim1b.x} shows how the precision of the MI and MLA estimates increase with additional observations.  As in \emph{Simulation 1a}, MLA uses this data more efficiently than MI does. However, due to the low accuracy and bias of the AC, additional unlabeled data improves precision less than one might expect. Both methods provide acceptably accurate confidence intervals as shown in Figure \ref{fig:sim.1b.coverage.x}. Figure \ref{fig:sim1b.z} in Appendix \ref{appendix:main.sim.plots} shows that, as in \emph{Simulation 1a}, effective correction for misclassifications of $X$ is required to consistently estimate $B_Z$, the coefficient of $Z$ on $Y$.  Inspecting results from methods that do not correct for differential error is useful for understanding their limitations. When few annotations of $X$ are observed, GMM is nearly as bad as the naïve estimator. PL is also visibly biased. Both improve when a greater proportion of the data is labeled since they combine AC-based estimates with the feasible estimator.

In sum, our simulations suggest that the MLA approach is superior in conditions of differential error.  Although estimates by the MI approach appear consistent, the method's practicality is limited by its inefficiency.

\subsection{\emph{Simulation 2a:} Nonsystematic Misclassification of a Dependent Variable}

 \begin{figure}[htbp!]
\begin{knitrout}
\definecolor{shadecolor}{rgb}{0.969, 0.969, 0.969}\color{fgcolor}
\includegraphics[width=\maxwidth]{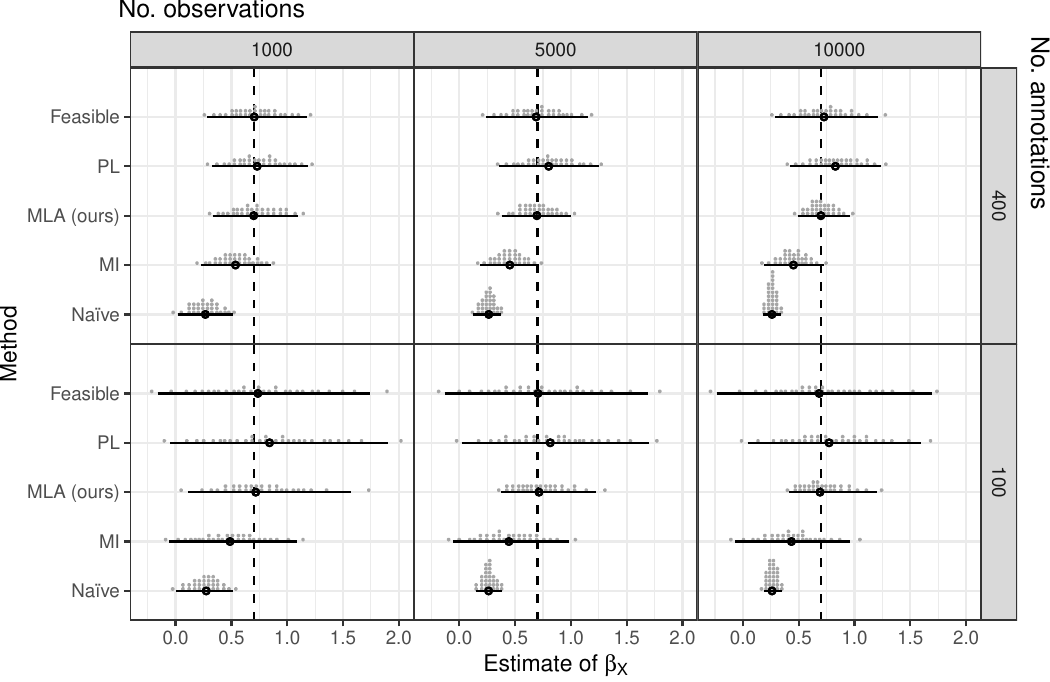} 
\end{knitrout}
\caption{Simulation 2a: Nonsystematic misclassification of a dependent variable. Only the MLA approach is consistent. \label{fig:sim2a.x}}
\end{figure}

Figure \ref{fig:sim2a.x} illustrates \emph{Simulation 2a}: nonsystematic misclassification of a dependent variable. This also introduces bias as evidenced by the naïve estimator's inaccuracy. Our MLA method is able to correct this error and provide consistent estimates.

Surprisingly, the MI estimator is inconsistent and does not improve with more human-labeled data.
The PL approach is also inconsistent, especially when only few of all observations are annotated manually. It is closer to recovering the true parameter than the MI or the naïve estimator, but provides only modest improvements in precision compared to the feasible estimator.   
It is clear that the precision of the MLA estimator improves with more observations data to a greater extent than the PL estimator.  
Figure \ref{fig:sim.2a.coverage.x} shows that when the amount of human-labled data is low, the 95\% confidence intervals of both the MLA and PL become inaccurate due to the poor finite-sample properties of the quadradic approximation for standard errors.

In brief, our simulations suggest that MLA is the best error correction method when random misclassifications affect the DV. It is the only consistent option and more efficient than the PL method, which is almost consistent.
 
\subsection{\emph{Simulation 2b}: Systematic Misclassification of a Dependent Variable}

\begin{figure}
\begin{knitrout}
\definecolor{shadecolor}{rgb}{0.969, 0.969, 0.969}\color{fgcolor}
\includegraphics[width=\maxwidth]{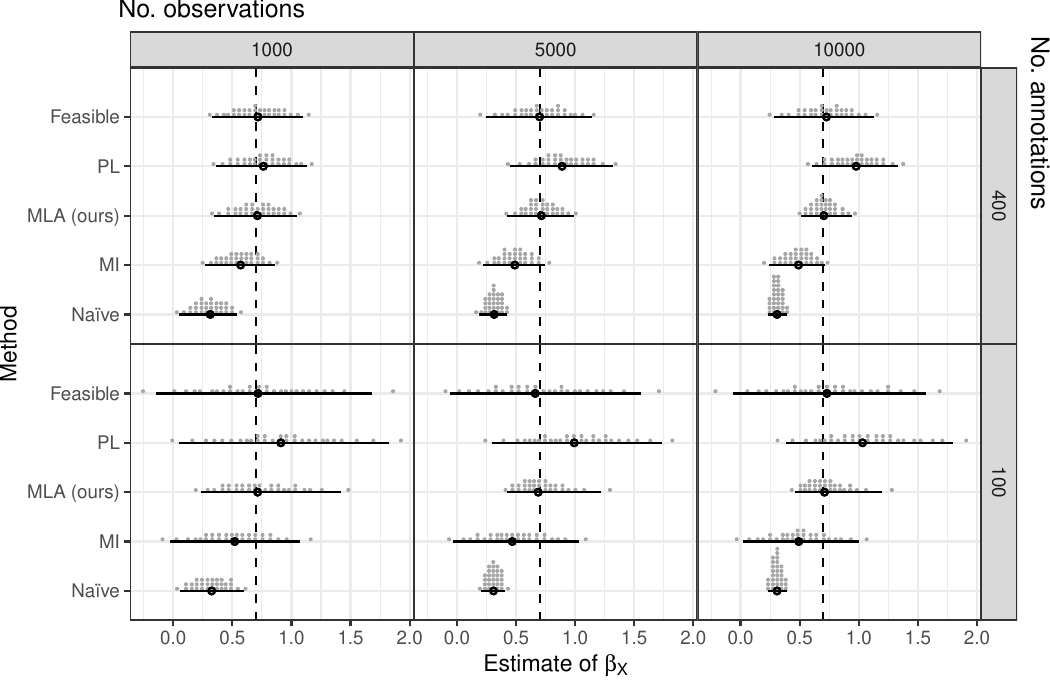} 
\end{knitrout}
\caption{Simulation 2b: Systematic misclassification of a dependent variable. Only the MLA approach is consistent. \label{fig:sim2b.x}}

\end{figure}

In \emph{Simulation 2b}, misclassifications of the dependent variable $Y$ are correlated with an independent variable $Z$. As shown in Figure \ref{fig:sim2b.x}, this causes dramatic bias in the naïve estimator. 
Similar to \emph{Simulation 2a}, MI is inconsistent. PL is also inconsistent because it does not account for $Y$ when correcting for misclassifications.
As in \emph{Simulation 1b}, our MLA method is consistent, but only does much better than the feasible estimator when the dataset is large.
Therefore, our simulations suggest that MLA is the best method when misclassifications in the DV are correlated with an IV. 
Figure \ref{fig:sim2b.z} in Appendix \ref{appendix:main.sim.plots} shows that the precision of estimates for the coefficient for $Z$ improves with additional data to a greater extent. 

\section{Transparency about Misclassification Is Not Enough—We Have To Fix It! Recommendations for Automated Content Analysis}

``Validate, Validate, Validate'' \citep[p. 269]{grimmer_text_2013} is one of the guiding mantras for automated content analysis. It reminds us that ACs can produce misleading results and of the importance of steps to ascertain validity, for instance by making misclassification transparent. 

Like \citet{grimmer_text_2013}, we are deeply concerned that computational methods may produce invalid evidence. In this sense, their validation mantra animates this paper. Critically evaluating the performance of measurement tools by providing transparency about misclassification rates via metrics such as precision or recall is the meaningful first step to build confidence on the tools, but it leaves unanswered an important question: Is comparing automated classifications to some external ground truth sufficient to claim that results are valid? Or is there something else we can do and should do? 

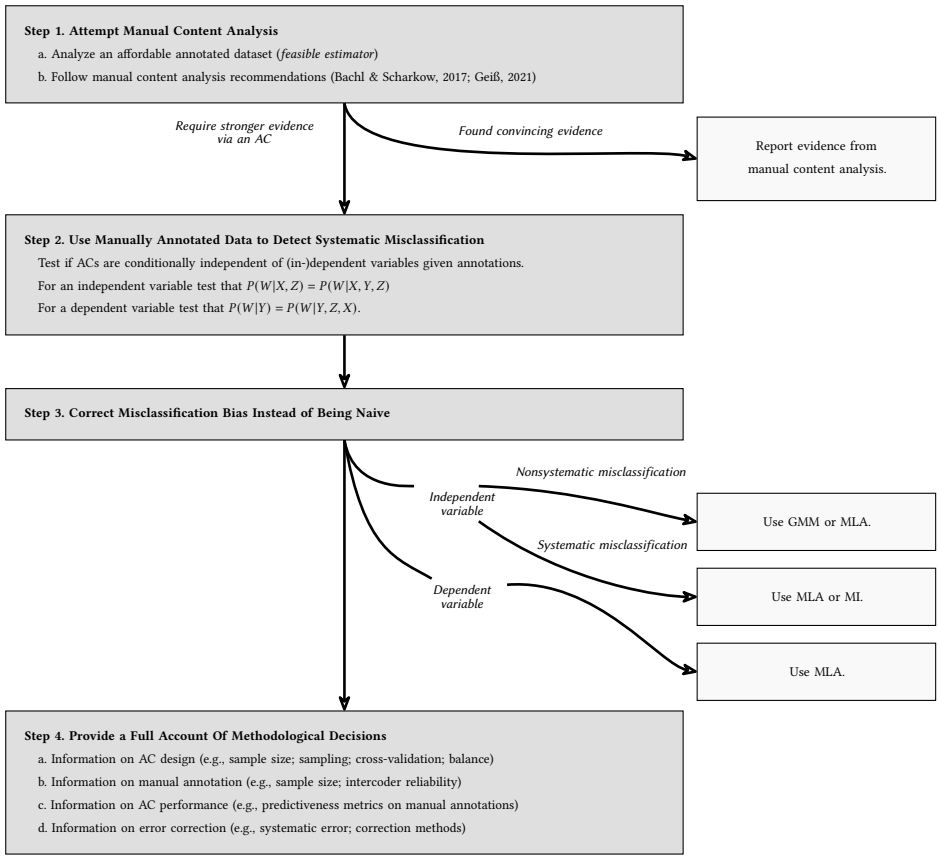
\begin{figure}[hbt!]
\centering
\tikzset{
  observed/.style={circle, draw},
  partly observed/.style 2 args={draw, fill=#2, path picture={
      \fill[#1, sharp corners] (path picture bounding box.south west) -|
      (path picture bounding box.north east) -- cycle;},
     circle},
    unobserved/.style={draw, circle, fill=gray!40},
    residual/.style={draw, rectangle},
   step box/.style={draw, rectangle, fill=gray!25, text width=6.7in, inner sep=0.2in, align=left, anchor=south west, execute at begin node=\setlength{\baselineskip}{4ex}},,
    decision box/.style={draw, diamond, fill=blue!20, text width=2in, align=center, inner sep=0.1in, anchor=south west, execute at begin node=\setlength{\baselineskip}{1ex},aspect=3},
    outcome box/.style={draw, rectangle, fill=gray!5, text width=2in, inner sep=0.25in, align=center, anchor=south west, execute at begin node=\setlength{\baselineskip}{4ex}},
        show curve controls/.style={
    postaction={
      decoration={
        show path construction,
        curveto code={
          \draw [blue] 
            (\tikzinputsegmentfirst) -- (\tikzinputsegmentsupporta)
            (\tikzinputsegmentlast) -- (\tikzinputsegmentsupportb);
          \fill [red, opacity=0.5] 
            (\tikzinputsegmentsupporta) circle [radius=.2ex]
            (\tikzinputsegmentsupportb) circle [radius=.2ex];
        }
      },
      decorate
    }},
    myarrow/.style={
    arrows={-Stealth[round, open]},
    scale=1,
    line width=1pt
    },
    myarrownotip/.style={
    scale=1,
    line width=1pt
    },
    mylabel/.style={
      text width=2.2in,
      align=center,
      execute at begin node =\setlength{\baselineskip}{1.3ex},
      inner sep=1ex,
      font={\mdseries\itshape\sffamily #1}
    }
}
\newcommand{\myindent}{\hspace{1em}}

\begin{tikzpicture}[every node/.style={transform shape}, scale=0.5]

    \node[step box] (manual) {\textbf{Step 1. Attempt Manual Content Analysis}\\ \myindent a. Analyze an affordable annotated dataset (\textit{feasible estimator}) \\\myindent b. Follow manual content analysis recommendations (Bachl \& Scharkow, 2017; Gei\ss, 2021) };

   \node[outcome box] (report_manual) [below right=0.2in of manual] {Report evidence from manual content analysis.}; 

   \node[step box] (test_systematic) [below left=0.2in of report_manual] {\textbf{Step 2. Use Manually Annotated Data to Detect Systematic Misclassification}\\ \myindent Test if ACs are conditionally independent of (in-)dependent variables given annotations. \\ \myindent For an independent variable test that $P(W|X,Z) = P(W|X,Y,Z)$ \\ \myindent For a dependent variable test that $P(W|Y) = P(W|Y,Z,X)$.};  

   \node[step box] (correct) [below=0.55in of test_systematic] {\textbf{Step 3. Correct Misclassification Bias Instead of Being Naive}};  

   \node[step box] (report) [below=2.8in of correct] {\textbf{Step 4. Provide a Full Account Of Methodological Decisions}\\ \myindent a. Information on AC design (e.g., sample size; sampling; cross-validation; balance)\\ \myindent b. Information on manual annotation (e.g., sample size; intercoder reliability)\\ \myindent c. Information on AC performance (e.g., predictiveness metrics on manual annotations) \\ \myindent d. Information on error correction (e.g., systematic error;  correction methods)};  

   \node[mylabel, anchor=south west] (independent) [below=8ex of correct,xshift=1.23in] {Independent\\ variable}; 
   \node[mylabel, anchor=south west] (dependent) [below=10ex of independent] {Dependent\\ variable};

   \node[outcome box] (outcome_systematic_iv) [below =3.8in of report_manual] {Use MLA or MI.}; 
   \node[outcome box] (outcome_nonsystematic_iv) [above =3ex of outcome_systematic_iv] {Use GMM or MLA.};


   \node[outcome box] (outcome_dv) [below =3ex of outcome_systematic_iv] {Use MLA.};


 \draw[myarrow] (manual.south) to [controls=+(280:2) and +(165:1.5)] (report_manual.west) { node [mylabel, pos=0.5, yshift=-5ex, xshift=5.5in] {Found convincing evidence}}; 

 \draw[myarrow] (test_systematic) to (correct);
 \draw[myarrow] (manual.south) to (test_systematic) {node [mylabel, pos=0.5, yshift=-5ex,xshift=2.5in] {Require stronger evidence\\ via an AC}}; 

 \draw[myarrownotip] (correct.south) to [controls=+(280:1.4) and  +(160:1.4)] (independent);

 \draw[myarrow] (independent.north)+(3ex,0) to [xshift=17ex, yshift=-8.5ex, controls=+(80:1.2) and +(170:1.2)] (outcome_nonsystematic_iv.west) {node [mylabel] {Nonsystematic misclassification}};

 \draw[myarrow] (independent.south)+(3ex,0) to [xshift=15ex, yshift=-8.8ex, controls=+(290:0.4) and +(180:0.8)] (outcome_systematic_iv.west) {node [mylabel] { Systematic misclassification}}; 

 \draw[myarrownotip] (correct.south) to [controls=+(280:3) and +(150:2)] (dependent);

 \draw[myarrow] (dependent)+(8ex,2ex) to [xshift=15ex,yshift=-4ex, controls=+(80:0.8) and +(170:1)] (outcome_dv.west);


 \draw[myarrow] (correct) to (report); 


\end{tikzpicture}
\vspace{1ex}
     \caption{Recommendations for Automated Content Analysis Study Design}
     \label{fig:FigureRecommendations}
\end{figure}

We think there is: Using statistical methods to not only quantify but also correct for misclassification. Our study provides several recommendations for using automated classifiers for statistical inference in confirmatory data analysis as summarized in Figure \ref{fig:FigureRecommendations}.

\subsection{Step 1: Attempt Manual Content Analysis}

Manual content annotation is often done \textit{post facto},  for instance to calculate predictiveness of an already existing AC such as Google's Perspective classifier. We propose to instead use manually annotated data \textit{ante facto}, i.e., before building or validating an AC.
Practically speaking, the main reason to use an AC is feasibility: to avoid the costs of manually coding a large dataset. 
One may for example need a large dataset to study an effect one assumes to be small. Manually labeling such a dataset is expensive. 
Often, ACs are seen as a cost-saving procedure but scholars often fail to consider the threats to validity posed by misclassification. 
Moreover, validating an existing AC or building a new AC is also expensive, for instance due to costs of computational resources or manual annotation of (perhaps smaller) test and training datasets.

We therefore caution researchers against preferring automated over manual content analysis unless doing so is necessary to obtain useful evidence. We agree with \citet{baden_three_2022} who argue that ``social science researchers may be well-advised to eschew the promises of computational tools and invest instead into carefully researcher-controlled, limited-scale manual studies'' (p. 11). In particular, we recommend using manually annotated data \textit{ante facto}: Researchers should begin by examining human-annotated data so to discern if an AC is necessary. In our simulations, the feasible estimator is less precise but consistent in all cases. So if fortune shines and this estimate sufficiently answers one's research question, manual coding is sufficient. Here, scholars should rely on existing recommendations for descriptive and inferential statistics when using manual content analysis \citep{geis_statistical_2021, bachl_correcting_2017}. If the feasible estimator however fails to provide convincing evidence, for example by not rejecting the null, manually annotated data is not wasted. It can be reused to build an AC or to correct misclassification bias.

\subsection{Step 2: Use Manually Annotated Data to Detect Systematic Misclassification}

As demonstrated in our simulations, knowing whether an AC makes systematic misclassifications is important: It determines which correction methods can work. 
Fortunately, manually annotated data can be used to detect systematic misclassification. 
For example, \citet{fong_machine_2021} suggest using Sargan's J-test of the null hypothesis that the product of the AC's predictions and regression residuals have an expected value of 0.
More generally, one can test if the data's conditional independence structures can be represented by Figures \ref{fig:simulation.1a} or \ref{fig:simulation.2a}. This can be done, for example, via likelihood ratio tests of $P(W|X,Z) = P(W|X,Y,Z)$ (if an AC measures an IV $X$) or of $P(W|Y) = P(W|Y,Z,X)$ (if an AC measures a DV $Y$) or by visual inspection of plots of relating misclassifications to other variables  \citep{carroll_measurement_2006}.
 We strongly recommend using such methods to test for systematic misclassification and to design an appropriate correction.

\subsection{Step 3: Correct for Misclassification Bias Instead of Being Naïve}

Across our simulations, we showed that the naïve estimator is biased. Testing different error correction methods, we found that these generate different levels of consistency, efficiency, and accuracy in uncertainty quantification. That said, our proposed MLA method should be considered as a versatile method because it is the only method capable of producing consistent estimates in prototypical situations studied here. We recommend the MLA method as the first ``go-to'' method.  As shown in Appendix \ref{appendix:robustness}, this method requires specifying a valid error model to obtain consistent estimates. One should take care that the model not have omitted variables including nonlinearities and interactions.
Our \texttt{misclassificationmodels} R package provides reasonable default error models and a user-friendly interface to facilitate adoption of our MLA method (see Appendix \ref{appendix:misclassificationmodels}).

When feasible, we recommend comparing the MLA approach to another error correction method. Consistency between two correction methods shows that results are robust independent of the correction method. If the AC is used to predict an IV, GMM is a good choice if error is nondifferential. Otherwise, MI can be considered.
Unfortunately, if the AC is used to predict a DV, our simulations do not support a strong suggestion for a second method. 
PL might be useful reasonable choice with enough manually annotated data and non-differential error.
This range of viable choices in error correction methods also  motivates our next recommendation.

\subsubsection{Step 4: Provide a Full Account of  Methodological Decisions}

Finally, we add our voices to those 
recommending that researchers report methodological decisions so other can understand and replicate their design \citep{pipal_if_2022, reiss_reporting_2022}, especially in the context of machine learning \citep{mitchell_model_2019}. These decisions include but are not limited to choices concerning test and training data (e.g., size, sampling, split in cross-validation procedures, balance), manual annotations (size, number of annotators, intercoder values, size of data annotated for intercoder testing), and the classifier itself (choice of algorithm or ensemble, different accuracy metrics). They extend to reporting error correction methods and the choices involved as proposed by our third recommendation.
In our review, we found that reporting such decisions is not yet common, at least in the context of SML-based text classification. 
When correcting for misclassification, uncorrected results will often provide a lower-bound on effect sizes; corrected analyses will provide more accurate but less conservative results. 
Therefore, both corrected and uncorrected estimates should be presented as part of making potential multiverses of findings transparent.

\section{Conclusion, Limitations, and Future Directions}

Misclassification bias is an important threat to validity in studies that use automated content analysis to measure variables.
As we showed in an example with data from the Perspective API, widely used and very accurate automated classifiers can cause type-I and type-II errors.
As evidence by our literature review, this problem has neither attracted enough attention within communication science \citep[but see][]{bachl_correcting_2017} nor in the broader computational social science community. 
Although reporting metrics of classifier performance on manually annotated validation data, for instance metrics like precision or recall, is important, such practices provide little protection from misclassification bias.
They use annotations to enact a transparency ritual to ward against misclassification bias, but annotations can do much more. With the right statistical model, they can correct misclassification bias.

We introduce \emph{Maximum Likelihood Adjustment} (MLA), a new method we designed to correct misclassification bias. We then use Monte Carlo simulations to 
evaluate the MLA approach in comparison to other error correction methods.  
Our MLA method is the only one that is effective across the full range of scenarios we tested. It is also straightforward to use. Our implementation in the R package \texttt{misclassificationmodels} provides a familiar formula interface for regression models.
Remarkably, our simulations show that our method can use even an automated classifier below common accuracy standards to obtain consistent estimates.  Therefore, low accuracy is not necessarily a barrier to using an AC.

Based on these results, we provide four recommendations for the future of automated content analysis: Researchers should (1) attempt manual content analysis before building or validating ACs to see whether human-labeled data is sufficient, (2) use manually annotated data to test for systematic misclassification and choose appropriate error correction methods, (3) correct for misclassification via error correction methods, and (4) be transparent about the methodological decisions involved in AC-based classifications and error correction. 

Our study has several limitations. First, the simulations and methods we introduce focus on misclassification by automated tools. They provisionally assume that human annotators do not make errors, especially not systematic ones. 
This assumption can be reasonable if intercoder reliability is very high but, as with ACs, this may not always be the case.

 Thus, it may be important to account for measurement error by human coders  \citep{bachl_correcting_2017} and by automated classifiers simultaneously. In theory, it is possible to extend our MLA approach in order to do so \citep{carroll_measurement_2006}. 
However, because the true values of content categories are never observed, accounting for automated and human misclassification at once requires latent variable methods that bear considerable additional complexity and assumptions \citep{pepe_insights_2007}. We leave the integration of such methods into our MLA framework for future work. In addition, our method requires an additional assumption that the error model is correct. As we argue in Appendix \ref{appendix:robustness} (section \ref{appendix:assumption}), this assumption is often acceptable.

Second, the simulations we present do not consider all possible factors that may influence the performance and robustness of error correction methods. We extend our simulations to investigate factors including classifier accuracy, heteroskedasticity, and violations of distributional assumptions in Appendix \ref{appendix:robustness}.

Third, the methods we test and the simulations we present can provide little guidance about how much validation data is required for a given level of statistical precision.  The question of how a researcher can optimally allocate a budget between costly validation data and cheap but error-prone data has been studied in the statistical literature on measurement error \citep[e.g.,][]{amorim_two-phase_2021}. Future work should develop and apply such methods for automated content analysis. 

More generally, we have shown that misclassification bias can result in misleading real-world findings, that it is too often unconsidered and unaddressed, and that statistical methods can be effective at correcting it. Yet, we wish to know how consequential misclassification bias is to real-world empirical studies so as to give more precise recommendations about when error correction methods are necessary. Future work that replicates past studies employing automated content analysis and extends them with error correction methods can show how important error correction methods are to a range of real-world study designs.

\section{Acknowledgements}
We thank the Computational Methods Hackathon pre-conference at the 2022 annual meeting of the International Communication Association (ICA). Dr. TeBlunthuis presented a workshop on misclassification bias at this hackathon and connected with Dr. Chan, who was already working on the same problem, and with Dr. Hase to begin our collaboration.  We also thank our colleagues who provided feedback on this manuscript at various stages including: Benjamin Mako Hill, Nick Vincent, Aaron Shaw, Ruijia Cheng, Kaylea Champion, and Jeremy Foote with the Community Data Science Collective as well as Abigail Jacobs, Manuel Horta Ribero, Nicolai Berk, and Marco Bachl. We also thank the anonymous reviewers with the computational methods division at ICA 2023 whose comments helped improve this manuscript and to the division's awards committee for the recognition. We similarly thank the anonymous reviewers at Communication Methods \& Measures for the invaluable comments. Any remaining errors are our own. Additional thanks to those who shared data with Dr. TeBlunthuis during the project's conception: Jin Woo Kim, Sandra Gonzalez-Bailon, and Manlio De Domenico. This work was facilitated through the use of advanced computational, storage, and networking infrastructure provided by the Hyak supercomputer system at the University of Washington.

\bibliographystyle{ACM-Reference-Format}
\bibliography{article.bib}

\clearpage
\appendix
\setcounter{page}{1}
\setcounter{secnumdepth}{3}
\addcontentsline{toc}{section}{Appendices}
\stepcounter{section}
\section{Perspective API Example}
\label{appendix:perspective}

Our example relies on the publicly available Civil Comments dataset \citep{cjadams_jigsaw_2019}. The dataset contains around 2 million comments collected from independent English-language news sites between 2015 and 2017. We use a subset of 448,000 comments which were manually annotated both for toxicity (\emph{toxicity}) and disclosure of identity (\emph{disclosure}) in a comment. The dataset also includes counts of likes each comment received (\emph{number of likes}).

Each comment was labeled by up to ten manual annotators (although selected comments were labeled by even more annotators). Originally, the dataset represents \emph{toxicity} and \emph{disclosure} as proportions of annotators who labeled a comment as toxic or as disclosing aspects of personal identity including race and ethnicity.  
For our analysis, we converted these proportions into indicators of the majority view to transform both variables to a binary scale. 

Our MLA method works in this scenario, using a uniform random sample of 10000 annotations, as shown in \ref{fig:real.data.example.app} below. 

\begin{figure}[htbp!]
\centering
\begin{subfigure}{\linewidth}
\begin{knitrout}
\definecolor{shadecolor}{rgb}{0.969, 0.969, 0.969}\color{fgcolor}
\includegraphics[width=\maxwidth]{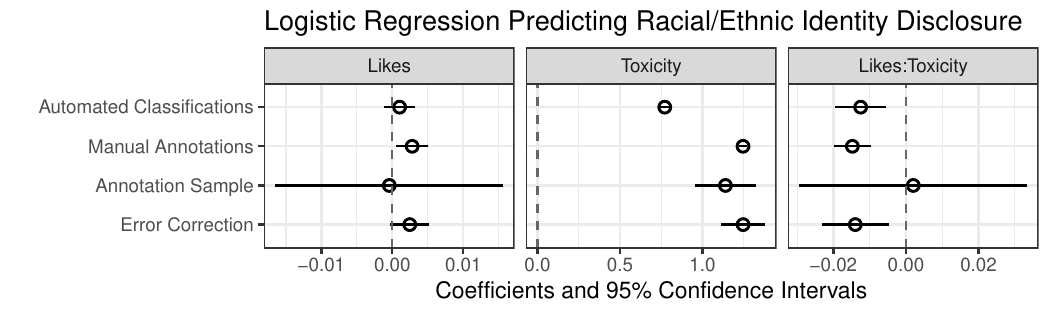} 
\end{knitrout}
\subcaption{\emph{Example 1}: Misclassification in an independent variable. \label{fig:real.data.example.iv.app}}
\end{subfigure}

\begin{subfigure}{\linewidth}
\begin{knitrout}
\definecolor{shadecolor}{rgb}{0.969, 0.969, 0.969}\color{fgcolor}
\includegraphics[width=\maxwidth]{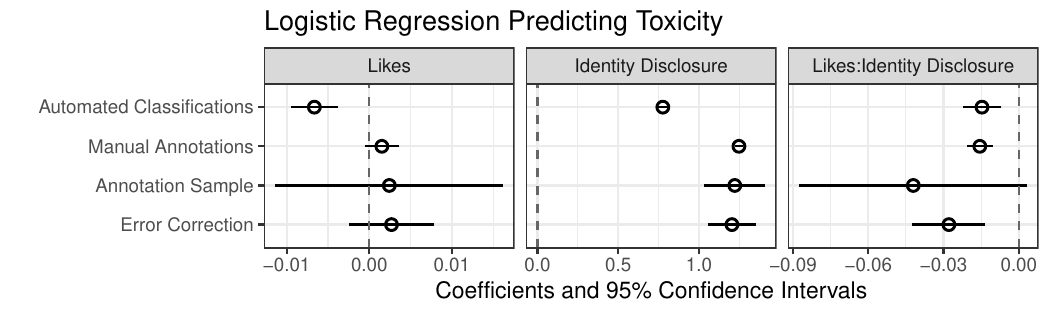} 
\end{knitrout}
\subcaption{\emph{Example 2}: Misclassification in a dependent variable. \label{fig:real.data.example.dv.app}}

\end{subfigure}
\caption{Real-data example including correction using MLA. \label{fig:real.data.example.app}}
\end{figure}

Our maximum-likelihood based error correction technique in this example requires specifying models for the Perspective's scores and, in the case where these scores are used as a covariate, a model for the human annotations.  In our first example, where toxicity was used as a covariate, we used the \emph{human annotations}, \emph{identity disclosure}, and the interaction of these two variables in the model for scores.  We did not include \emph{likes} in this model because its correlation with misclassification is low ($\rho=0.007;~ 95\%~CI=[0,0.01]$). Our model for the human annotations is an intercept-only model.

In our second example, where toxicity is the outcome, we use the fully interacted model of the \emph{human annotations}, \emph{identity disclosure}, and \emph{likes} in our model for the human annotations because all three variables are correlated with the Perspective scores.

\section{Systematic Literature Review} \label{appendix:lit.review}

To inform our simulations and recommendations, we reviewed studies using SML for text classification.

\subsection{Identification of Relevant Studies}
\begin{figure}
    \centering
    \includegraphics{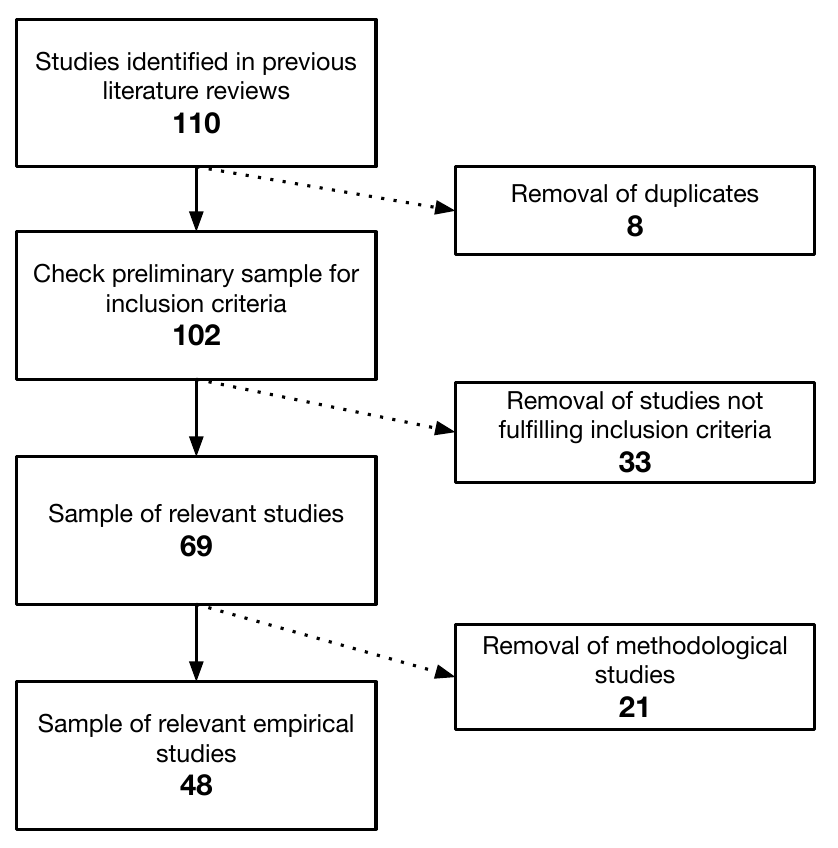}
    \caption{Identifying relevant studies for the literature review}
    \label{fig:FigureA1}
\end{figure}

Our sample was drawn from four recent reviews on the use of AC within the context of communication science and the social sciences more broadly \citep{baden_three_2022, hase_computational_2022, junger_unboxing_2022, song_validations_2020}. Authors of respective studies had either already published their data in an open-science approach or thankfully shared their data with us when contacted.
From their reviews, we collected \emph{N} = 110 studies that included some type of SML (for an overview, see Figure \ref{fig:FigureA1}). 

We first removed 8 duplicate studies identified by several reviews. Two coders then coded the remaining \emph{N} = 102 studies of our preliminary sample for relevance. After an intercoder test (\emph{N} = 10, $\alpha$ = .89), we excluded studies not fulfilling inclusion criteria, here studies not including any SML approach and studies only using SML for data cleaning, not data analysis—for instance to sort out topically irrelevant articles. Next, we removed studies focusing on methodologically advancing SML-based ACs since these studies often include far more robustness and validity tests than commonly employed in empirical settings. Subsequently, all relevant empirical studies (\emph{N} = 48) were coded in further detail. 

\subsection{Manual Coding of Relevant Empirical Studies}
For manual coding, we created a range of variables (for an overview, see Table \ref{tab:TableA1}). Based on data from the Social Sciences Citation Index (SSCI), we identified whether studies were published in journals classified as belonging to \emph{Communication} and their \emph{Impact} according to their H index. In addition, two authors manually coded...
\begin{itemize}
 \item the type of variables created via SML-based ACS using the variables \emph{Dichotomous} (0 = No, 1 = Yes), \emph{Categorical} (0 = No, 1 = Yes), \emph{Ordinal} (0 = No, 1 = Yes), and \emph{Metric} (0 = No, 1 = Yes), 
 \item whether variables were used in descriptive or multivariate analyses using the variables \emph{Descriptive} (0 = No, 1 = Yes), \emph{Independent} (0 = No, 1 = Yes), and \emph{Dependent} (0 = No, 1 = Yes), 
 \item how classifiers were trained and validated via manually annotated data using the variables \emph{Size Training Data} (Open String), \emph{Size Test Data} (Open String), \emph{Size Data Intercoder Test} (Open String), \emph{Intercoder Reliability} (Open String), and \emph{Accuracy of Classifier} (Open String),
 \item whether articles mentioned and/or corrected for misclassifications using the variables \emph{Error Mentioned} (0 = No, 1 = Yes) and \emph{Error Corrected} (0 = No, 1 = Yes).
\end{itemize}

\begin{table}
  \caption{Variables Coded for Relevant Empirical Studies}
  \label{tab:TableA1}
  \begin{tabular}{l l l l}         \toprule
  Category               & Variable                      & Krippendorf's $\alpha$  & \% or \emph{M} (\emph{SD}) \\ \midrule
  Type of Journal        & \emph{Communication}          & n.a.                    & 67\%  \\
                         & \emph{Impact}                 & n.a.                    & \emph{M = 4} \\
  Type of Variable       & \emph{Dichotomous}            & 0.86                    & 50\%  \\
                         & \emph{Categorical}            & 1                       & 23\% \\
                         & \emph{Ordinal}                & 0.85                    & 10\% \\
                         & \emph{Metric}                 & 1                       & 35\% \\
  Use of Variable        & \emph{Descriptive}            & 0.89                    & 90\% \\
                         & \emph{Independent}            & 1                       & 44\% \\
                         & \emph{Dependent}              & 1                       & 40\% \\
  Information on Classifier & \emph{Size Training Data}  & 0.95                    & 67\%  \\
                         & \emph{Size Test Data}      & 0.79                    & 52\%  \\
                         & \emph{Size Data Intercoder Test}  & 1      & 44\%  \\
                         & \emph{Intercoder Reliability}  & 0.8             & 56\%  \\
                         & \emph{Accuracy of Classifier}  & 0.77                   & 85\%  \\
  Measurement Error      & \emph{Error Mentioned}        & 1                       & 19\% \\ 
                         & \emph{Error Corrected}        & 1                       & 2\% \\ \bottomrule
  \end{tabular}
\end{table}

\subsection{Results}

SML-based ACs were most often used to create dichotomous measurements (\emph{Dichotomous}: 50\%), followed by variables on a metric (\emph{Metric}: 35\%), categorical (\emph{Categorical}: 23\%), or ordinal scale (\emph{Ordinal}: 10\%). Almost all studies used SML-based classifications to report descriptive statistics on created variables (\emph{Descriptive}: 90\%). However, many also used these in downstream analyses, either as dependent variables (\emph{Dependent}: 40\%) or independent variables (\emph{Independent}: 44\%) in statistical models. 

Only slightly more than half of all studies included information on the size of training or test sets (\emph{Size Training Data}: 67\%, \emph{Size Test Data}: 52\%). Even fewer included information on the size of manually annotated data for intercoder testing (\emph{Size Data Intercoder Test}: 44\%) or respective reliability values (\emph{Intercoder Reliability}: 56\%). Lastly, not all studies reported how well their classifier performed by using metrics such as precision, recall, or F1-scores (\emph{Accuracy of Classifier}: 85\%). Lastly,  few studies exlicitly mentioned the issue of misclassification (\emph{Error Mentioned}: 19\%, with only a single study correcting for such (\emph{Error Corrected}: 2\%).

\section{Other Error Correction Methods}
\label{appendix:other.methods}
Statisticans have introduce a range of other error correction methods which we did not test in our simulations. Here, we briefly discuss three additional methods and explain why we did not include them in our simulations.

\emph{Simulation extrapolation} (SIMEX) simulates the process generating measurement error to model how measurement error affects an analysis and ultimately to approximate an analysis with no measurement error \citep{carroll_measurement_2006}. SIMEX is a very powerful and general method that can be used without manually annotated data, but may be more complicated than necessary to correct measurement error from ACs when manually annotated data is available. Likelihood methods are easy to apply to misclassification so SIMEX seems unnecessary \citep{carroll_measurement_2006}.

\emph{Score function methods} derive estimating equations for models without measurement error and then solve them either exactly or using numerical integration \citep{carroll_measurement_2006, yi_handbook_2021}. 
The main advantage of score function methods may have over likelihood-based methods is that they do not require distributional assumptions about mismeasured independent variables. This advantage has limited use in the context of ACs because binary classifications must follow Bernoulli distributions.

We also do not consider \emph{Bayesian methods} (aside from the Amelia implementation of MI) because we expect these are analogous to the maximum likelihood methods we consider. Bayesian methods may have other advantages resulting from posterior inference and may generalize to a wide range of applications. However, specifying prior distributions introduces additional methodological complexity and posterior inference would be computationally intensive, making Bayesian methods less convenient for Monte Carlo simulations. 

\section{Maximum Likelihood Adjustment}
\label{appendix:mla}

This appendix provides a more rigorous explanation of our maximum likelihood adjustment method presenting first the general theory and then with how MLA is used in our simulations. 

\subsection{Deriving MLA When an AC Measures an Independent Variable}
\label{appendix:derivation}
To explain why the MLA approach is effective, we follow \citet{carroll_measurement_2006} and begin by observing the following fact from basic probability theory:

\begin{align}
    P(Y,W) &= \sum_{x}{P(Y,W,X=x)} 
    \label{eq:mle.covariate.chainrule.1}\\
    &= \sum_{x}{P(Y|W,X=x)P(W,X=x)} 
    \label{eq:mle.covariate.chainrule.2}\\
    &= \sum_{x}{P(Y,X=x)P(W|Y,X=x)}  \label{eq:mle.covariate.chainrule.3} \\
    &= \sum_{x}{P(Y|X=x)P(W|Y,X=x)P(X=x)} \label{eq:mle.covariate.chainrule.4}
\end{align}
\noindent 
Equation \ref{eq:mle.covariate.chainrule.1} integrates $X$ out of the joint probability of $Y$ and $W$ by summing over its possible values $x$. If $X$ is binary, this means adding the probability given $x=1$ to the probability given $x=0$.  When $X$ is observed, say $x=0$, then $P(X=0)=1$ and $P(X=1)=0$. As a result, only the true value of $X$ contributes to the likelihood. However, when $X$ is unobserved, all of its possible values contribute. In this way, integrating out $X$ allows us to include data where $X$ is not observed in the likelihood. 

Equation \ref{eq:mle.covariate.chainrule.2} uses the chain rule of probability to factor the joint probability $P(Y,W)$ of $Y$ and $W$ into $P(Y|W,X=x)$, the conditional probability of $Y$ given $W$ and $X$, and $P(W,X=x)$, the joint probability of $W$ and $X$. This lets us see how maximizing $\mathcal{L}_{\Theta}(\Theta;y,w)$, the joint likelihood of parameters $\Theta$ given observed data $y$ and $w$, accounts for the uncertainty of automated classifications. For each possible value $x$ of $X$, it weights the model of the outcome $Y$ by the probability that $x$ is the true value and that the AC output $w$.

Equation \ref{eq:mle.covariate.chainrule.3} shows a different way to factor the joint probability $P(Y,W)$ so that $W$ is not in the model of $Y$. Since $X$ and $W$ are correlated, if $W$ is in the model for $Y$, the estimation of $B_1$ will be biased.  By including $Y$ in the model for $W$, Equation \ref{eq:mle.covariate.chainrule.3} can account for differential measurement error. 

Equation \ref{eq:mle.covariate.chainrule.4} factors $P(Y,X=x)$ the joint probability of $Y$ and $X$ into $P(Y|X=x)$, the conditional probability of $Y$ given $X$, $P(W|X=x,Y)$, the conditional probability of $W$ given $X$ and $Y$, and $P(X=x)$ the probability of $X$.  This shows that fitting a model for $Y$ given $X$ in this framework, such as the regression model $Y = B_0 + B_1 X + B_2 Z$ requires including the \emph{exposure model} for $P(X=x)$.  Without validation data, $P(X=x)$ is difficult to calculate without strong assumptions \citep{carroll_measurement_2006}, but $P(X=x)$ can easily be estimated using a sample of validation data.

Equations \ref{eq:mle.covariate.chainrule.1}--\ref{eq:mle.covariate.chainrule.4} demonstrate the generality of this method because the conditional probabilities may be calculated using a wide range of probability models. 
 For simplicity, we have focused on linear regression for the probability of $Y$ and logistic regression for the probability of $W$ and the probability of $X$. However, more flexible probability models such as generalized additive models (GAMs) or Gaussian process classification may be useful for modeling nonlinear conditional probability functions \citep{williams_bayesian_1998}.

\subsection{When an AC Measures the Dependent Variable}

Again, we will maximize $\mathcal{L}_{\Theta}(\Theta;y,w)$, the joint likelihood of the parameters $\Theta$ given the observed data on the outcome $y$ and automated classifications $w$ measuring the dependent variable $Y$ \citep{carroll_measurement_2006}.
We again use the law of total probability to integrate out $Y$ and the chain rule of probability to factor the joint probability into $P(Y)$, the probability of $Y$, and $P(W|Y)$, the conditional probability of $W$ given $Y$.

\begin{align}
    P(Y,W) &= \sum_{y}{P(Y=y,W)} \\
        &= \sum_{y}{P(Y=y)P(W|Y=y)}
\end{align}

As above, the conditional probability of $W$ given $Y$ must be estimated using a model that describes the conditional dependence of $W$ on $Y$.

\subsection{Instantiating MLA in Simulation 1: When an Automated Classifier Predicts an Independent Variable}

In general, if we want to estimate a model $f_{\Theta_y}(y|x, z;\Theta_y)$ for $y$ given $x$ and $z$ with parameters $\Theta_y$, we can use AC classifications $w$ predicting $x$ to gain statistical power without introducing misclassification bias by maximizing $\mathcal{L}(\Theta|y,w)$, the likelihood of the parameters $\Theta = \{\Theta_y, \Theta_w, \Theta_x\}$ given observed data $y$ and $w$  \citep{carroll_measurement_2006}.
The joint model for $y$ and $w$ can be factored into the product of three terms: $f_{\Theta_y}(y|x,z;\Theta_y)$, the model with parameters $\Theta_y$ we want to estimate, $p(w|x,y,z; \Theta_w)$, a model for $w$ having parameters $\Theta_w$, and $p(x|z; \Theta_x)$, a model for $x$ having parameters $\Theta_x$.
Estimating these three models is sufficient to estimate a joint model for the dependent variable and automated classifications and thereby obtain a consistent estimate despite misclassification. $P(W=w|x,y; \Theta_w)$ is called the \emph{error model} and $P(X=x|z; \Theta_x)$ is called the \emph{exposure model} \citep{carroll_measurement_2006}.

To illustrate, consider the regression model  $y=B_0 + B_1 x + B_2 z + \varepsilon$  and automated classifications $w$ of the independent variable $x$.
We can assume that the probability of $w$ follows a logistic regression model of $y$, $x$, and $z$ and that the probability of $x$ follows a logistic regression model of $z$. In this case, the likelihood model below is sufficient to consistently estimate the parameters $\Theta = \{\Theta_y, \Theta_w, \Theta_x\} = \{\{B_0, B_1, B_2, \sigma\}, \{\alpha_0, \alpha_1, \alpha_2, \alpha_3\}, \{\gamma_0, \gamma_1\}\}$.
$$
\mathcal{L}(\Theta; y, w) = \prod_{i=1}^{N}{\begin{cases}
f_{\Theta_y}(y_i|x_i, z_i;\Theta_Y)p_{\Theta_w}(W=w_i|x_i, y_i, z_i; \Theta_w)p_{\Theta_x}(X=x_i|z_i; \Theta_x)~\mathrm{if}~x_i~\mathrm{is~observed} \\ 
\sum_{x}f_{\Theta_y}(y_i|x_i=x, z_i;\Theta_Y)p_{\Theta_w}(W=w_i|x_i=x, y_i, z_i; \Theta_w)p_{\Theta_x}(X=x|z_i; \Theta_x)~\mathrm{otherwise} \end{cases}}
\label{eq:covariate.reg.general}$$
\begin{align}
    f_{\Theta_y}(y_i| x_i, z_i; \Theta_y) &= \frac{1}{\sigma \sqrt{2\pi}}e^{-\frac{1}{2}(\frac{(y_i - (B_0 + B_1 x_i + B_2 z_i))}{\sigma})^2}\\
    p_{\Theta_w}(W=w_i| x_i, y_i, z_i; \Theta_w) &= \begin{cases} \frac{1}{1 + e^{-(\alpha_0 + \alpha_1 y_i + \alpha_2 x_i + \alpha_3 z_i)}}~\mathrm{if}~ w_i = 1 \\ \frac{e^{-(\alpha_0 + \alpha_1 y_i + \alpha_2 x_i + \alpha_3 z_i)}}{1 + e^{-(\alpha_0 + \alpha_1 y_i + \alpha_2 x_i + \alpha_3 z_i)}} ~\mathrm{if}~w_i = 0 \end{cases} \\
\label{eq:covariate.logisticreg.w} \\
    p_{\Theta_x}(X=x_i| z_i; \Theta_x) &= \begin{cases}\frac{1}{1 + e^{-(\gamma_0 + \gamma_1 z_i)}}~\mathrm{if}~x_i=1 \\
    \frac{e^{-(\gamma_0 + \gamma_1 z_i)}}{1 + e^{-(\gamma_0 + \gamma_1 z_i)}}~\mathrm{if}~x_i=0
    \end{cases}
\end{align}

\noindent where $f_{\Theta_y}(y_i| x_i, z_i; \Theta_y)$ is the density of the normal distribution with mean $B_0 + B_1 x_i + B_2 x_i$ and variance $\sigma^2$.  Note that Equation \ref{eq:covariate.reg.general} models differential error (i.e., $Y$ is not assumed independent of $W$ conditional on $X$ and $Z$) via a linear relationship between $w$ and $y$.  When error is nondifferential, the dependence between $w$ and $y$ can be removed from Equations \ref{eq:covariate.reg.general} and \ref{eq:covariate.logisticreg.w}. 

Estimating the three conditional probabilities in practice requires specifying models on which the validity of the method depends.
This framework is very general and a wide range of probability models, such as generalized additive models (GAMs) or Gaussian process classification, may be used to model $p_{\Theta_w}(W=w_i| x_i, y_i, z_i; \Theta_w)$ and $p_{\Theta_x}(X=x_i|z_i;\Theta_x)$ \citep{williams_bayesian_1998}.

\subsection{Instantiating MLA in Simulation 2: When an Automated Classifier Predicts a Dependent Variable}

We now turn to the case when an AC makes classifications $w$ that predict a discrete dependent variable $y$. 
In our second real-data example, $w$ is the Perspective API's toxicity classifications and $y$ is the true value of toxicity.
This case is simpler than the case above where an AC is used to measure an independent variable $x$ because there is no need to specify a model for the probability of $x$. 
If we assume that the probability of $y$ follows a logistic regression model of $x$ and $z$ and allow $w$ to be biased and to directly depend on $x$ and $z$, then maximizing the following likelihood is sufficient to consistently estimate the parameters $\Theta = \{\Theta_y, \Theta_w\} = \{\{B_0, B_1, B_2\},\{\alpha_0, \alpha_1, \alpha_2, \alpha_3\}\}$.

$$
    \mathcal{L}(\Theta;y,w) = \prod_{i=0}^{N}{\begin{cases}
    p_{\Theta_y}(Y=y_i | x_i, z_i; \Theta_y)p_{\Theta_w}(W=w_i|x_i, z_i, y_i; \Theta_w)~\mathrm{if}~y_i~\mathrm{is~observed} \\
    \sum_{y}{p_{\Theta_y}(Y=y | x_i, z_i; \Theta_y)p_{\Theta_w}(W=w_i|x_i, z_i, y_i=y; \Theta_w)}~\mathrm{otherwise}\end{cases}}
    \label{eq:depvar.general}
$$ 

\begin{align}
    p_{\Theta_y}(Y=y_i| x_i, z_i; \Theta_y) &= \begin{cases} \frac{1}{1 + e^{-(B_0 + B_1 x_i + B_2 z_i)}}~\mathrm{if}~y_i = 1 \\ 
    \frac{e^{-(B_0 + B_1 x_i + B_2 z_i)}}{1 + e^{-(B_0 + B_1 x_i + B_2 z_i)}}~\mathrm{if}~y_i=0\end{cases}\\ 
    p_{\Theta_w}(W=w_i | y_i, x_i, z_i; \Theta_w) &= \begin{cases} \frac{1}{1 + e^{-(\alpha_0 + \alpha_1 y_i + \alpha_2 x_i + \alpha_3 z_i)}}~\mathrm{if}~w_i=1 \\
    \frac{e^{-(\alpha_0 + \alpha_1 y_i + \alpha_2 x_i + \alpha_3 z_i)}}{1 + e^{-(\alpha_0 + \alpha_1 y_i + \alpha_2 x_i + \alpha_3 z_i)}}~\mathrm{if}~w_i=0
    \end{cases}
    \label{eq:depvar.w}
\end{align}

If the AC's errors are conditionally independent of $X$ and $Z$ given  $W$, the dependence of $w$ on $x$ and $z$ can be omitted from equations \ref{eq:depvar.general} and \ref{eq:depvar.w}.

We implement these methods in \texttt{R} using the \texttt{optim} library for maximum likelihood estimation.  Our implementation supports models specified using \texttt{R}'s formula syntax. It can fit linear and logistic regression models when an AC measures an independent variable and logistic regression models when an AC measures the dependent variable. Our implementation provides two methods for approximating confidence intervals: The Fischer information quadratic approximation and the profile likelihood method provided in the \texttt{R} package \texttt{bbmle}.  The Fischer approximation usually works well in simple models fit to large samples and is fast enough for practical use for the large number of simulations we present. However, the profile likelihood method may provide more accurate confidence intervals \citep{carroll_measurement_2006}.

\subsection{Comment on Model Assumptions}
\label{appendix:assumption}

How burdensome is the assumption that the error model be able to consistently estimate the conditional probability of $W$ given $Y$?  If this assumption were much more difficult than those already accepted by the model for $Y$ given $X$ and $Z$, one would fear that using the MLA correction method introduces greater validity threats than it removes. In particular, one may worry that unobserved variables $U$ are omitted from our model for $P(Y,W)$.  As demonstrated in Appendix \ref{appendix:robustness} (section \ref{appendix:misspec}), the MLA method is less effective when variables are omitted from the error model. 

However, if we believe our outcome model for $P(Y|X,Z)$ is consistent this threat is substantially reduced.  If one can assume a model for $P(Y|X,Z)$, it is often reasonable to assume the variables needed to model $P(W|X,Y,Z)$ are observed.
Furthermore, since $W$ is an output from an automated classifier it depends only on the classifier's features, which are observable in principle. As a result, and as suggested by \citet{fong_machine_2021}, one should consider including all such features in the error model.  

However, due to the highly nonlinear nature of machine learning classifiers, specifying the functional form of the error model may require care in practice.  One option is to calibrate an AC's to one's dataset and thereby obtain accurate estimates of its predicted probabilities.

\section{misclassificationmodels: The R package} \label{appendix:misclassificationmodels}

The package provides a function to conduct regression analysis but also corrects for misclassification using information from manually annotated data. The function is very similar to \textbf{glm()} but with two changes:

\begin{itemize}
\item The formula interface has been extended with the double-pipe operator to denote proxy variable. For example, \textbf{x || w} indicates that \textit{w} is the proxy of the ground truth \textit{x}.
\item The manually annotated data must be provided via the argument \textit{data2}
\end{itemize}

The following snippet shows a typical scenario, here for correcting misclassifications in an independent variable:
\lstset{style=mystyle}
\begin{lstlisting}[language=R, caption=A demo of misclassificationmodels]
library(misclassificationmodels)
## research_data contains the following columns: y, w, z
## val_data contains the following columns: y, w, x, z
# w is a proxy of x
res <- glm_fixit(formula = y ~ x || w + z, 
                 data = research_data, 
                 data2 = val_data)
summary(res)
\end{lstlisting}

For more information about the package,  please see here: \url{https://osf.io/pyqf8/?view_only=c80e7b76d94645bd9543f04c2a95a87e}.

\section{Additional plots for Simulations 1 and 2}
\label{appendix:main.sim.plots}

\begin{figure}[htbp!]
\begin{knitrout}
\definecolor{shadecolor}{rgb}{0.969, 0.969, 0.969}\color{fgcolor}
\includegraphics[width=\maxwidth]{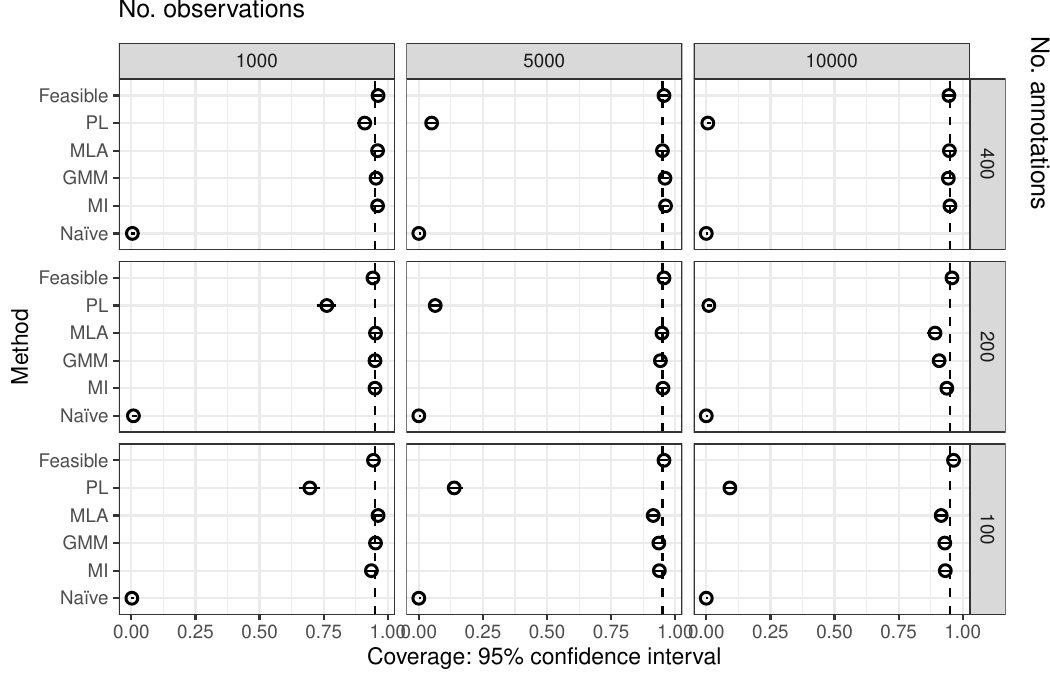} 
\end{knitrout}
\caption{Coverage of 95\% confidence intervals for $B_X$ in \emph{Simulation 1a}. MLA has imperfect coverage when the number of annotations is low as a result of the poor finite sample performance of the quadratic approximation.}
\label{fig:sim.1a.coverage.x}
\end{figure}

\begin{figure}[htbp!]
\begin{knitrout}
\definecolor{shadecolor}{rgb}{0.969, 0.969, 0.969}\color{fgcolor}
\includegraphics[width=\maxwidth]{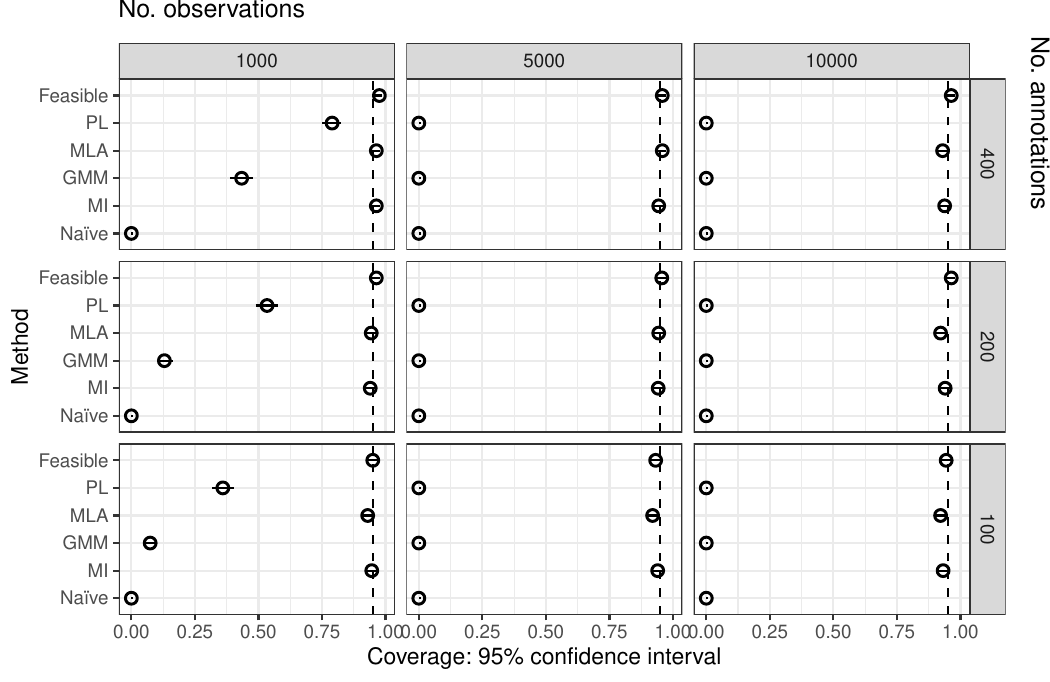} 
\end{knitrout}
\caption{Coverage of 95\% confidence intervals for $B_X$ in \emph{Simulation 1b}.}
\label{fig:sim.1b.coverage.x}
\end{figure}

\begin{figure}[htbp!]
\begin{knitrout}
\definecolor{shadecolor}{rgb}{0.969, 0.969, 0.969}\color{fgcolor}
\includegraphics[width=\maxwidth]{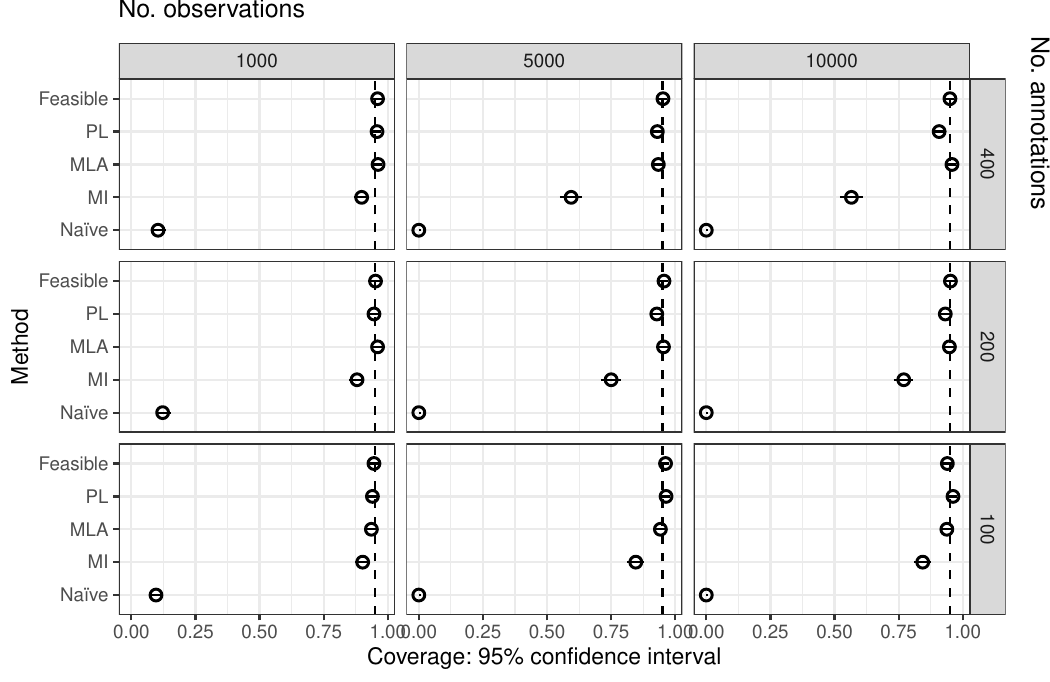} 
\end{knitrout}
\caption{Coverage of 95\% confidence intervals for $B_X$ in \emph{Simulation 2a}. PL has imperfect coverage as a result of the quadratic approximation's poor sample performance.}
\label{fig:sim.2a.coverage.x}
\end{figure}

\begin{figure}[htbp!]
\begin{knitrout}
\definecolor{shadecolor}{rgb}{0.969, 0.969, 0.969}\color{fgcolor}
\includegraphics[width=\maxwidth]{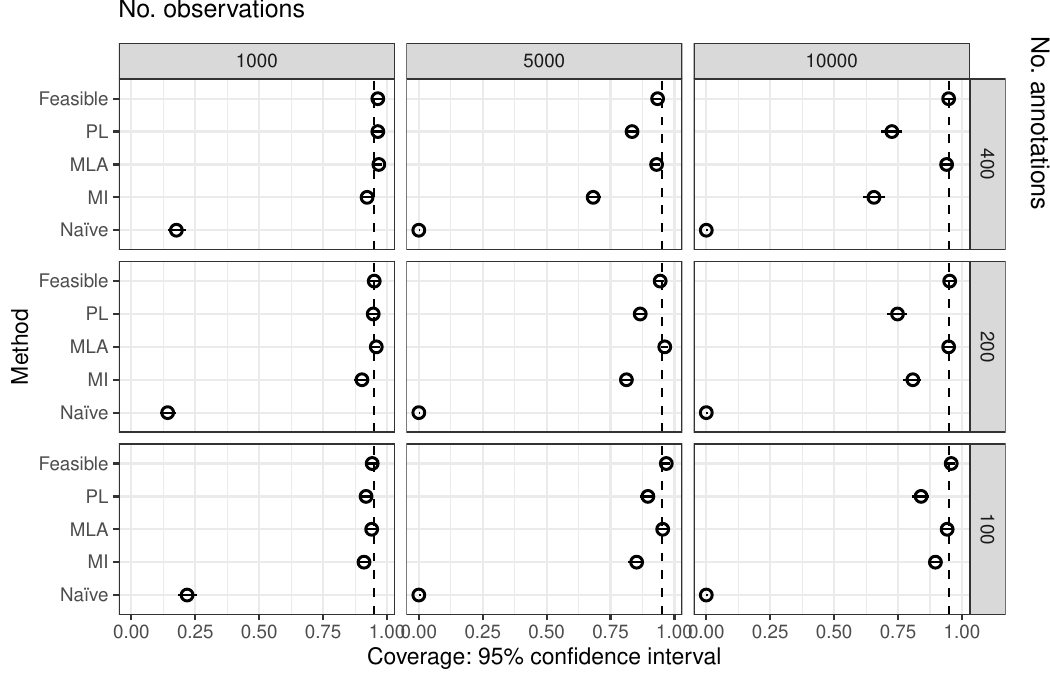} 
\end{knitrout}
\caption{Coverage of 95\% confidence intervals for $B_X$ in \emph{Simulation 2b}.}
\label{fig:sim.2b.coverage.x}
\end{figure}

\begin{figure}[htbp!]
\begin{knitrout}
\definecolor{shadecolor}{rgb}{0.969, 0.969, 0.969}\color{fgcolor}
\includegraphics[width=\maxwidth]{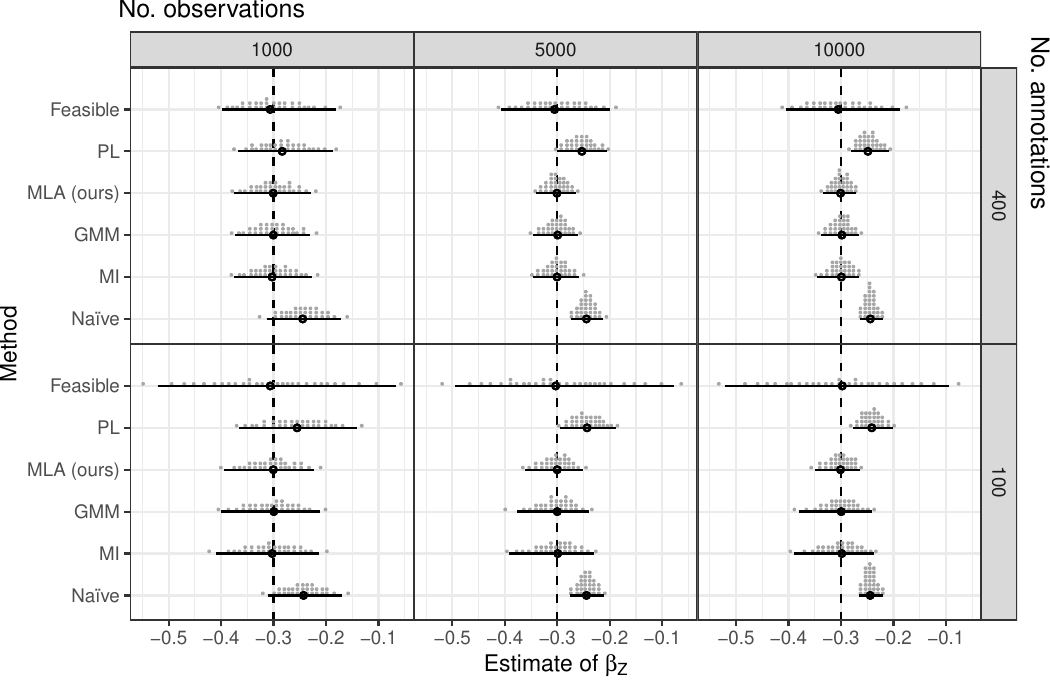} 
\end{knitrout}
\caption{Estimates of $B_Z$ in \emph{Simulation 1a}. All error correction methods except PL obtain precise and accurate estimates of of $B_Z$ given sufficient validation data.}
\label{fig:sim1a.z}
\end{figure}

\begin{figure}[htbp!]
\begin{knitrout}
\definecolor{shadecolor}{rgb}{0.969, 0.969, 0.969}\color{fgcolor}
\includegraphics[width=\maxwidth]{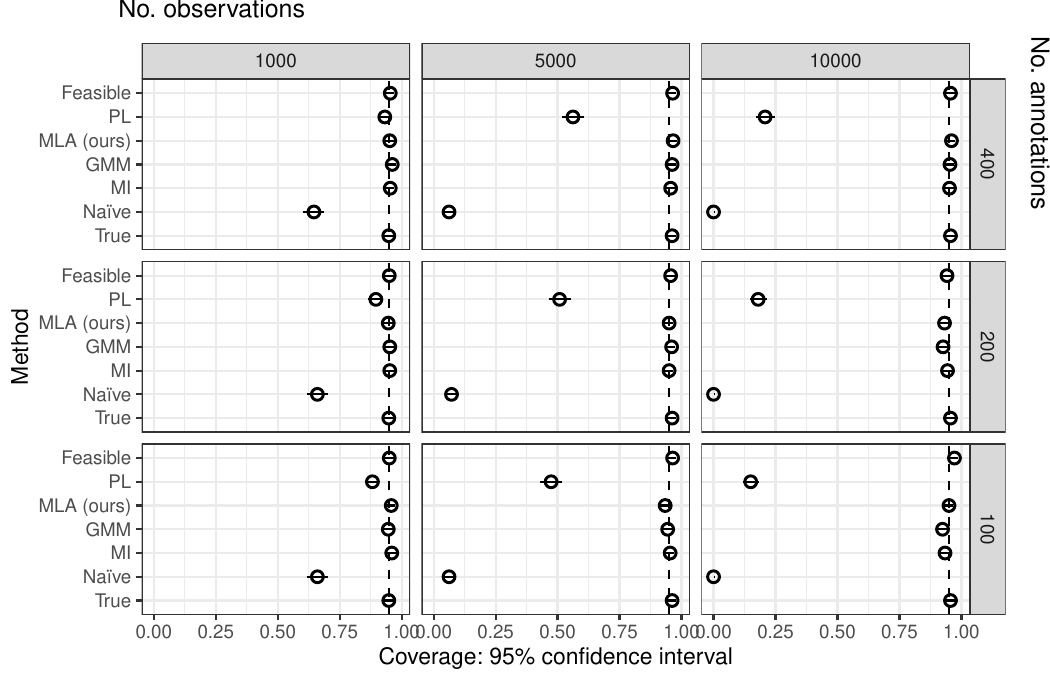} 
\end{knitrout}
\caption{Coverage of 95\% confidence intervals for $B_Z$ in \emph{Simulation 1a}.}
\label{fig:sim1a.z.coverage.z}
\end{figure}

\begin{figure}[htbp!]
\begin{knitrout}
\definecolor{shadecolor}{rgb}{0.969, 0.969, 0.969}\color{fgcolor}
\includegraphics[width=\maxwidth]{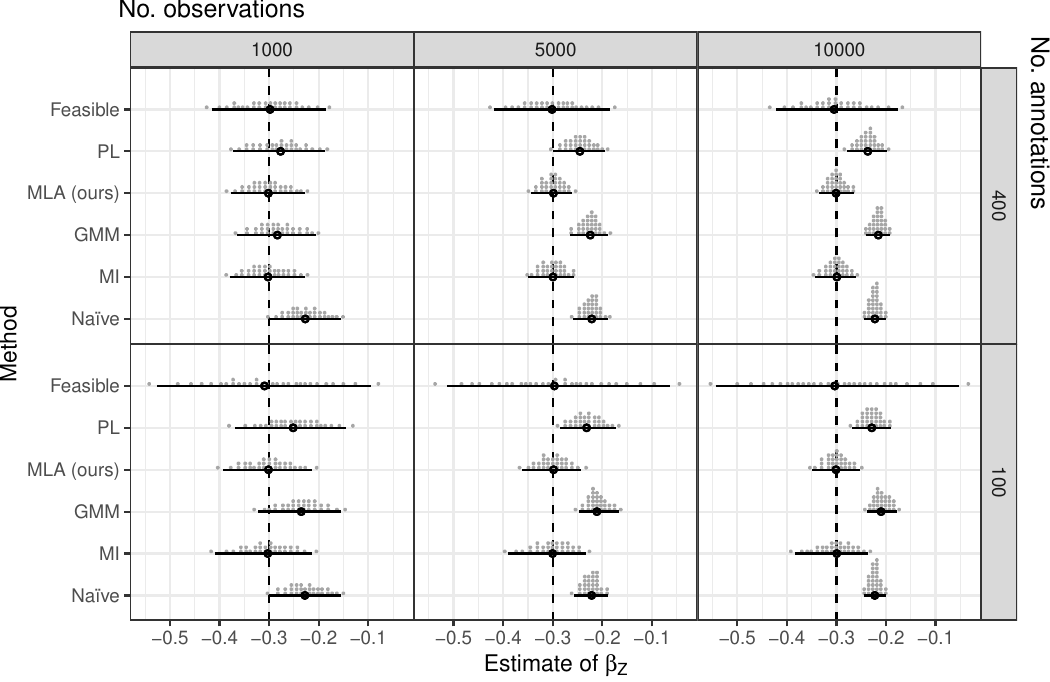} 
\end{knitrout}
\caption{Estimates of $B_Z$ in \emph{Simulation 1b}.  Only MI and our MLA approach obtain consistent estimates of $B_Z$.\label{fig:sim1b.z}}
\end{figure}

\begin{figure}[htbp!]
\begin{knitrout}
\definecolor{shadecolor}{rgb}{0.969, 0.969, 0.969}\color{fgcolor}
\includegraphics[width=\maxwidth]{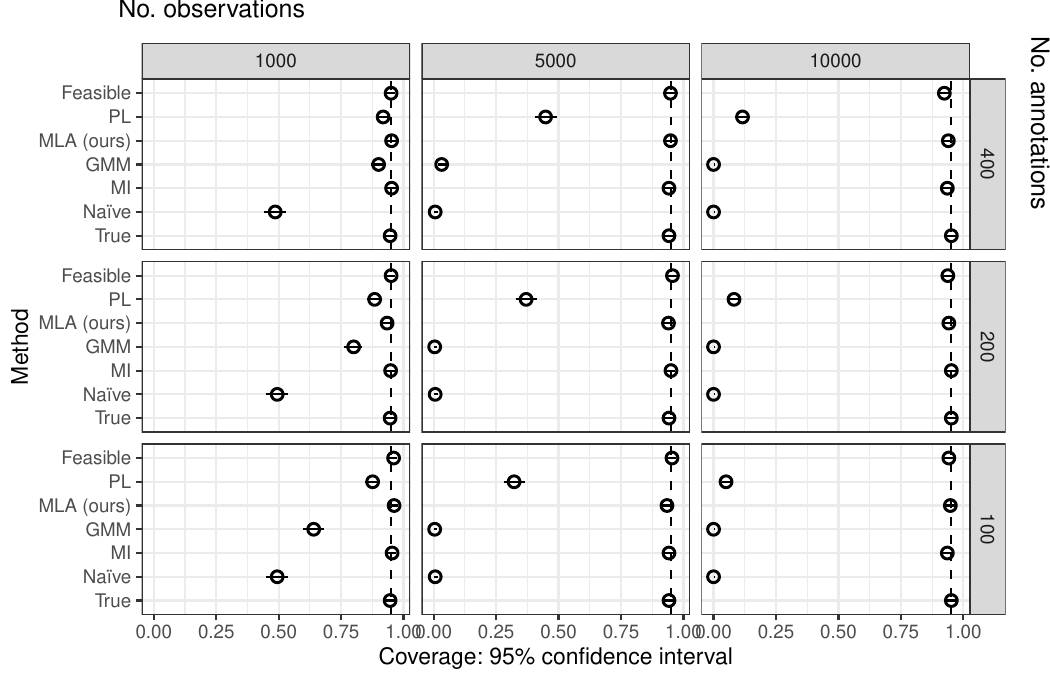} 
\end{knitrout}
\caption{Coverage of 95\% confidence intervals for $B_Z$ in \emph{Simulation 1b}.}
\label{fig:sim1b.z.coverage.z}
\end{figure}

\begin{figure}[htbp!]
\begin{knitrout}
\definecolor{shadecolor}{rgb}{0.969, 0.969, 0.969}\color{fgcolor}
\includegraphics[width=\maxwidth]{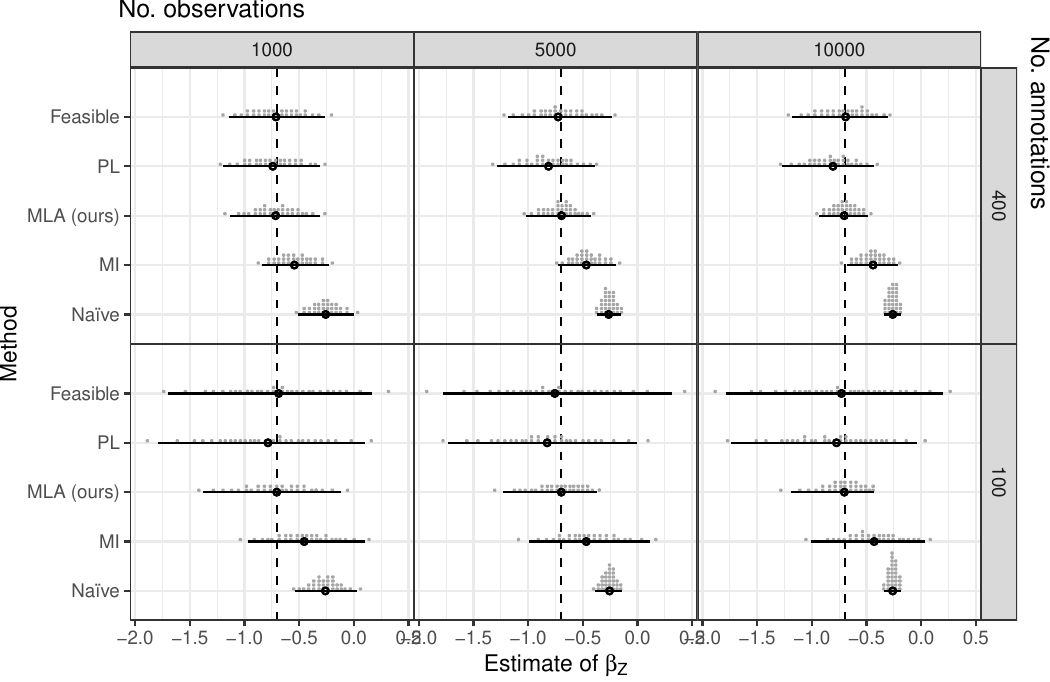} 
\end{knitrout}
\caption{Estimates of $B_Z$ in \emph{simulation 2a}. Only our MLA approach is consistent for estimating $B_Z$.}
\label{fig:sim2a.z}
\end{figure}

\begin{figure}[htbp!]
\begin{knitrout}
\definecolor{shadecolor}{rgb}{0.969, 0.969, 0.969}\color{fgcolor}
\includegraphics[width=\maxwidth]{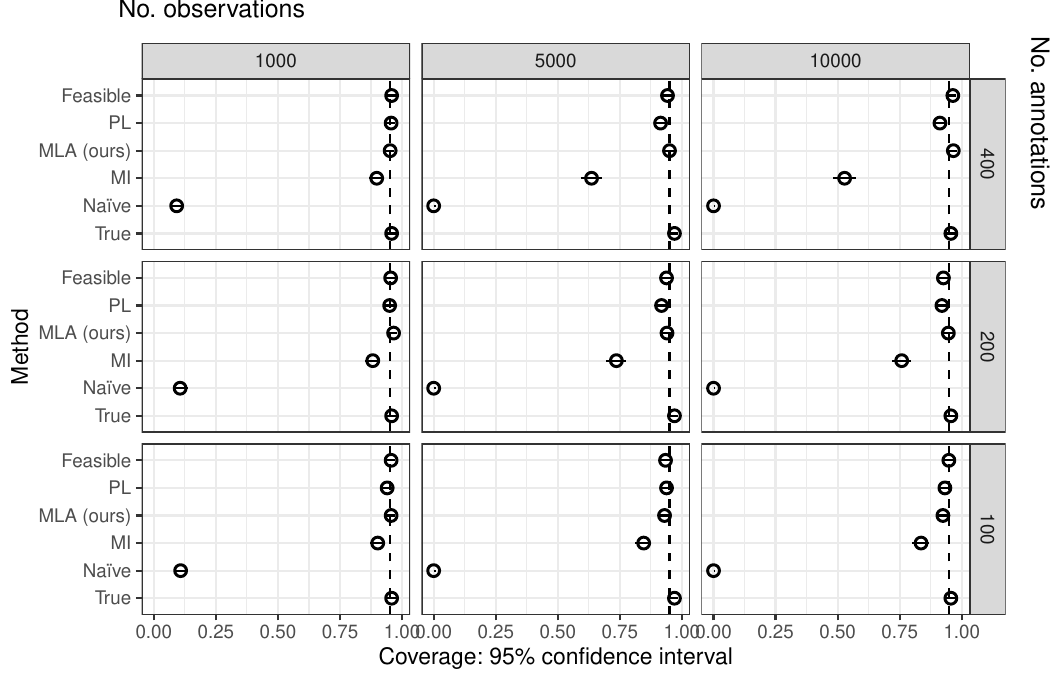} 
\end{knitrout}
\caption{Coverage of 95\% confidence intervals for $B_Z$ in \emph{Simulation 2a}.}
\label{fig:sim2a.z.coverage.z}
\end{figure}

\begin{figure}[htbp!]
\begin{knitrout}
\definecolor{shadecolor}{rgb}{0.969, 0.969, 0.969}\color{fgcolor}
\includegraphics[width=\maxwidth]{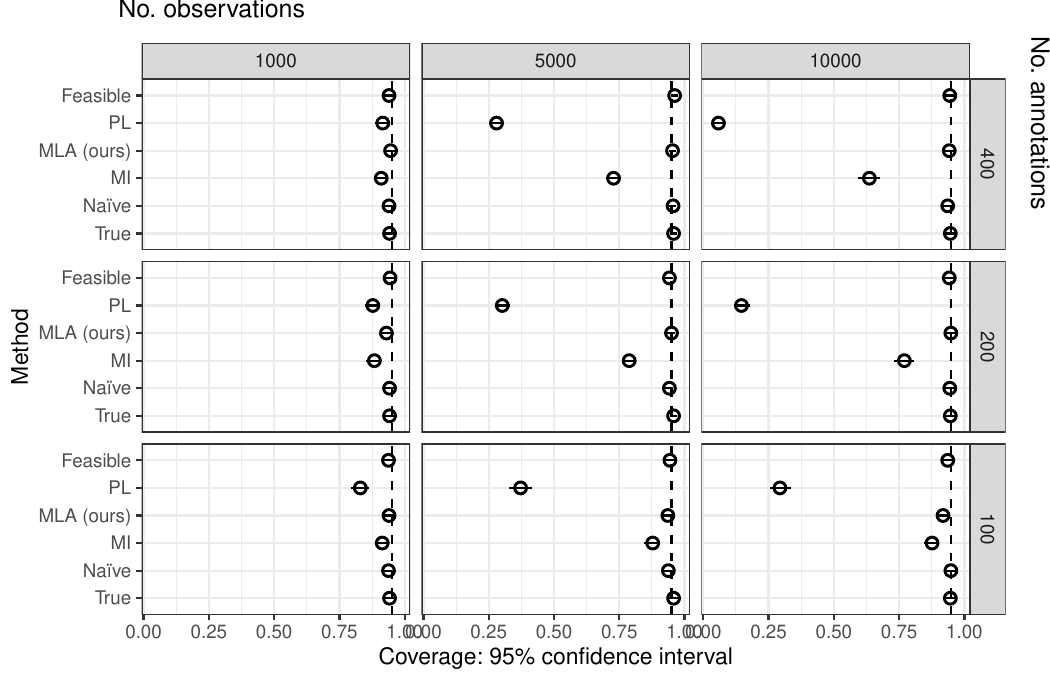} 
\end{knitrout}
\caption{Coverage of 95\% confidence intervals for $B_Z$ in \emph{Simulation 2b}.}
\label{fig:sim2b.z.coverage.z}
\end{figure}

Appendix \ref{appendix:main.sim.plots} includes addition plots for our main simulations across \emph{Simulation 1a-2b}. First, Figures \ref{fig:sim.1a.coverage.x}, \ref{fig:sim.1b.coverage.x}, \ref{fig:sim.2a.coverage.x}, and \ref{fig:sim.2b.coverage.x} plot the coverage of the 95\% confidence intervals over the simulations reported in our main text. Then, we present visualizations of results for the independent variables unreported in the main text. Figure \ref{fig:sim1a.z} visualizes estimates of $B_Z$ in \emph{Simulation 1a}, with respective confidence intervals in Figure \ref{fig:sim1a.z.coverage.z}. Figure \ref{fig:sim1b.z} visualizes estimates of $B_Z$ in \emph{Simulation 1b}, with respective confidence intervals in Figure \ref{fig:sim1b.z.coverage.z}. Figure \ref{fig:sim2a.z} visualizes estimates of $B_Z$ in \emph{Simulation 2a}, with respective confidence intervals in Figure \ref{fig:sim2a.z.coverage.z}. Lastly, Figure \ref{fig:sim2b.z} visualizes estimates of $B_Z$ in \emph{Simulation 2b}, with respective confidence intervals in Figure \ref{fig:sim2b.z.coverage.z}.

\section{Robustness Tests}\label{appendix:robustness}

Appendix \ref{appendix:robustness} presents robustness tests that extend our simulations to further explore the capabilities and limitations of error correction methods. In the following sections, we show what happens when the error model is misspecified (section \ref{appendix:misspec}), when the accuracy of the classifier varies (section \ref{appendix:accuracy}), when the classified variable is not balanced but skewed  (section \ref{appendix:imbalanced}), when the degree of systematic misclassification changes (section \ref{appendix:degreebias}), when the correlation between $X$ and $Y$ is stronger or weaker (section \ref{appendix:robustness.v}), and when parametric assumptions of MLA are violated (section \ref{appendix:robustness.vi}).

\subsection{Robustness Test I: Misspecification of the Error Correction Model}
\label{appendix:misspec}
In \emph{Simulation 1b} and \emph{2b}, the MLA method was able to correct systematic misclassification using the error models in equations \ref{eq:covariate.reg.general} and \ref{eq:depvar.general}.
However, this depends on the error model consistently estimating the conditional probability of automated classifications given the true value and the outcome.
If the misclassifications and the outcome are conditionally dependent on an omitted variable $Z$, this will not be possible. 
Here, we demonstrate how misspecification of the error correction model affects results in the context of misclassification in an independent variable (see section \ref{appendix:misspec.iv}) and a dependent variable (see section \ref{appendix:misspec.dv}). 

\subsubsection{Systematic Misclassification of an Independent Variable}
\label{appendix:misspec.iv}
\begin{figure}[htpb!]
\begin{subfigure}{0.95\textwidth}
\begin{knitrout}
\definecolor{shadecolor}{rgb}{0.969, 0.969, 0.969}\color{fgcolor}
\includegraphics[width=\maxwidth]{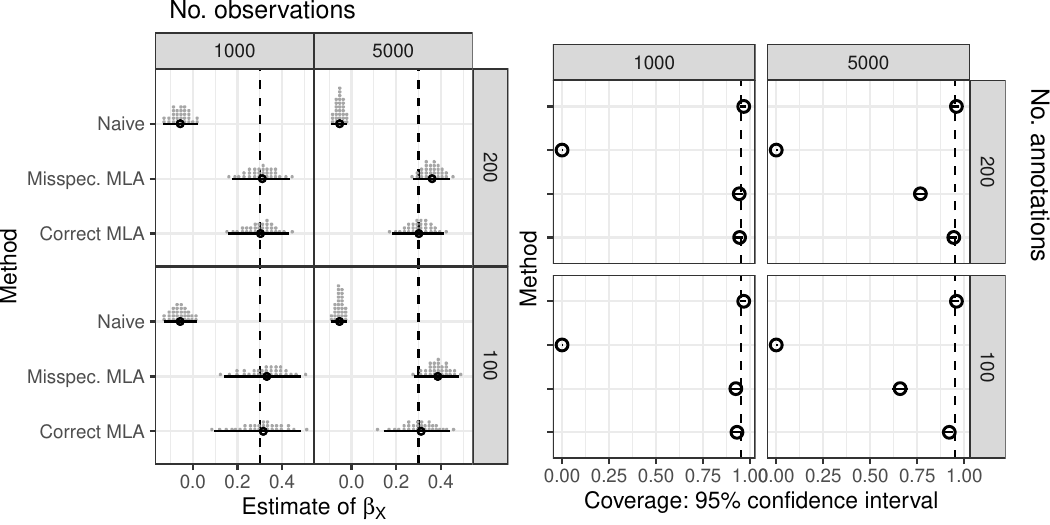} 
\end{knitrout}
\label{fig:iv.noz.x}
\caption{Estimates of $B_X$ are close to the true value despite the misspecified error correction model.}
\end{subfigure}

\begin{subfigure}{0.95\textwidth}
\begin{knitrout}
\definecolor{shadecolor}{rgb}{0.969, 0.969, 0.969}\color{fgcolor}
\includegraphics[width=\maxwidth]{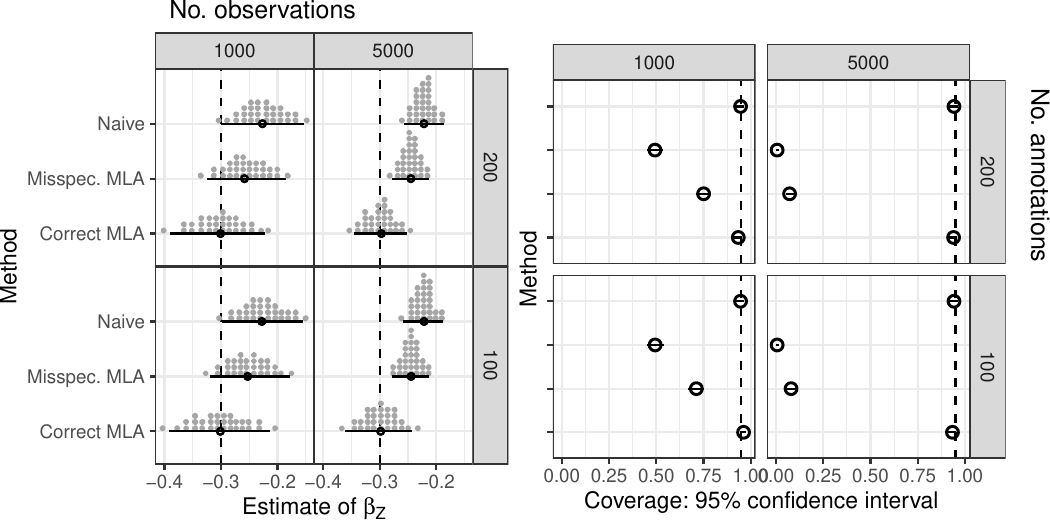} 
\end{knitrout}
\label{fig:iv.noz.z}
\caption{Estimates of $B_Z$ are biased given a misspecified error correction model.}
\end{subfigure}
\caption{Robustness Test I: Misspecification of the Error Correction Model, Simulation 1b}
\label{fig:iv.noz}
\end{figure}

Repeating \emph{Simulation 1b}, what happens when the error model is misspecified? Figure \ref{fig:iv.noz} visualizes effects on $B_X$ (upper panel) and $B_Z$ (lower panel). It shows that a misspecified MLA model is unable to fully correct misclassification bias: Although estimates of $B_X$ are close to the true estimate and estimates of $B_Z$ are better than the näive estimator, $\widehat{B_Z}$ is still clearly biased.

\subsubsection{Systematic Misclassification of a Dependent Variable}
\label{appendix:misspec.dv}
\begin{figure}[htpb!]
\begin{subfigure}{0.95\textwidth}
\begin{knitrout}
\definecolor{shadecolor}{rgb}{0.969, 0.969, 0.969}\color{fgcolor}
\includegraphics[width=\maxwidth]{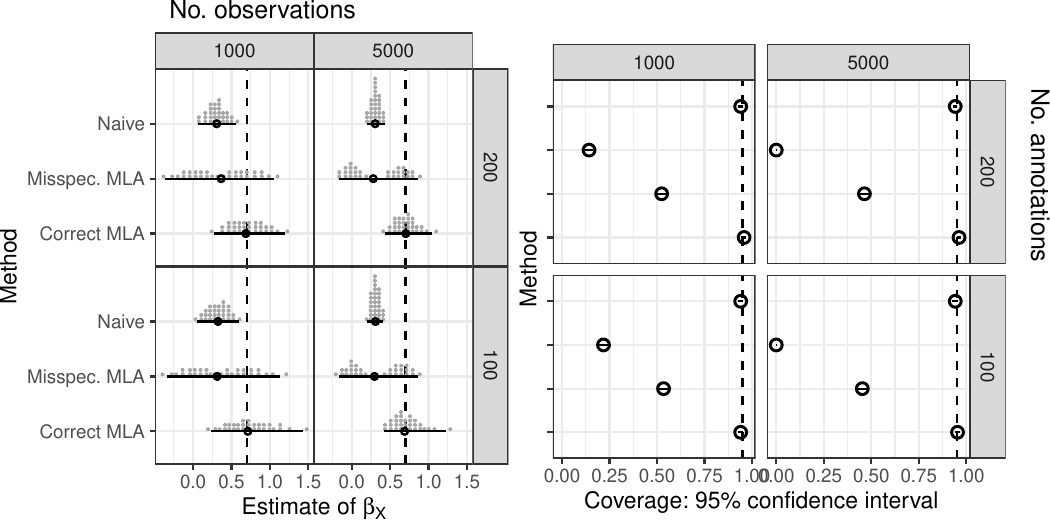} 
\end{knitrout}
\label{fig:dv.noz.x}
\caption{Estimates of $B_X$ are close to the true value despite the misspecified error correction model.}
\end{subfigure}

\begin{subfigure}{0.95\textwidth}
\begin{knitrout}
\definecolor{shadecolor}{rgb}{0.969, 0.969, 0.969}\color{fgcolor}
\includegraphics[width=\maxwidth]{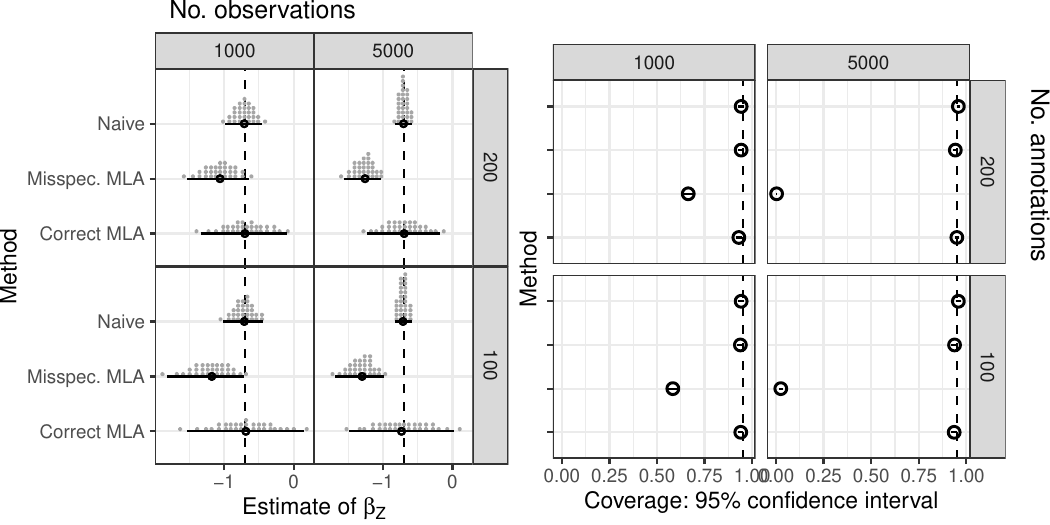} 
\end{knitrout}
\label{fig:dv.noz.z}
\caption{Estimates of $B_Z$ are biased given a misspecified error correction model.}
\end{subfigure}
\caption{Robustness Test I: Misspecification of the Error Correction Model, Simulation 2b}
\label{fig:dv.noz}
\end{figure}

Next, we repeat \emph{Simulation 2b} with a misspecified error correction model. Figure \ref{fig:dv.noz} shows that a misspecified error model is, again, unable to obtain consistent estimates of $B_Z$. 

\clearpage

\subsection{Robustness Test II: Varying Accuracy of the Automated Classifier}
\label{appendix:accuracy}

According to our literature review, the accuracy of reported classifiers varies considerably. But how does the performance of the classifier affect error correction methods and remaining bias in inferential modeling? To test this, we repeat \emph{Simulation 1a} (see Section \ref{appendix:iv.predacc}) and \emph{Simulation 2a} (see Section \ref{appendix:dv.predacc}) to show how varying accuracy of the AC affects estimates of independent variables $B_X$ and $B_Z$. Here, we let classifier accuracy range
from 60\% to 95\%. We present results for a scenario with 5,000 classifications and 200 manual annotations. 

\subsubsection{Varying Accuracy of an AC Predicting an Independent Variable}
\label{appendix:iv.predacc}
\begin{figure}[htpb!]

\begin{knitrout}
\definecolor{shadecolor}{rgb}{0.969, 0.969, 0.969}\color{fgcolor}
\includegraphics[width=\maxwidth]{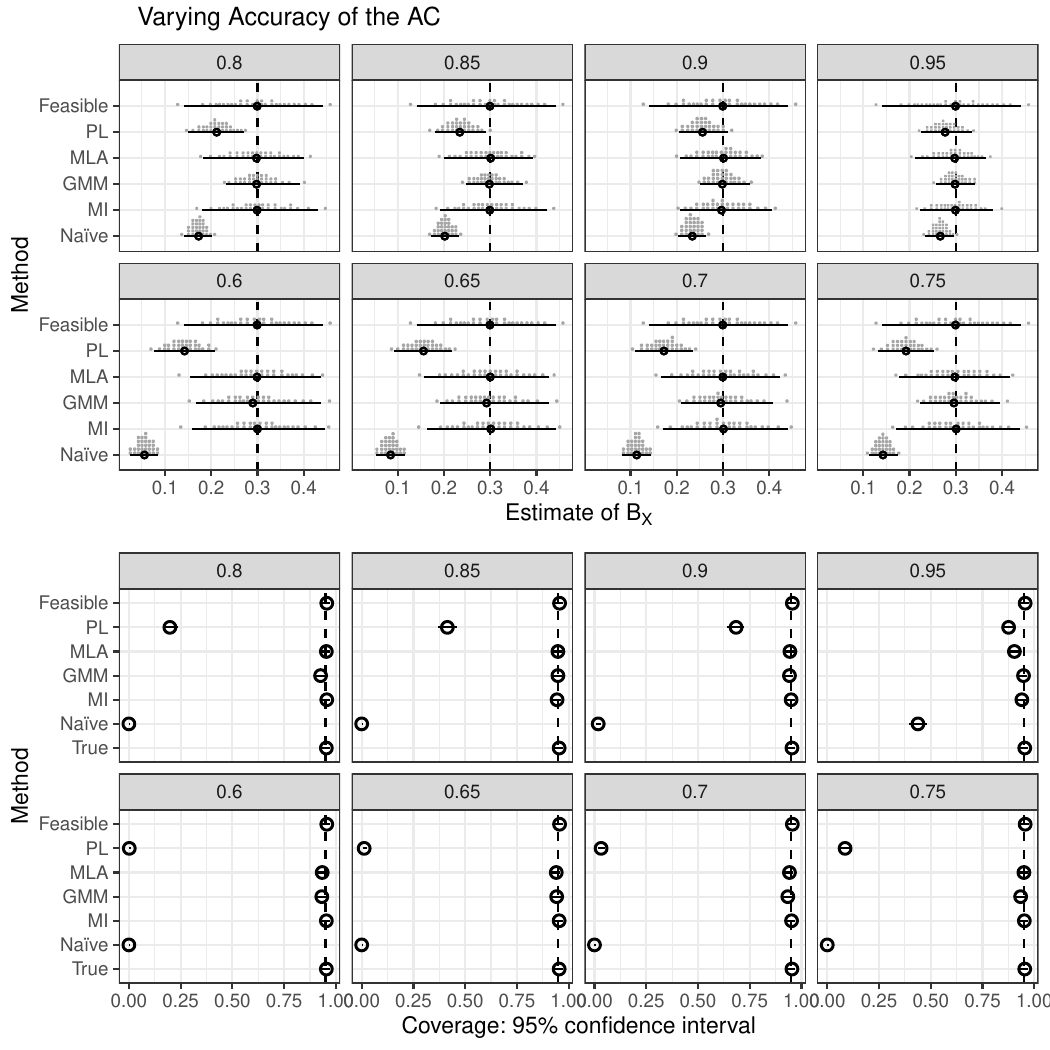} 
\end{knitrout}
\caption{Estimates of $B_X$ improve with higher accuracy of the AC.\label{fig:robustness.2iv.x}}
\end{figure}

\begin{figure}
\begin{knitrout}
\definecolor{shadecolor}{rgb}{0.969, 0.969, 0.969}\color{fgcolor}
\includegraphics[width=\maxwidth]{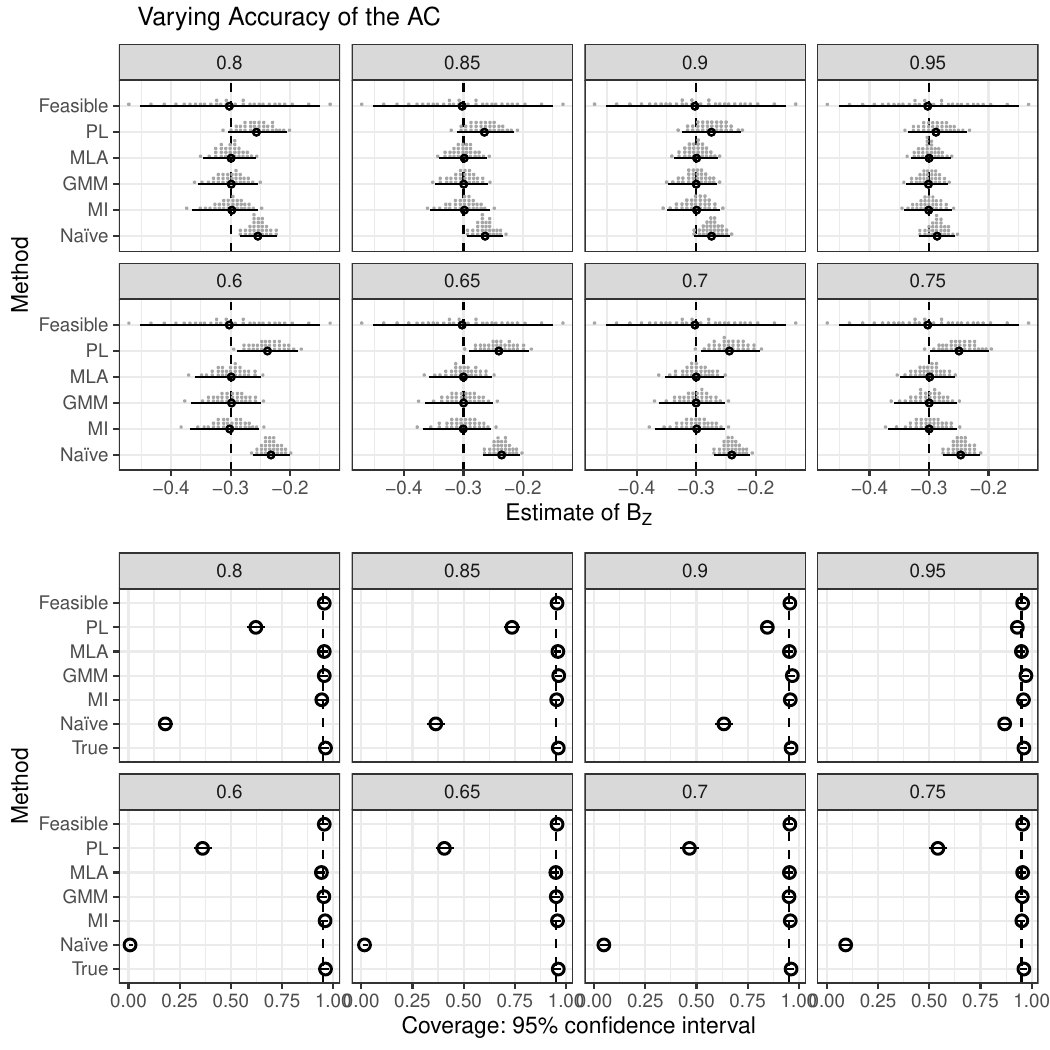} 
\end{knitrout}
\caption{Estimates of $B_Z$ improve with higher accuracy of the AC.\label{fig:robustness.2iv.z}}
\end{figure}

In Figures \ref{fig:robustness.2iv.x} and \ref{fig:robustness.2iv.z}, we present results extending \emph{Simulation 1a} where the independent variable is created via an AC. As expected, a more accurate classifier causes less misclassification bias. All the error correction methods also provide more precise estimates when used with a more accurate classifier. 

\clearpage

\clearpage

\subsubsection{Varying Accuracy of an AC Predicting a Dependent Variable}
\label{appendix:dv.predacc}
\begin{figure}[htpb!]
\begin{knitrout}
\definecolor{shadecolor}{rgb}{0.969, 0.969, 0.969}\color{fgcolor}
\includegraphics[width=\maxwidth]{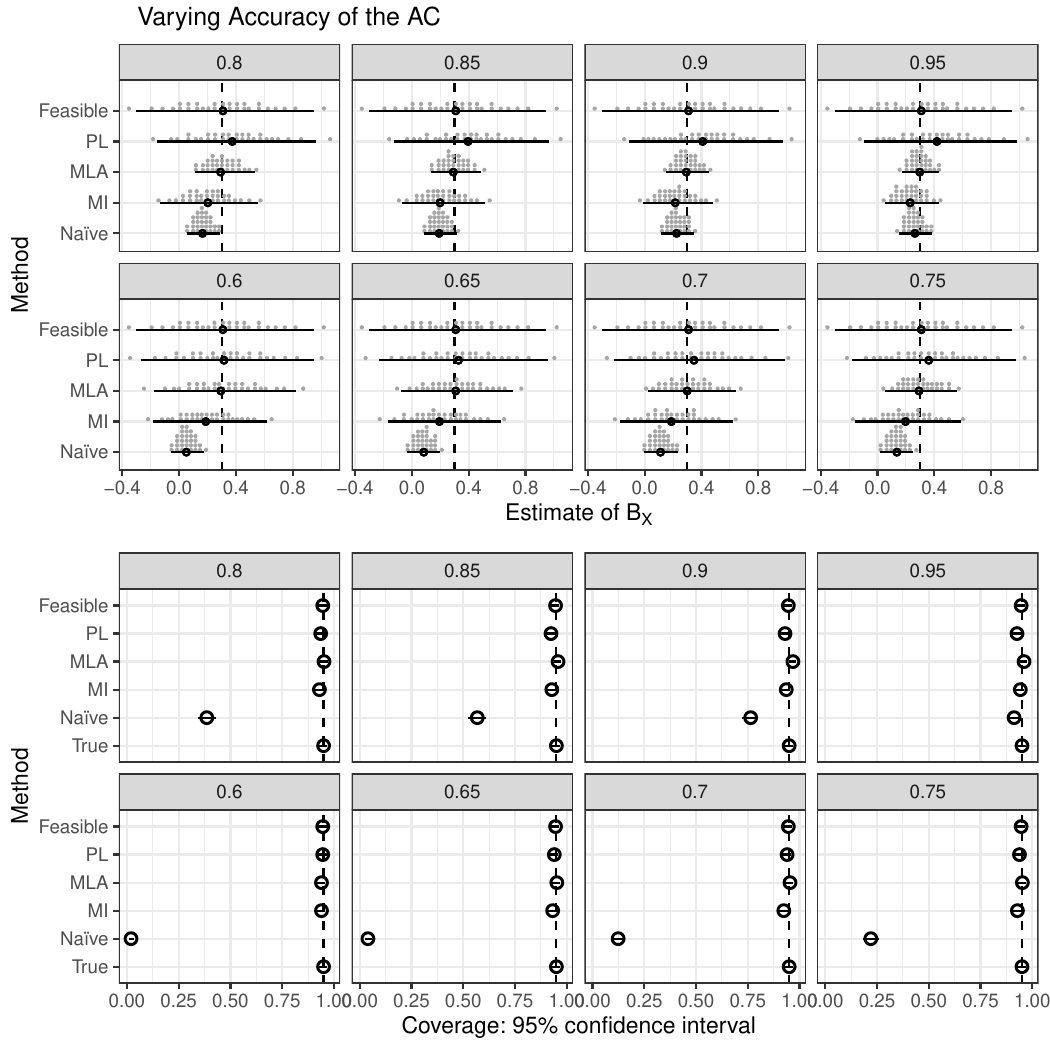} 
\end{knitrout}
\caption{Estimates of $B_X$ improve with higher accuracy of the AC.\label{fig:robustness.2dv.x}}
\end{figure}

\begin{figure}
\begin{knitrout}
\definecolor{shadecolor}{rgb}{0.969, 0.969, 0.969}\color{fgcolor}
\includegraphics[width=\maxwidth]{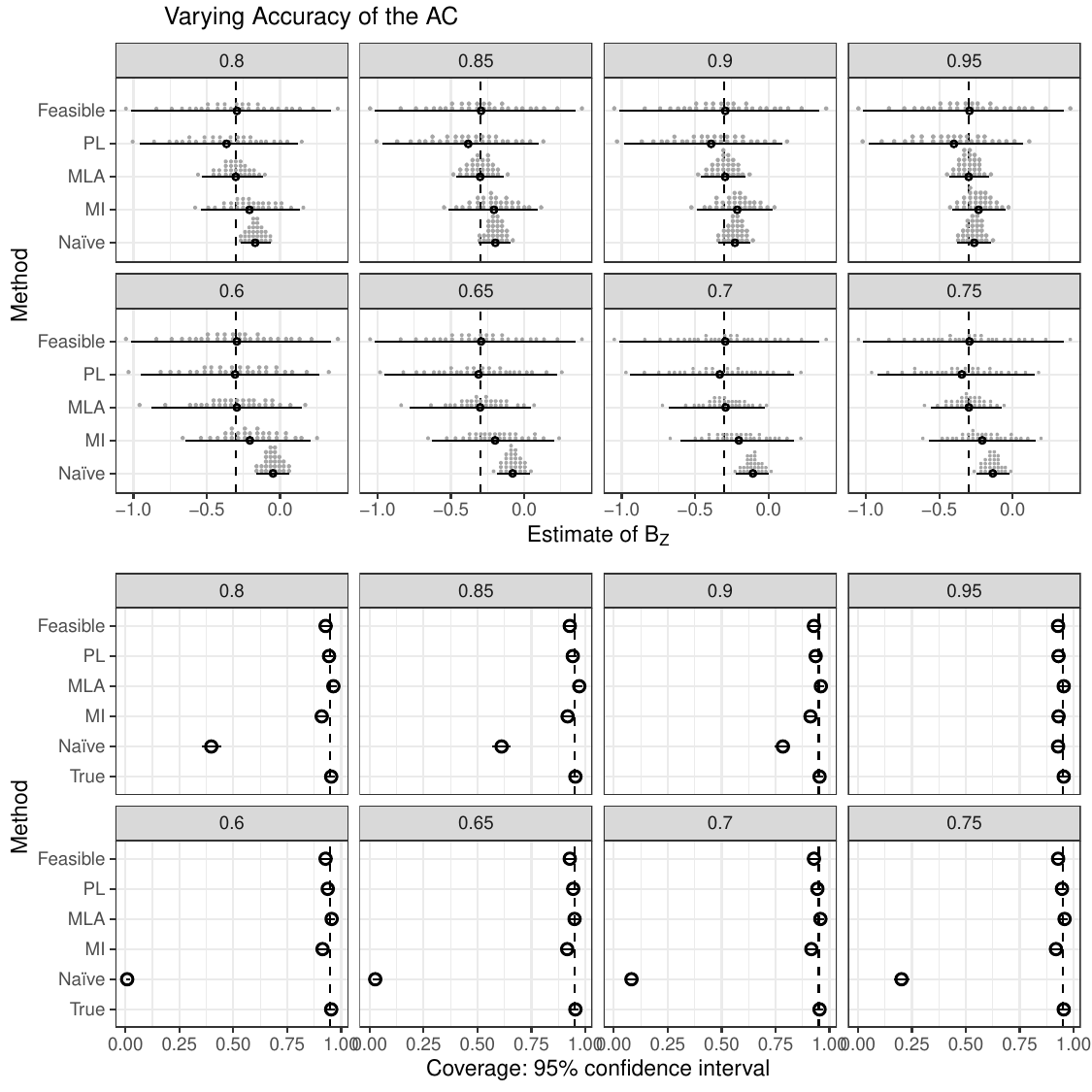} 
\end{knitrout}
\caption{Estimates of $B_Z$ improve with higher accuracy of the AC.\label{fig:robustness.2dv.z}}
\label{fig:dv.predacc}
\end{figure}

We now repeat these simulations for \emph{Simulation 2a}, where the dependent variable is created via an AC. As figures \ref{fig:robustness.2dv.x} and \ref{fig:robustness.2dv.z} show, the patterns are similar: error correction methods provide more precise estimates when used with a more accurate classifier.

\clearpage

\clearpage

\subsection{Robustness Test III: Misclassification in Imbalanced Variables}
\label{appendix:imbalanced}

For simplicity, our main simulations include balanced classified variables.  However, classifiers are often used to measure imbalanced variables, which can be more difficult to predict.  As a next robustness test, we therefore replicate \emph{Simulation 1a} (see section \ref{appendix:imbalanced.iv}) and \emph{Simulation 2a} (see section \ref{appendix:imbalanced.dv}) to analyze whether the MLA error correction method performs similarly well with imbalanced classified variables. We do so for the scenario with 5,000 classifications and 200 manual annotations.

\subsubsection{Imbalance in Classified Independent Variables}
\label{appendix:imbalanced.iv}

\begin{figure}[htpb!]

\begin{knitrout}
\definecolor{shadecolor}{rgb}{0.969, 0.969, 0.969}\color{fgcolor}
\includegraphics[width=\maxwidth]{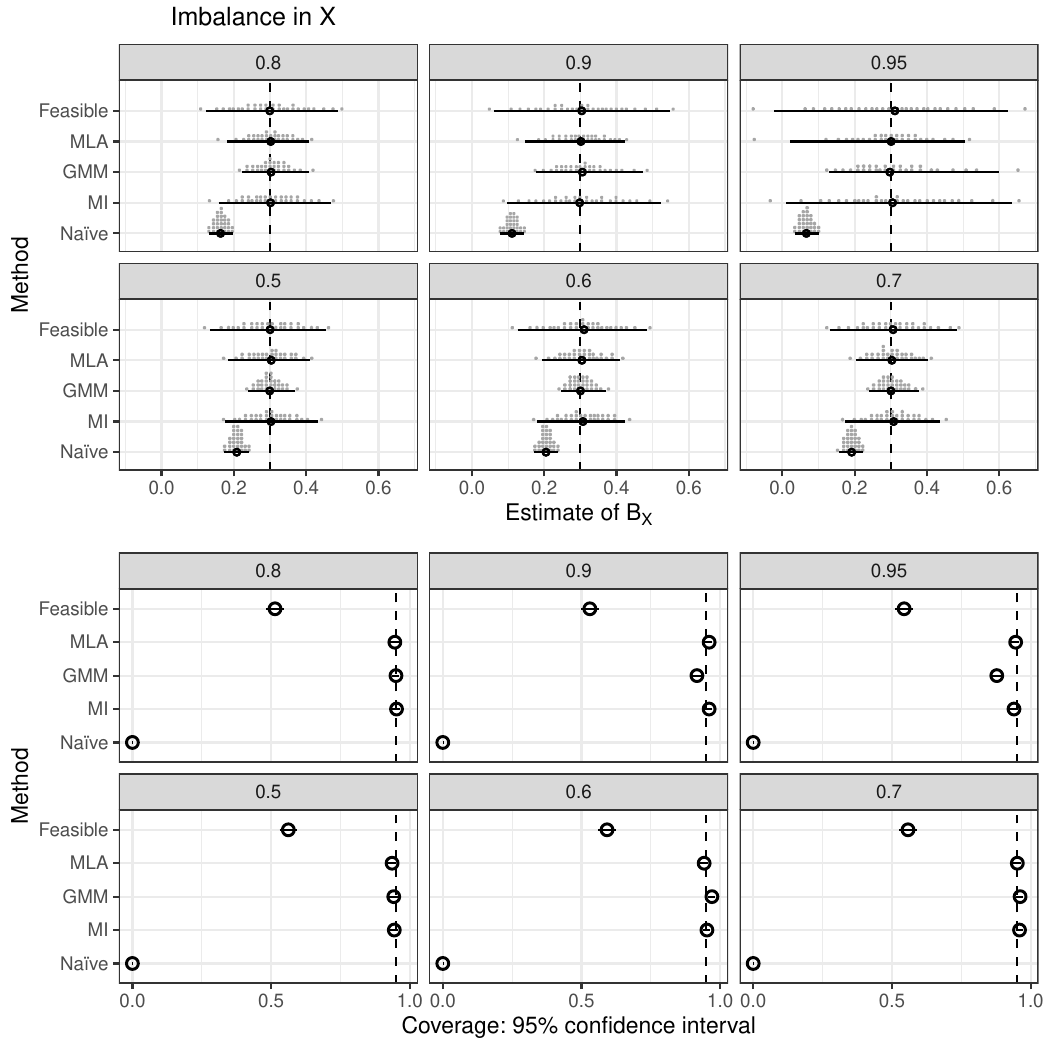} 
\end{knitrout}
\caption{Estimates of $B_X$ are close to true values given imbalance in $X$. \label{fig:iv.imbalanced.bx}}
\end{figure}

\begin{figure}
\begin{knitrout}
\definecolor{shadecolor}{rgb}{0.969, 0.969, 0.969}\color{fgcolor}
\includegraphics[width=\maxwidth]{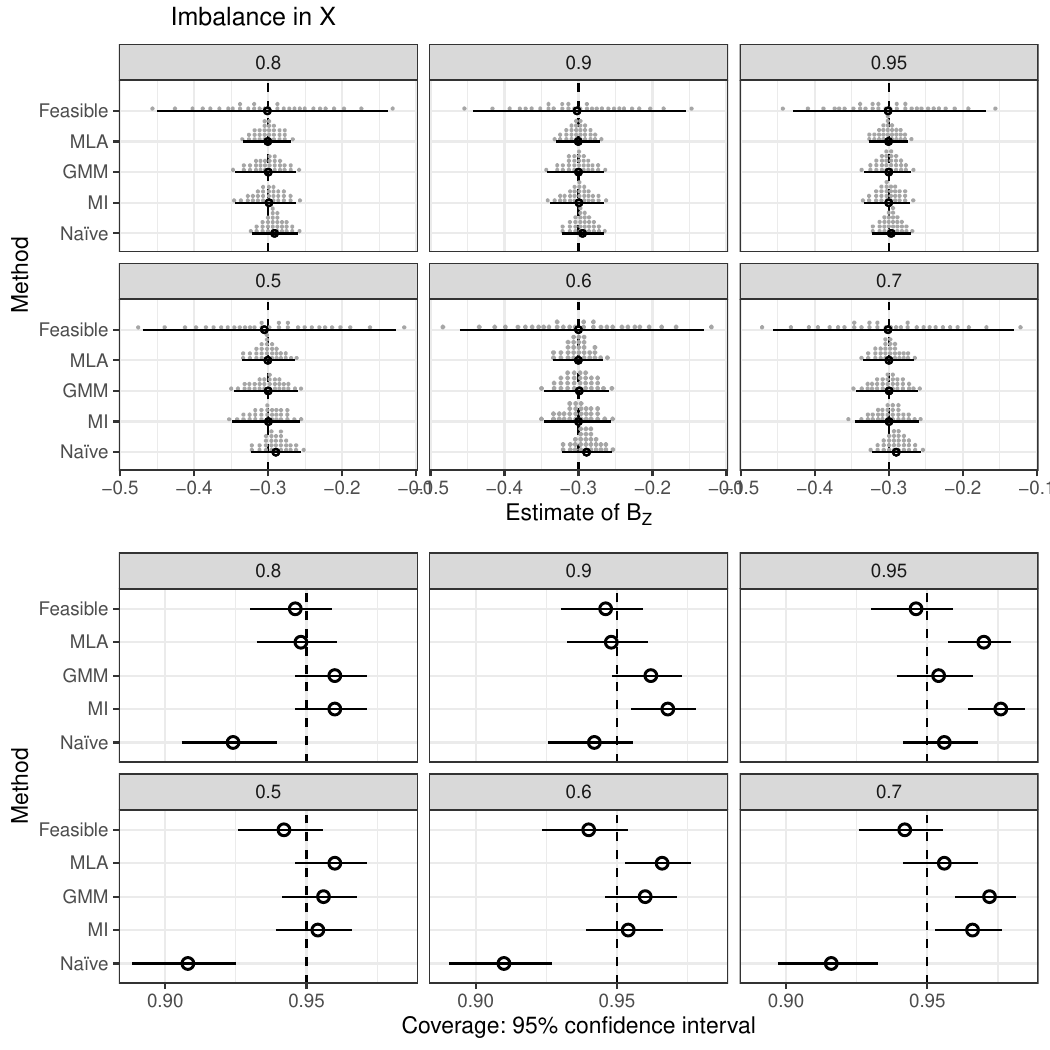} 
\end{knitrout}
\caption{Estimates of $B_Z$ are close to true values given imbalance in $X$.}
\label{fig:iv.imbalanced}
\end{figure}

Replicating \emph{Simulation 1a}, Figure \ref{fig:iv.imbalanced} illustrates that our MLA method performs similarly well with imbalance in classified independent variables.
However, the quality of uncertainty quantification of methods tends to degrade as imbalance increases as shown in Figure \ref{fig:iv.imbalanced}.  This suggests that imbalanced data requires additional validation data for effective misclassification correction.  Please note that the PL approach has a very large range of estimates and is thus excluded in Figures \ref{fig:iv.imbalanced.bx} and \ref{fig:iv.imbalanced} for readability.

\clearpage

\subsubsection{Imbalance in Classified Dependent Variables}
\label{appendix:imbalanced.dv}
\begin{figure}[htpb!]
\begin{knitrout}
\definecolor{shadecolor}{rgb}{0.969, 0.969, 0.969}\color{fgcolor}
\includegraphics[width=\maxwidth]{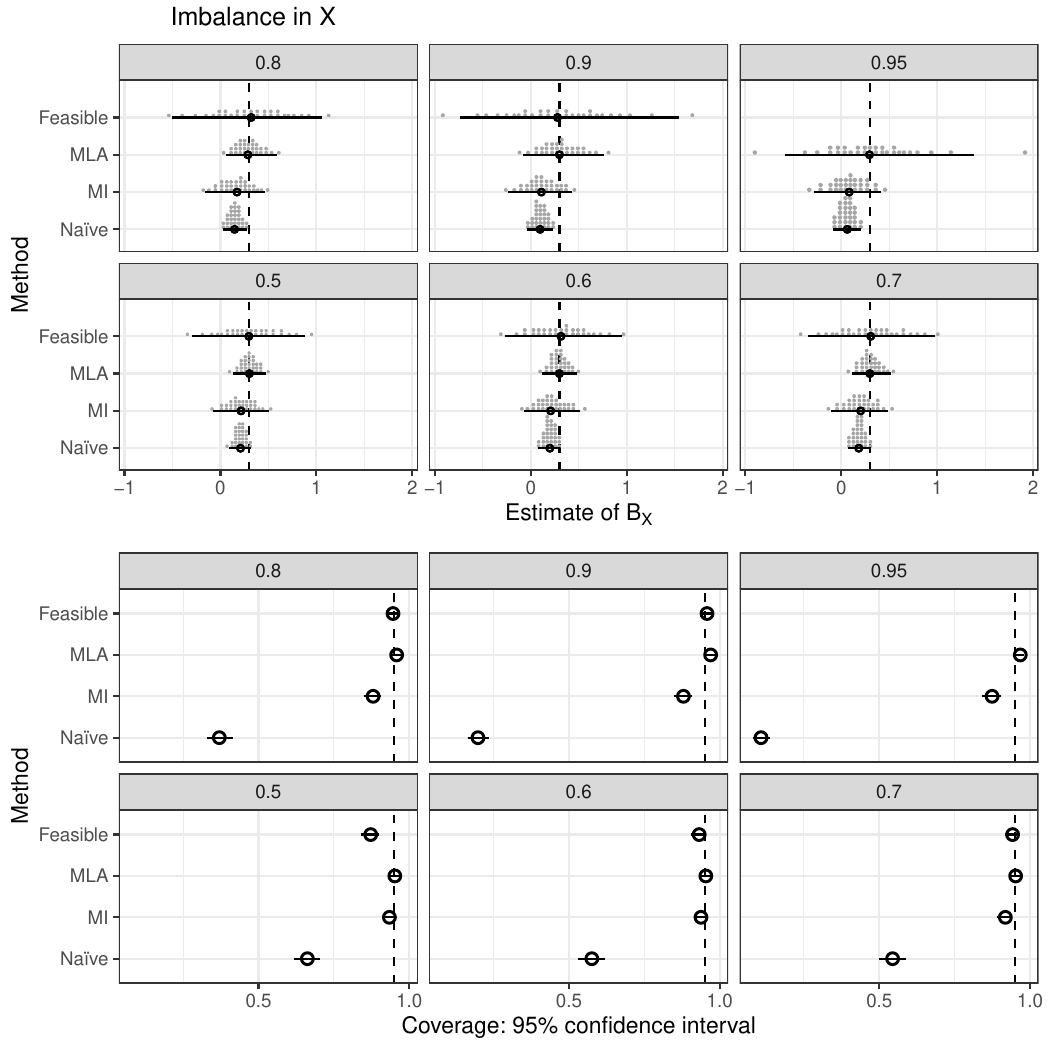} 
\end{knitrout}
\caption{Estimates of $B_X$ are close to true values given imbalance in $Y$. \label{fig:dv.imbalanced.x}}
\end{figure}

\begin{figure}
\begin{knitrout}
\definecolor{shadecolor}{rgb}{0.969, 0.969, 0.969}\color{fgcolor}
\includegraphics[width=\maxwidth]{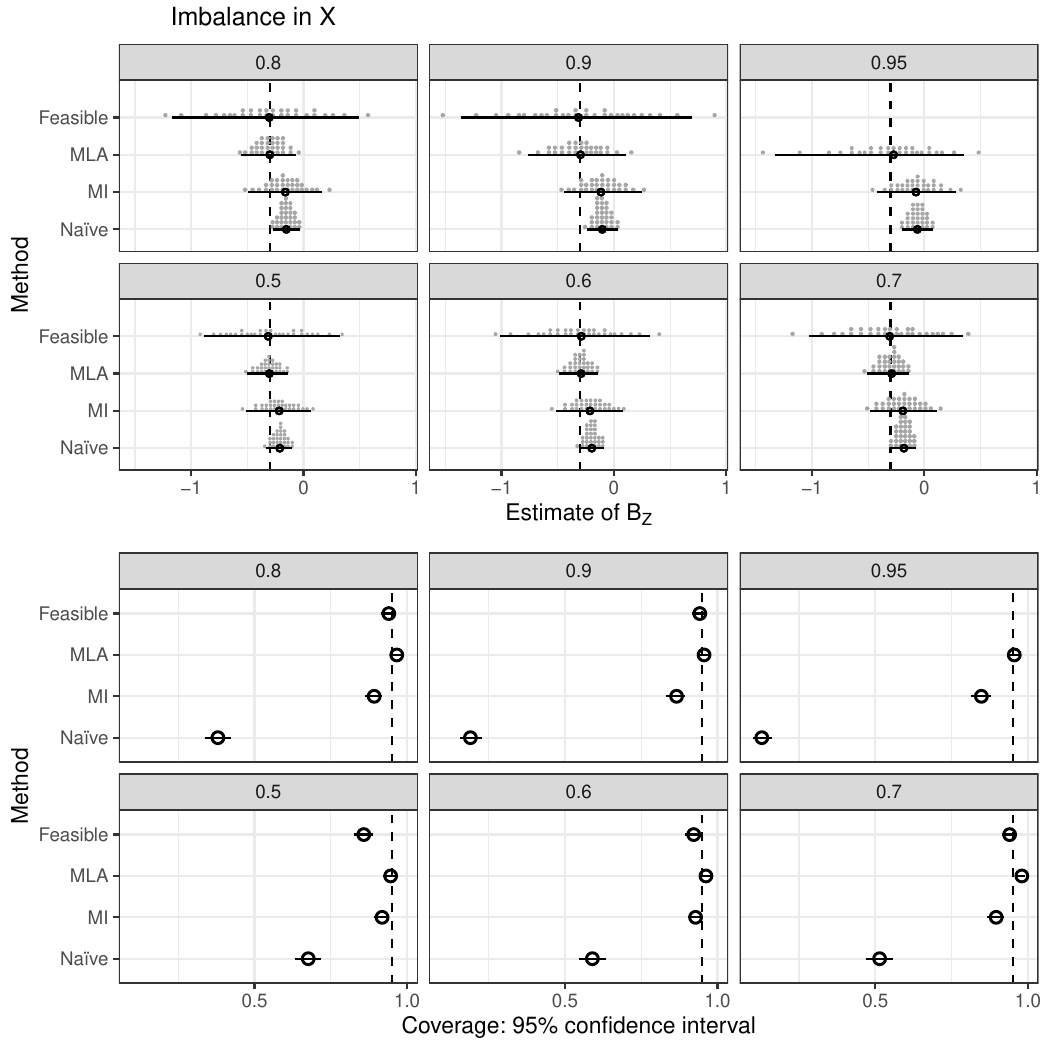} 
\end{knitrout}
\caption{Estimates of $B_Z$ are close to true values given imbalance in $Y$. \label{fig:dv.imbalanced.z}}
\end{figure}

Replicating \emph{Simulation 2a}, Figures \ref{fig:dv.imbalanced.x} and \ref{fig:dv.imbalanced.z} further illustrate that our MLA method performs similarly well with imbalance in classified dependent variables. The PL approach is, again, removed due to the large range of its estimates. Similarly, the estimates for the feasible method are omitted when the probability of $X$ is 0.95 due to high variance. 

\clearpage

\subsection{Robustness Test IV: Different Degrees of Systematic Misclassification}
\label{appendix:degreebias}

Lastly, we explore what happens if misclassification is more or less systematic. To do so, we replicate \emph{Simulation 1b} (see section \ref{appendix:degreebias.iv}) and \emph{Simulation 2b} (see section \ref{appendix:degreebias.dv}) with 1000 classifications and 100 manual annotations.  We vary the amount of systematic misclassification in \emph{Simulation 1b} via the logistic regression coefficient of $Y$ on $W$ while keeping the overall classifier accuracy close to 0.73. In \emph{Simulation 2b}, we similarly use a range of values for the coefficient of $Z$ on $W$.

\subsubsection{Systematic Misclassification in an Independent Variable}
\label{appendix:degreebias.iv}

Replicating \emph{Simulation 1b}, Figure \ref{fig:iv.degreebias} underlines that our MLA method performs  well even for higher degrees of systematic misclassification in the independent variable.

\begin{figure}[htpb!]
\begin{knitrout}
\definecolor{shadecolor}{rgb}{0.969, 0.969, 0.969}\color{fgcolor}
\includegraphics[width=\maxwidth]{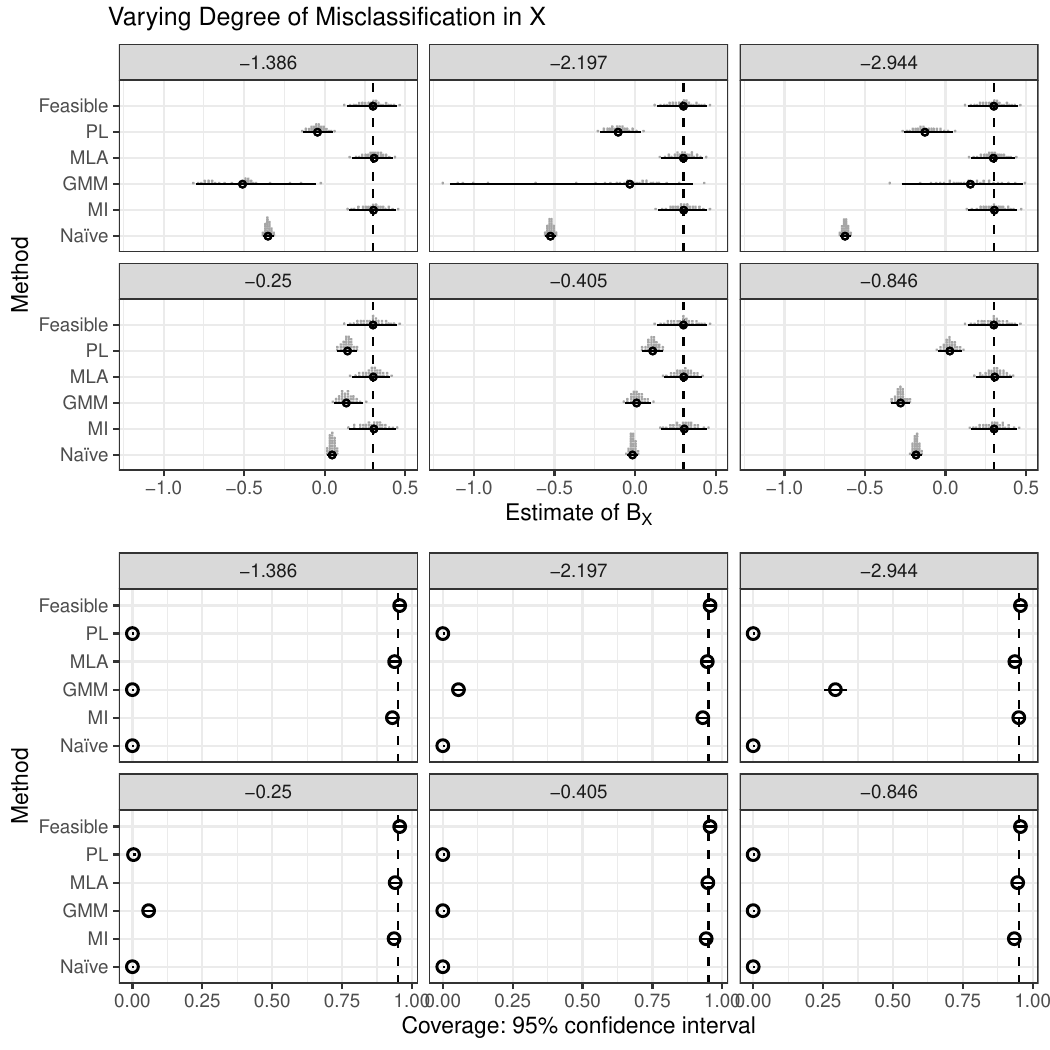} 
\end{knitrout}
\caption{Estimates of $B_X$ are close to true values given different degrees of misclassication in $X$.}
\end{figure}

\clearpage

\begin{figure}
\begin{knitrout}
\definecolor{shadecolor}{rgb}{0.969, 0.969, 0.969}\color{fgcolor}
\includegraphics[width=\maxwidth]{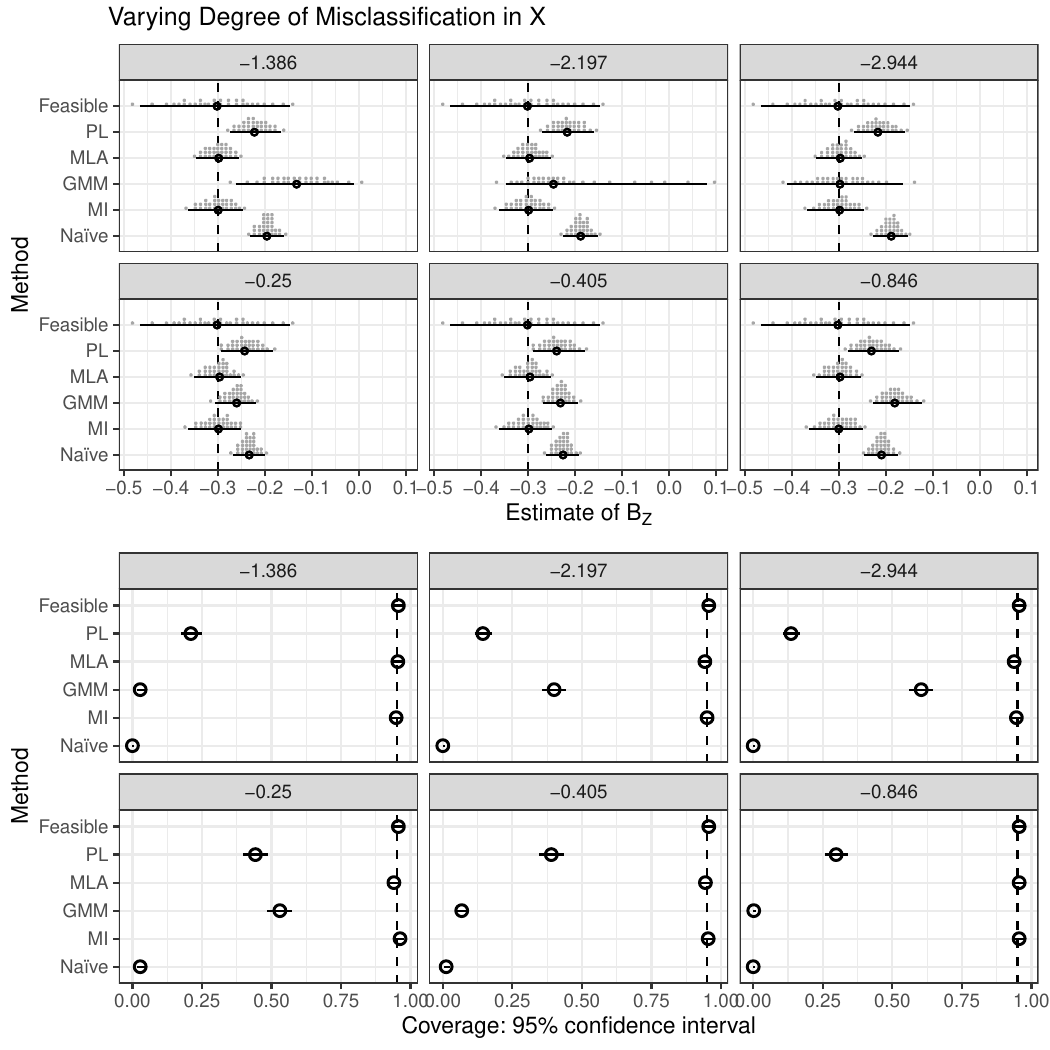} 
\end{knitrout}

\caption{Estimates of $B_Z$ are close to true values given different degrees of misclassication in $X$.}
\end{figure}

\subsubsection{Systematic Misclassification in a Dependent Variable}
\label{appendix:degreebias.dv}
\begin{figure}
\begin{knitrout}
\definecolor{shadecolor}{rgb}{0.969, 0.969, 0.969}\color{fgcolor}
\includegraphics[width=\maxwidth]{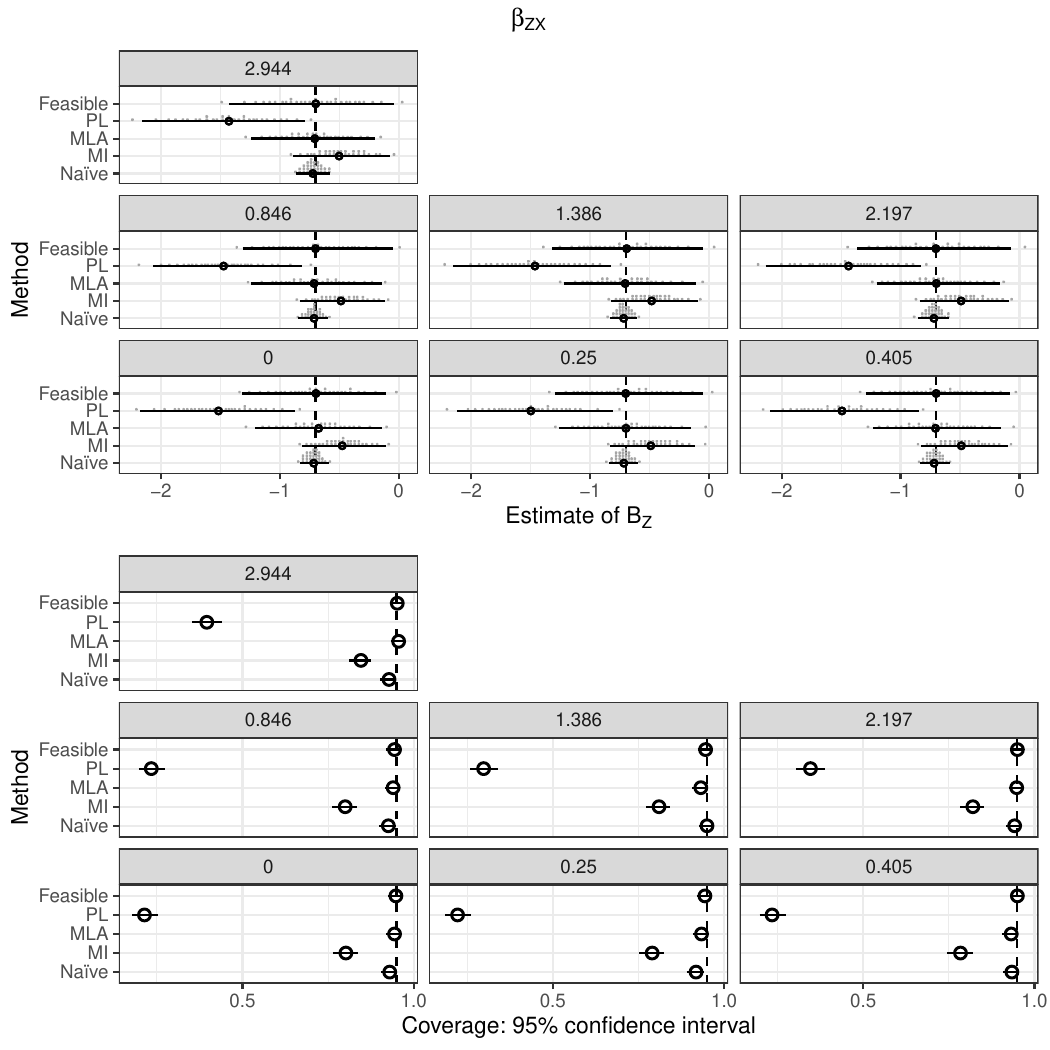} 
\end{knitrout}
\caption{As the correlation between $X$ and $Z$ increases, misclassification of $X$ causes greater bias in estimates of $\beta_X$. MLA and MI are effective correcting this bias.}
\label{fig:dv.degreebias}
\end{figure}

Replicating \emph{Simulation 2b}, Figure \ref{fig:dv.degreebias} comes to similar conclusions. In the case of systematic misclassification in the dependent variable, we can observe that the bias in the naïve estimator switches from negative to positive as systematic misclassification increases. 

\begin{figure}[htpb!]
\begin{knitrout}
\definecolor{shadecolor}{rgb}{0.969, 0.969, 0.969}\color{fgcolor}
\includegraphics[width=\maxwidth]{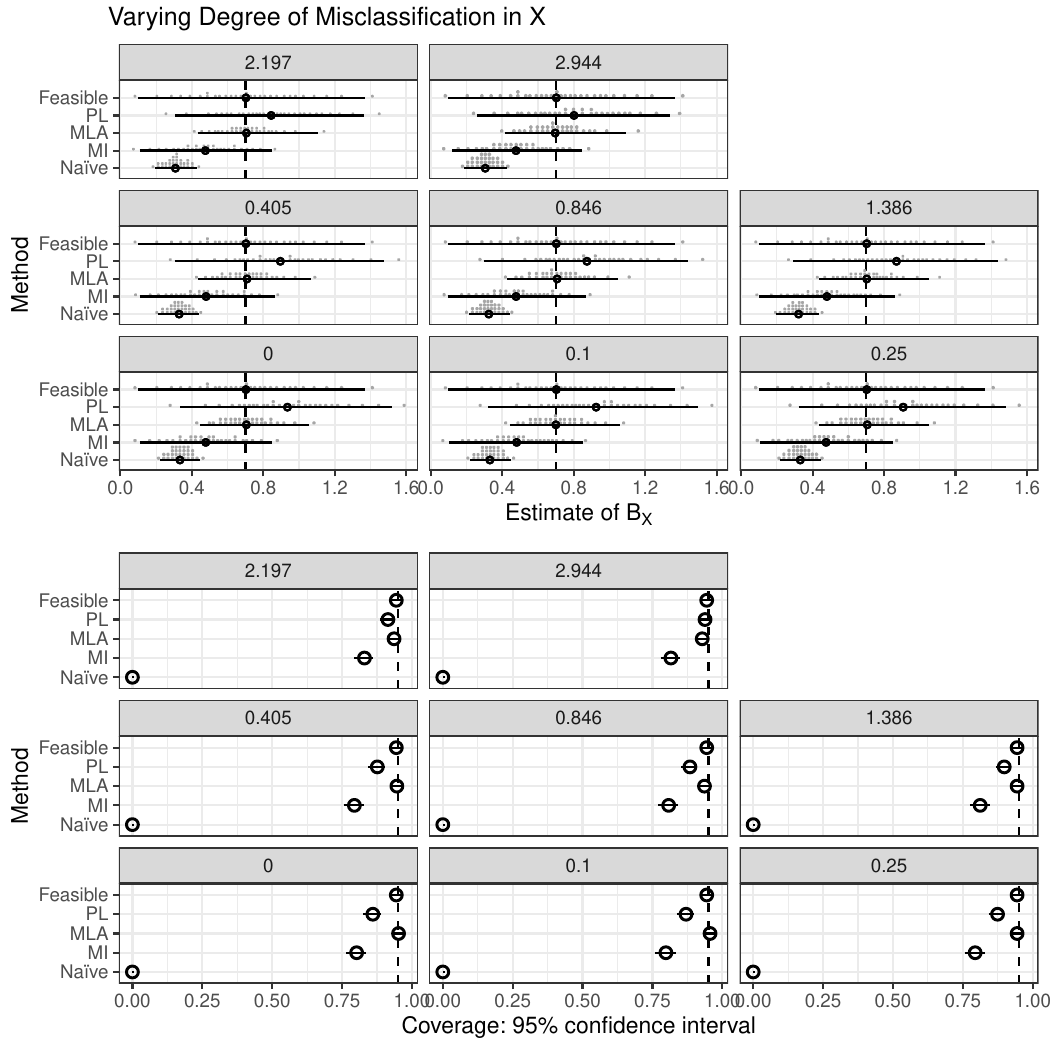} 
\end{knitrout}
\caption{Estimates of $B_X$ are close to true values given different degrees of misclassication in $Y$.}
\end{figure}
\clearpage
\begin{figure}
\begin{knitrout}
\definecolor{shadecolor}{rgb}{0.969, 0.969, 0.969}\color{fgcolor}
\includegraphics[width=\maxwidth]{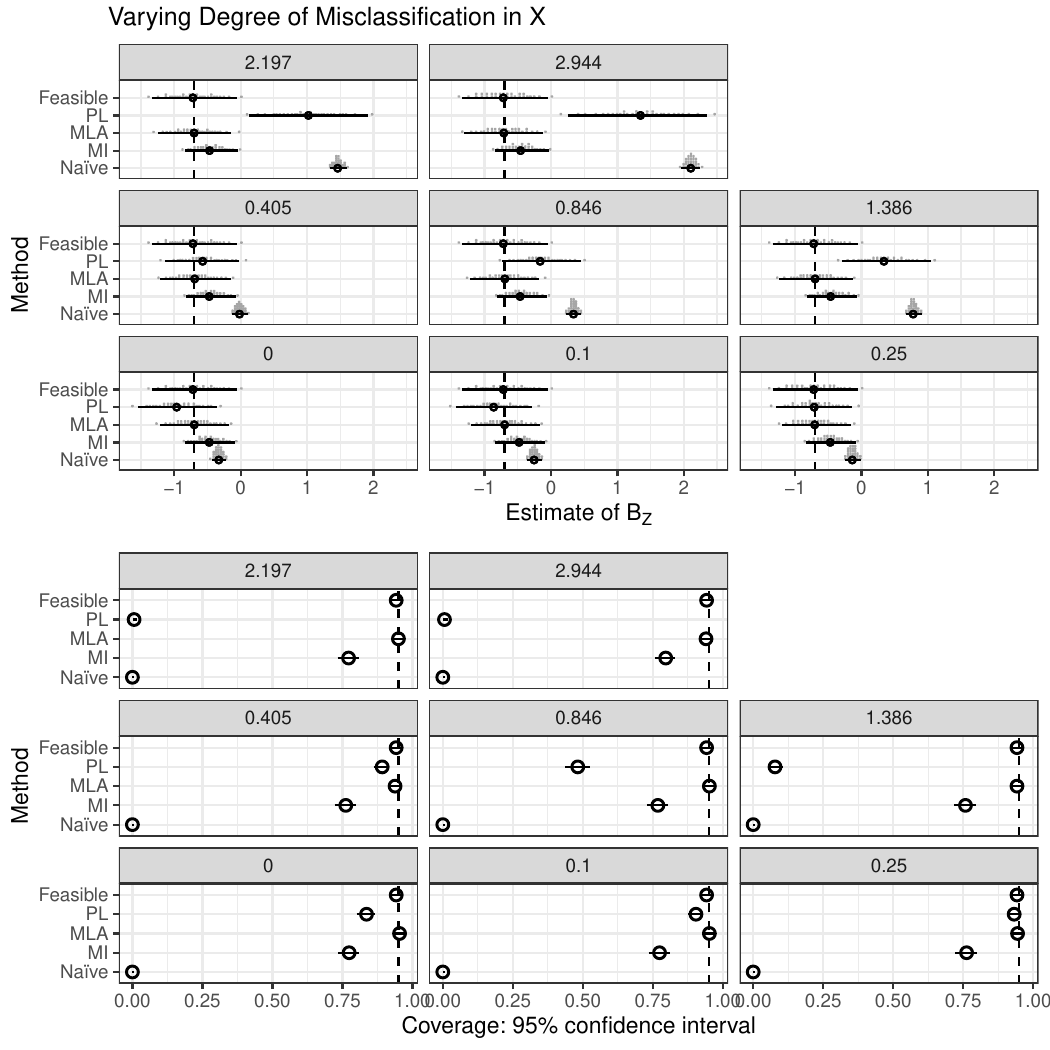} 
\end{knitrout}
\caption{Estimates of $B_Z$ become inconsistent with increasing misclassification in $Y$.}
\end{figure}

\subsection{Robustness Test V: Varying the Correlation between $X$ and $Z$}
\label{appendix:robustness.v}

In our main simulations, $X$ and $Z$ are weakly correlated (Pearson's $\rho=0.24$). Here we explore what happens when this correlation is more or less strong. We do this by repeating example 1.b with 200 annotations and 5000 classifications and various values for the coefficient of $Z$ on $X$ in the generative model. This  coefficient is set to 1 in our main simulations.

\subsubsection{Increasing correlation between $X$ and $Z$ with misclassification in the IV}

\begin{figure}
\begin{knitrout}
\definecolor{shadecolor}{rgb}{0.969, 0.969, 0.969}\color{fgcolor}
\includegraphics[width=\maxwidth]{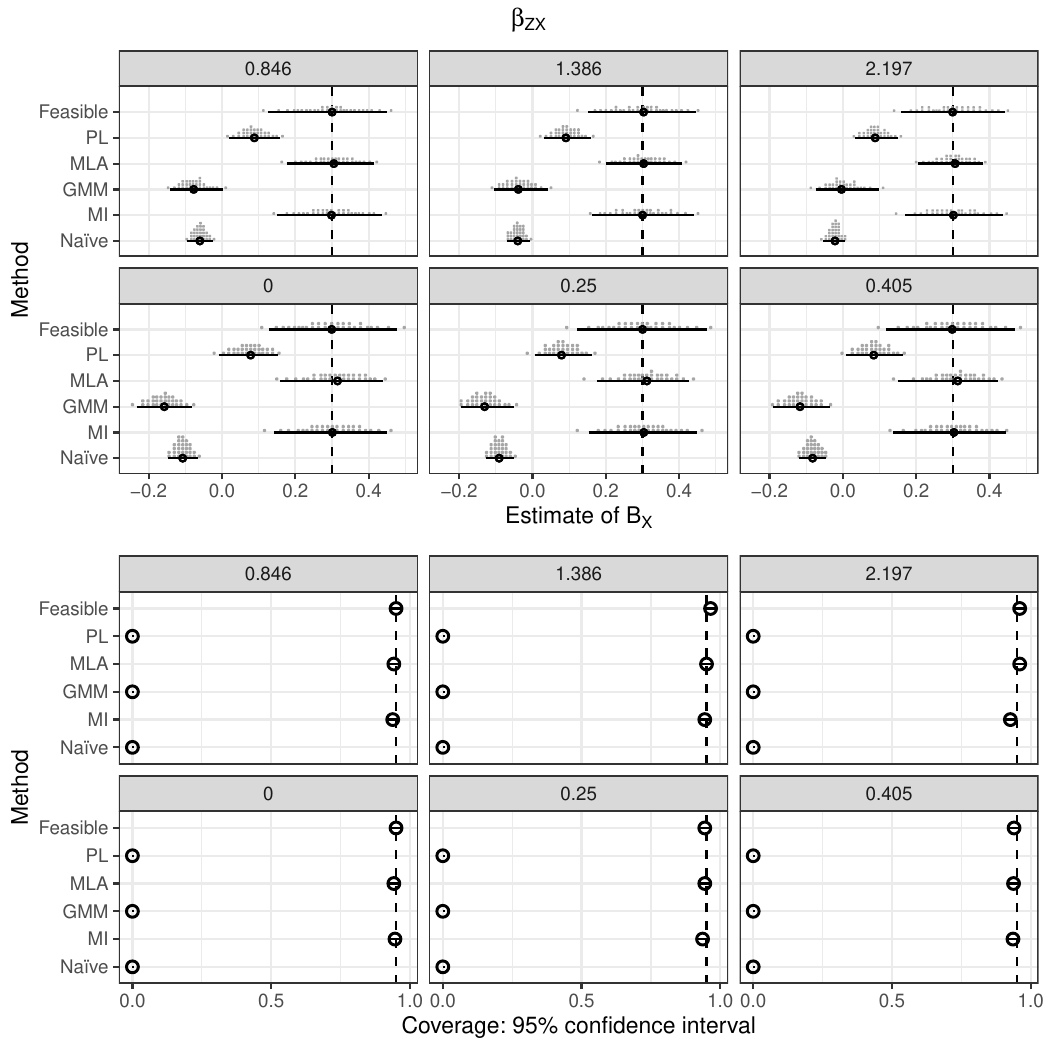} 
\end{knitrout}
\caption{The correlation between $X$ and $Z$ has little effect on estimation of $\beta_X$.}
\end{figure}

\begin{figure}
\begin{knitrout}
\definecolor{shadecolor}{rgb}{0.969, 0.969, 0.969}\color{fgcolor}
\includegraphics[width=\maxwidth]{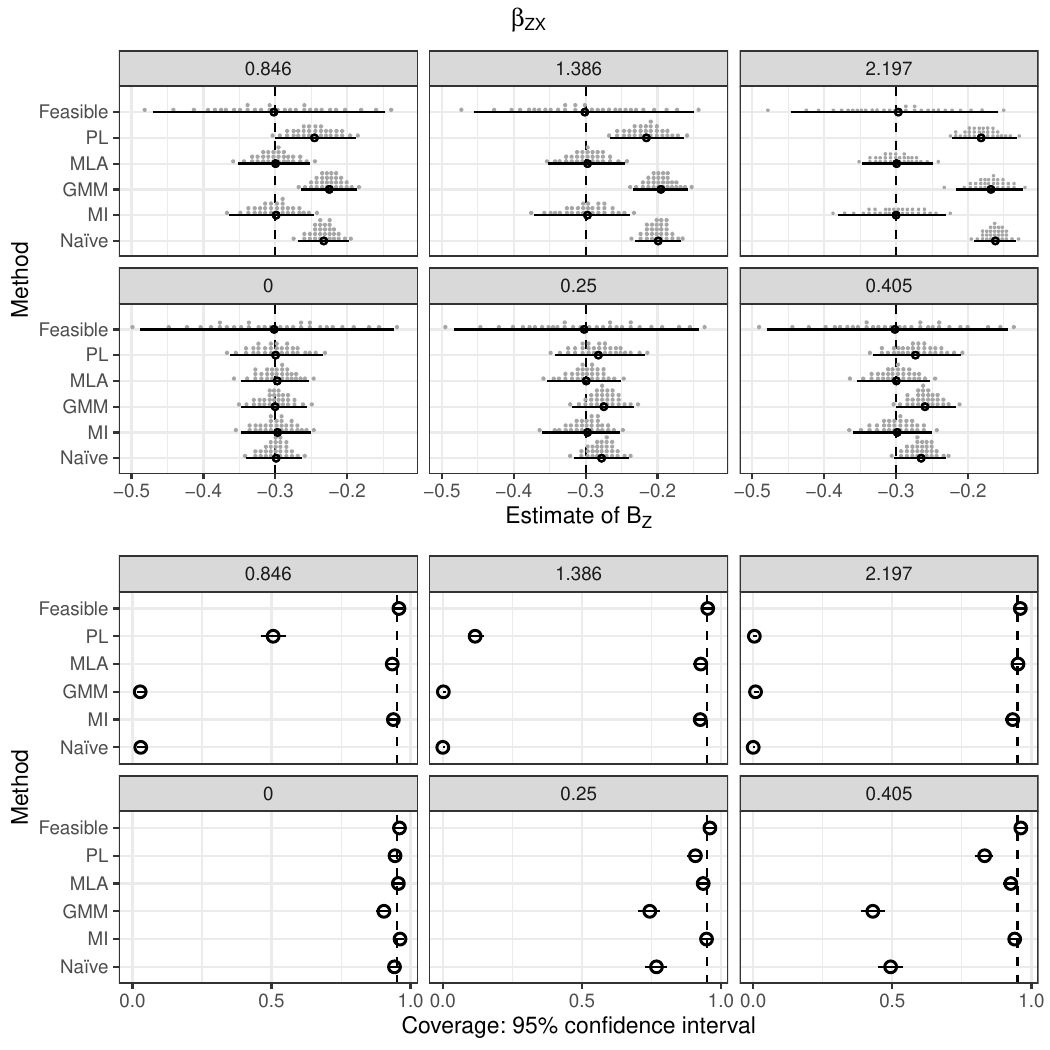} 
\end{knitrout}
\caption{As the correlation between $X$ and $Z$ increases, misclassification of $X$ causes greater bias in estimates of $\beta_X$. MLA and MI are effective in correcting this bias.}
\end{figure}
\clearpage

\subsubsection{Increasing correlation between $X$ and $Z$ with misclassification in the DV}

\begin{figure}
\begin{knitrout}
\definecolor{shadecolor}{rgb}{0.969, 0.969, 0.969}\color{fgcolor}
\includegraphics[width=\maxwidth]{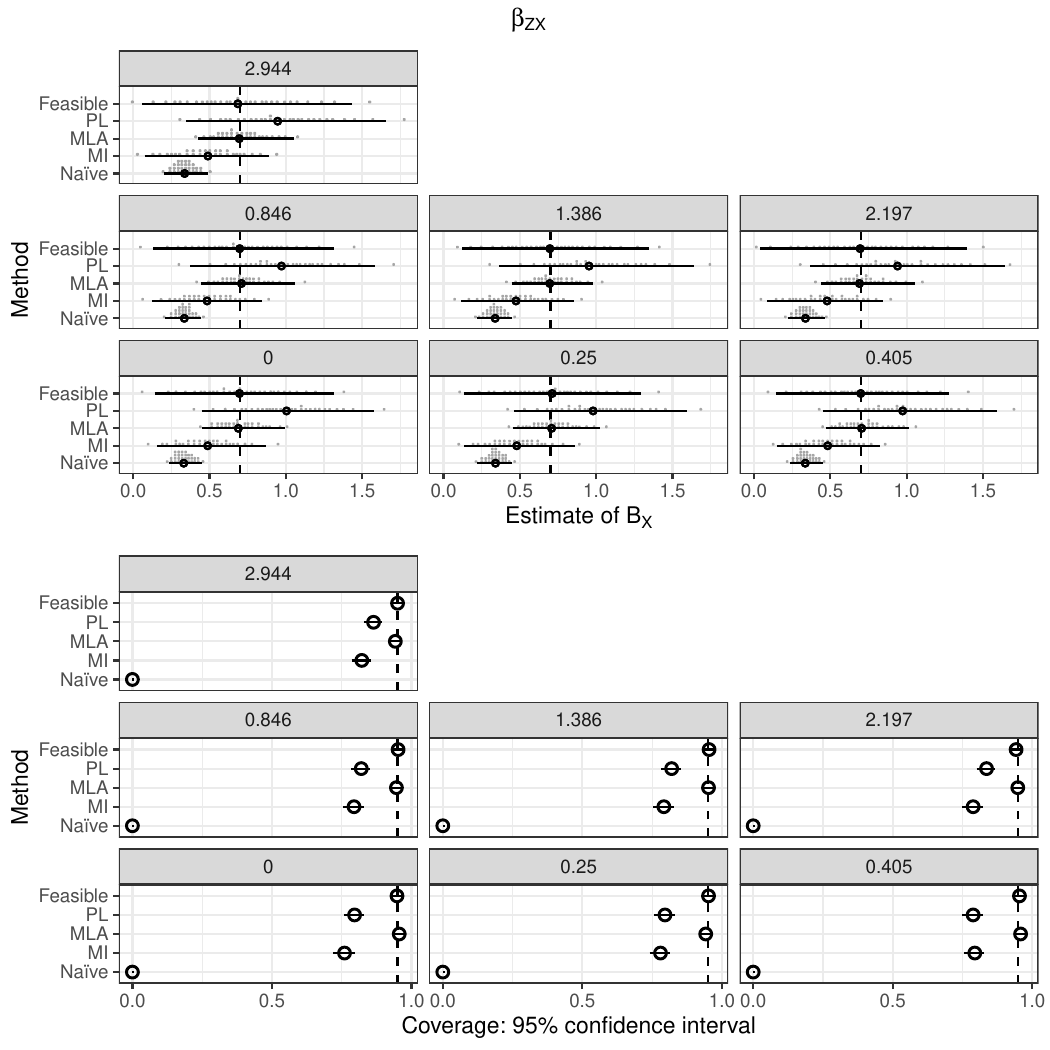} 
\end{knitrout}
\caption{The correlation between $X$ and $Z$ has little effect on estimation of $\beta_X$.}
\end{figure}

\clearpage
\subsection{Robustness Test VI: Non-normal outcomes}
\label{appendix:robustness.vi}
Our main simulations for misclassified independent variables have a normally distributed outcome. Our implementation of MLA in this scenario assumes that the outcome is  normally distributed. Here we explore what happens when this parametric assumption is violated.  Following the ``high signal-to-noise ratio'' simulation from \citet{fong_machine_2021}, we repeat simulations 1a and 1b with 200 annotations and 5,000 classifications and an outcome with a skewed, non-normal distribution.  This distribution is constructed by adding the normal distribution from simulation 1a to a second normal distribution with mean 0 and varying standard deviations ($\sigma_2$). 
 We exclude PL from this robustness check because the its estimates become unstable when the signal-to-noise ratio is high.

\begin{figure}
\begin{knitrout}
\definecolor{shadecolor}{rgb}{0.969, 0.969, 0.969}\color{fgcolor}
\includegraphics[width=\maxwidth]{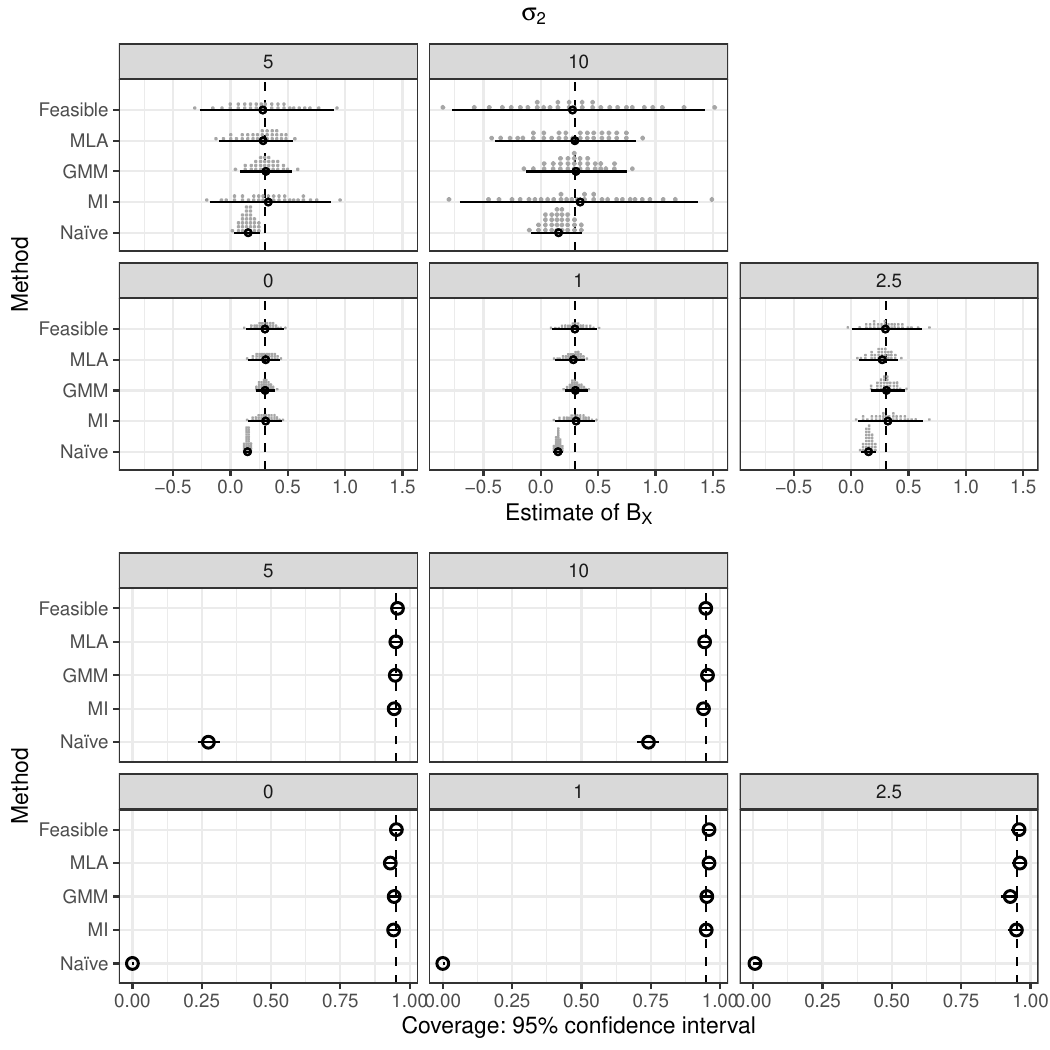} 
\end{knitrout}
\caption{Repeating \textit{Simulation 1a}, as $\sigma_2$ increases, the distribution of the outcome becomes increasingly non-normal, and error correction methods remain effective at estimating $B_X$ even as estimates become less precise. \label{fig:robustness.6a.cor.dv.x}}

\end{figure}

\begin{figure}
\begin{knitrout}
\definecolor{shadecolor}{rgb}{0.969, 0.969, 0.969}\color{fgcolor}
\includegraphics[width=\maxwidth]{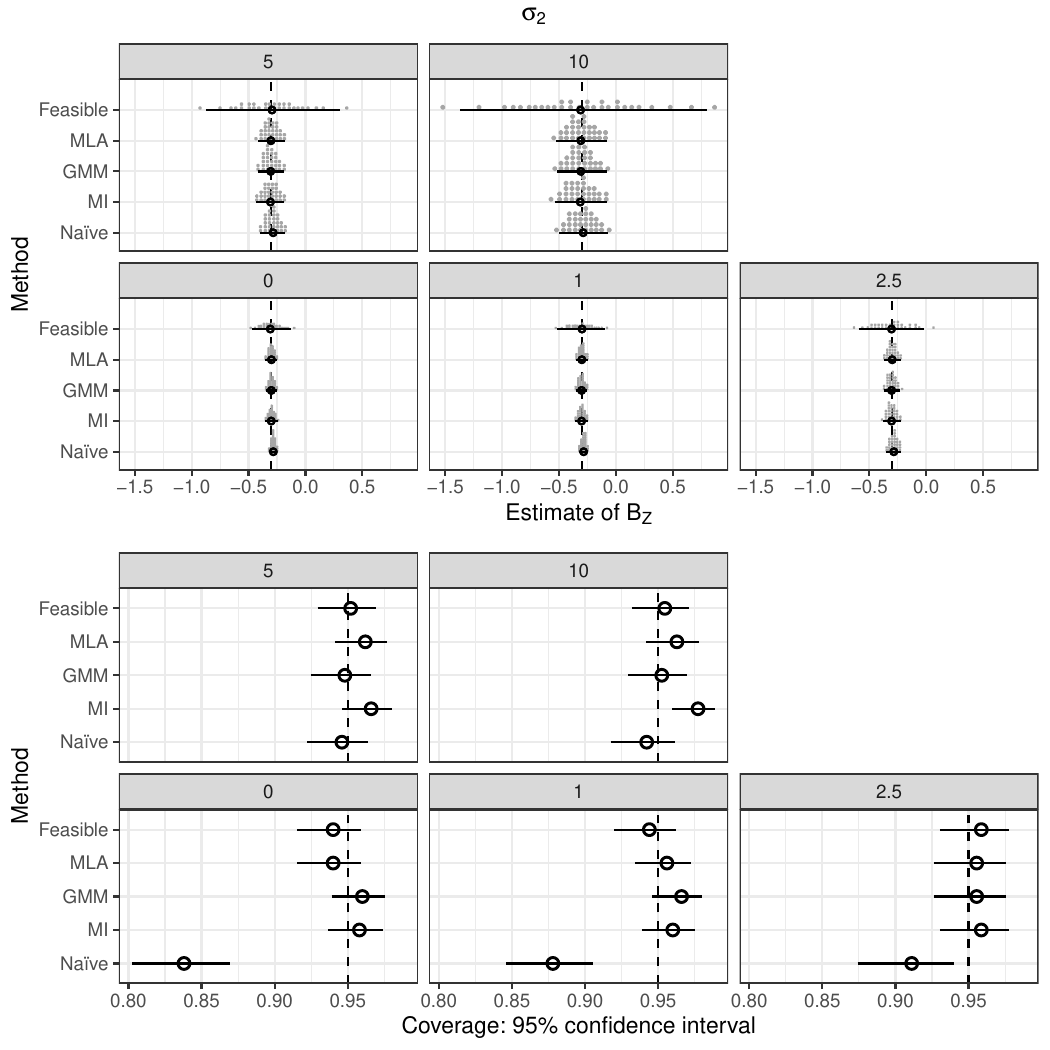} 
\end{knitrout}
\caption{Repeating \textit{Simulation 1a}, as $\sigma_2$ increases, the distribution of the outcome becomes increasingly non-normal, and error correction methods remain effective at estimating $B_Z$ even as estimates become less precise. 
\label{fig:robustness.6a.cor.dv.z}}

\end{figure}

Figures \ref{fig:robustness.6a.cor.dv.x} and \ref{fig:robustness.6a.cor.dv.z} show the results from extending simulation 1a, in which misclassification is random. Perhaps surprisingly, MLA is robust as the outcome becomes increasingly non-normal, however GMM is more robust. 

\begin{figure}
\begin{knitrout}
\definecolor{shadecolor}{rgb}{0.969, 0.969, 0.969}\color{fgcolor}
\includegraphics[width=\maxwidth]{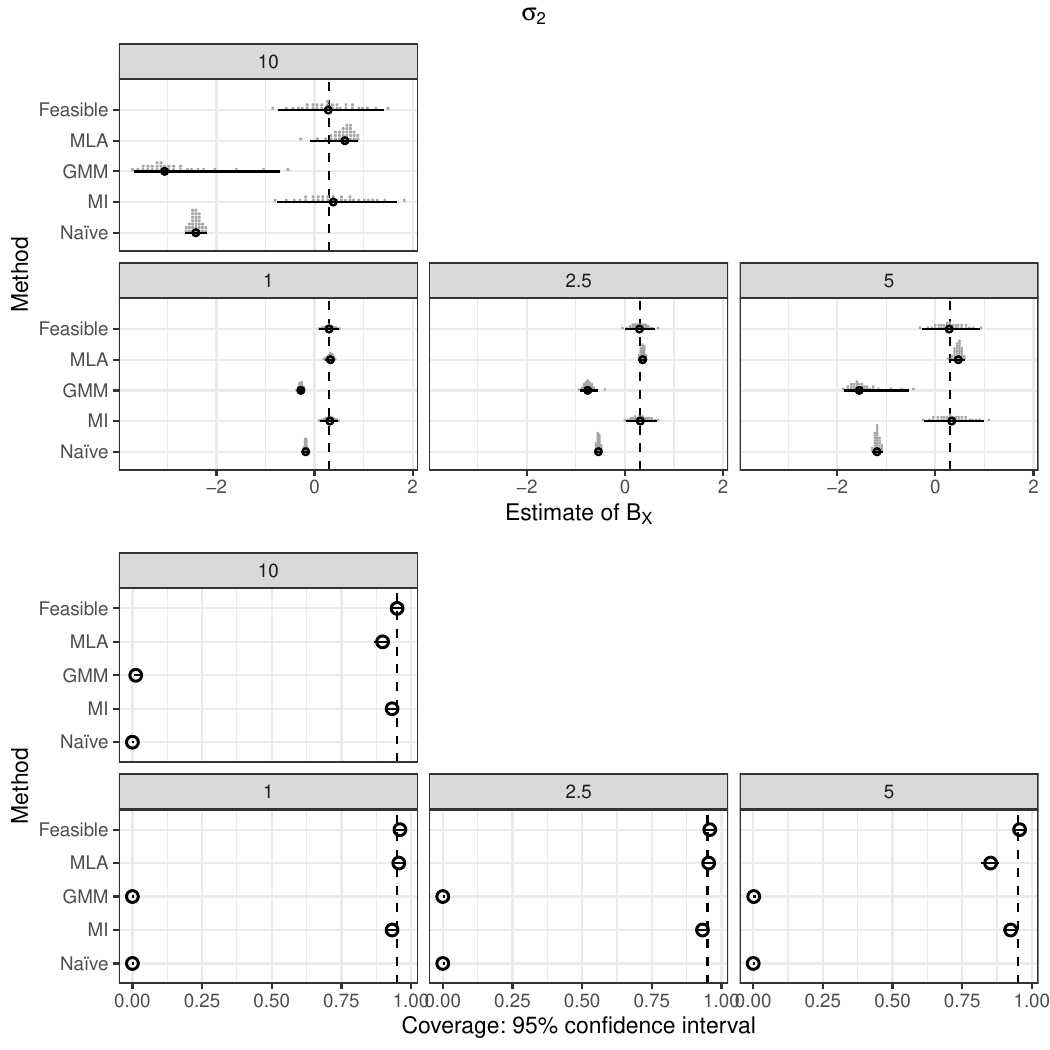} 
\end{knitrout}
\caption{Repeating \textit{Simulation 2b}, as $\sigma_2$ increases, MLA and MI lose effectiveness at estimating $B_x$. MI loses efficiency while MLA is biased and confidence interval coverage degrades. \label{fig:robustness.6b.cor.dv.x}}

\end{figure}

\begin{figure}
\begin{knitrout}
\definecolor{shadecolor}{rgb}{0.969, 0.969, 0.969}\color{fgcolor}
\includegraphics[width=\maxwidth]{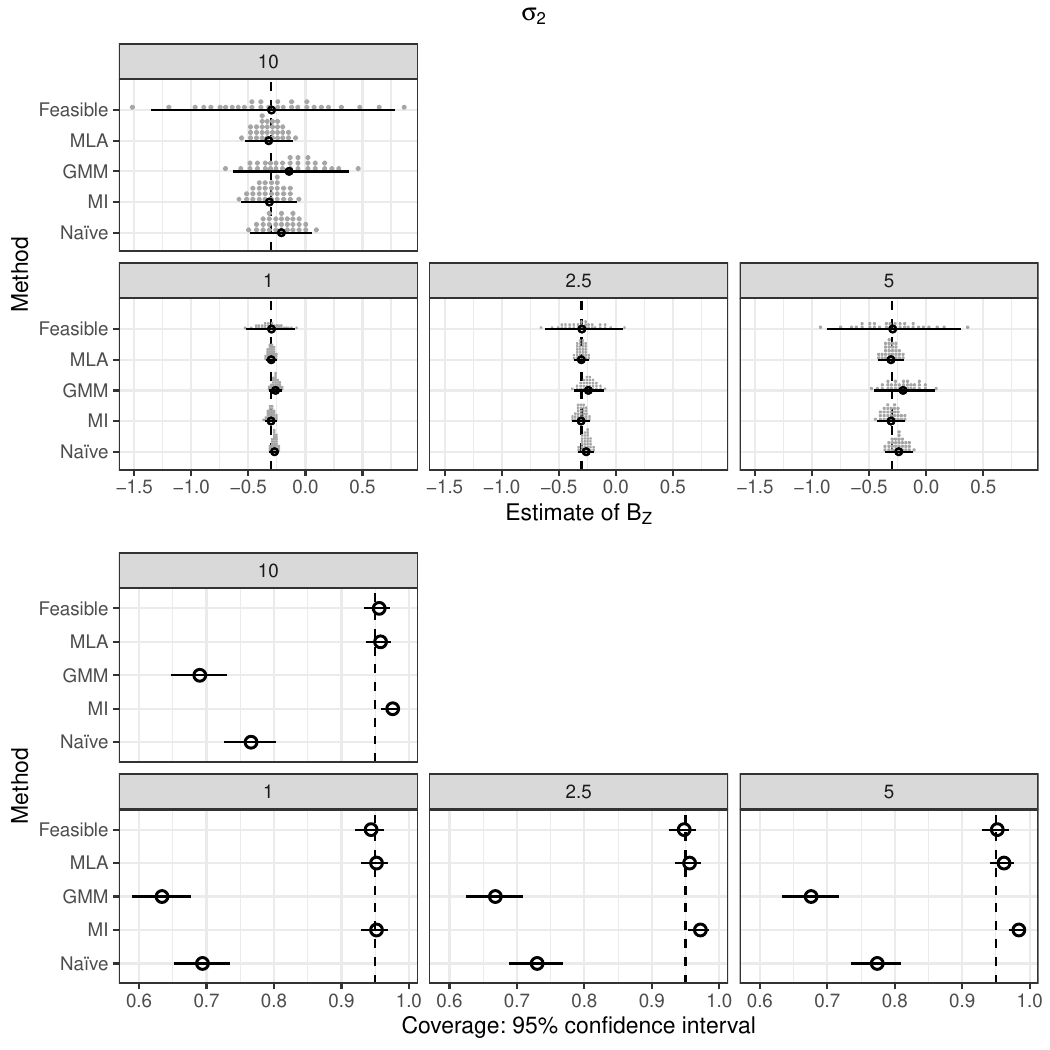} 
\end{knitrout}
\caption{Repeating \textit{Simulation 2b}, as $\sigma_2$ increases, MLA and MI remain effective at estimating $B_Z$.  \label{fig:robustness.6b.cor.dv.z}}

\end{figure}

Figures \ref{fig:robustness.6b.cor.dv.x} and \ref{fig:robustness.6b.cor.dv.z} extend simulation 1b where misclassification is systematic (differential error) and make the trade-off between GMM and MLA clear.  GMM is theoretically robust to the distribution of the outcome, but GMM is unable to correct differential error. MLA  is sensitive to parametric assumptions as its bias and imperfect confidence interval coverage when the outcome becomes increasingly non-normal demonstrate, but it is relatively effective when misclassification in an IV is correlated with the DV.  Note that this limitation of MLA can be corrected by using the appropriate parametric distribution in the likelihood. When differential misclassification is strong, such as in simulation 1b,  GMM's estimates can be more biased than MLA's.

\end{document}